\documentclass{article} % For LaTeX2e
\usepackage{iclr2027_conference,times}

\newif\iflink
\linktrue

\newif\ifarxiv
\arxivtrue      % arXiv version
\newcommand{\rev}[1]{#1}

\usepackage{amsmath,amsfonts,bm}

\def\eqref#1{equation~\ref{#1}}
\def\1{\bm{1}}

\DeclareMathAlphabet{\mathsfit}{\encodingdefault}{\sfdefault}{m}{sl}
\SetMathAlphabet{\mathsfit}{bold}{\encodingdefault}{\sfdefault}{bx}{n}

\usepackage{hyperref}
\usepackage{url}
 \usepackage{booktabs}
 \usepackage{helvet}
\usepackage{amssymb,mathtools,amsmath}
\usepackage{longtable,tabularx}
\usepackage{array}
\usepackage{multirow}
\usepackage[table]{xcolor}

\usepackage{subcaption}

\usepackage{tikz}
\usetikzlibrary{
    arrows.meta,
    positioning,
    calc,
    fit,
    backgrounds,
    decorations.pathreplacing,
    shapes.geometric
}

\usepackage{threeparttable}

\usepackage{xcolor}
\definecolor{actLight}{HTML}{AEC7E8}   \definecolor{actDark}{HTML}{1F77B4}
\definecolor{agLight}{HTML}{98DF8A}    \definecolor{agDark}{HTML}{2CA02C}
\definecolor{pgLight}{HTML}{FF9896}    \definecolor{pgDark}{HTML}{D62728}
\definecolor{saeKone}{HTML}{8C564B}    \definecolor{saeKstar}{HTML}{E377C2}

\definecolor{tfblue}{HTML}{1F5AA6}
\definecolor{tfpurple}{HTML}{6543A5}
\definecolor{tfgreen}{HTML}{287A52}
\definecolor{tforange}{HTML}{D86B16}
\definecolor{tfgray}{HTML}{666666}
\definecolor{tflightgray}{HTML}{F5F5F5}
\definecolor{tflightblue}{HTML}{F4F8FF}
\definecolor{tflightpurple}{HTML}{F8F4FF}
\definecolor{tflightgreen}{HTML}{F4FBF7}
\definecolor{tflightorange}{HTML}{FFF7F0}
\definecolor{tfline}{HTML}{B9B9B9}

\tikzset{
  panel/.style={draw=tfgray, rounded corners=2.5mm, line width=.8pt, fill=white},
  subpanel/.style={draw=tfblue!70, rounded corners=1.8mm, line width=.7pt, fill=white},
  box/.style={draw=tfgray!70, rounded corners=1.2mm, line width=.55pt, fill=white, align=left},
  bluebox/.style={draw=tfblue!75, rounded corners=1.2mm, line width=.65pt, fill=tflightblue, align=center},
  purplebox/.style={draw=tfpurple!75, rounded corners=1.2mm, line width=.65pt, fill=tflightpurple, align=center},
  greenbox/.style={draw=tfgreen!80, rounded corners=1.2mm, line width=.65pt, fill=tflightgreen, align=center},
  orangebox/.style={draw=tforange!85, rounded corners=1.2mm, line width=.65pt, fill=tflightorange, align=center},
  flow/.style={-{Stealth[length=2.8mm,width=2.2mm]}, draw=tfgray!75, line width=1.1pt},
  panelflow/.style={-{Stealth[length=4.8mm,width=5.0mm]}, draw=tfgray!70, line width=2.2pt},
  thinflow/.style={-{Stealth[length=2.1mm,width=1.6mm]}, draw=tfgray!80, line width=.7pt},
  methodprior/.style={draw=tfgray!70, rounded corners=1.6mm, line width=.65pt, fill=white, align=center},
  methodnew/.style={draw=tfpurple!80, rounded corners=1.6mm, line width=.9pt, fill=tflightpurple, align=center},
}

\usepackage[hyperfirst=false, nonumberlist, nostyles, nogroupskip]{glossaries}
\glsdisablehyper 
\glsunsetall

\newacronym{gradiend}{GRADIEND}{GRADient IEND}
\newacronym{actiend}{ACTIEND}{ACTivation IEND}
\newacronym{iend}{IEND}{Interpretable ENcoder Decoder}
\newacronym{agiend}{AGIEND}{Activation Gradient IEND}

\newacronym{sae}{SAE}{Sparse Autoencoder}
\newacronym{caa}{CAA}{Contrastive Activation Addition}
\newacronym{cga}{CGA}{Contrastive Gradient Addition}
\newacronym{caga}{CAGA}{Contrastive Activation Gradient Addition}

\newcommand{\iend}{\acrshort{iend}}
\newcommand{\gradiend}{\acrshort{gradiend}}
\newcommand{\actiend}{\acrshort{actiend}}
\newcommand{\sae}{\acrshort{sae}}
\newcommand{\caa}{\acrshort{caa}}
\newcommand{\cga}{\acrshort{cga}}
\newcommand{\agiend}{\acrshort{agiend}}
\newcommand{\caga}{\acrshort{caga}}

\usepackage{graphicx}
\usepackage{pgfplots}
\pgfplotsset{compat=1.18}

\ifarxiv
  \newcommand{\taskicon}[2][0.85]{%
    \scalebox{#1}{%
      \raisebox{-0.15em}{%
        \includegraphics[height=1em]{img/fontawesome_task_icons/#2.pdf}%
      }%
    }%
  }
    \newcommand{\ctaskicon}[2][0.85]{%
    \scalebox{#1}{%
      \raisebox{-0.15em}{%
        \includegraphics[height=1em]{img/fontawesome_task_icons_colored/#2.pdf}%
      }%
    }%
  }
\else
  \usepackage{fontawesome7}

  \newcommand{\taskicon}[2][0.85]{%
    \scalebox{#1}{\faIcon{#2}}%
  }
    \newcommand{\ctaskicon}[2][0.85]{%
    \scalebox{#1}{\faIcon{#2}}%
  }
\fi

\newcommand{\taskGender}{\mbox{\taskicon{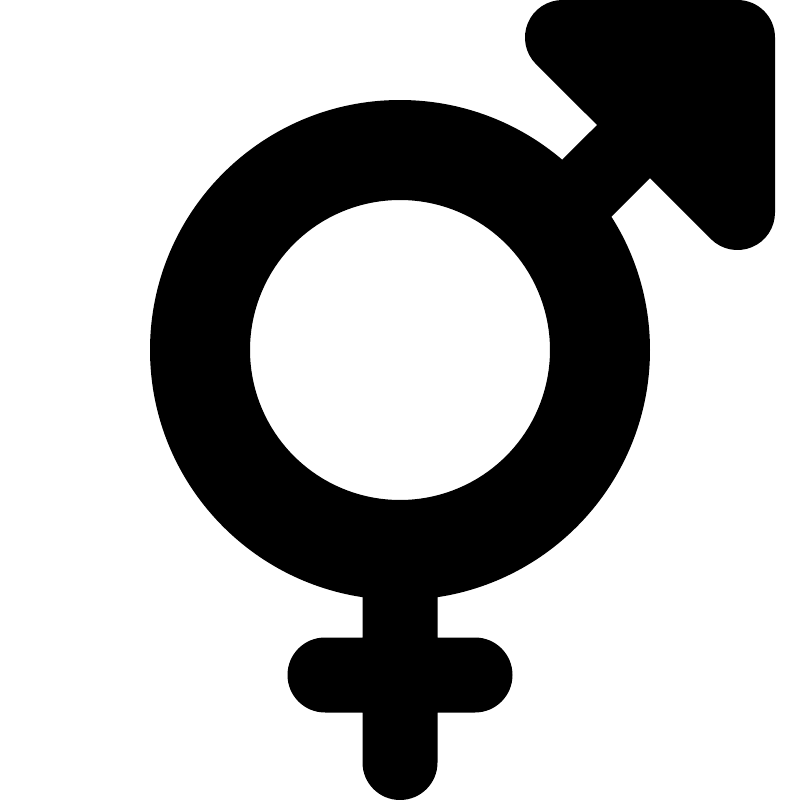}\hspace{1pt}\textsc{Gender}}}    % \faIcon{mars-and-venus}, \faIcon{person-half-dress}, \faIcon{genderless}
\newcommand{\taskEmotion}{\mbox{\taskicon{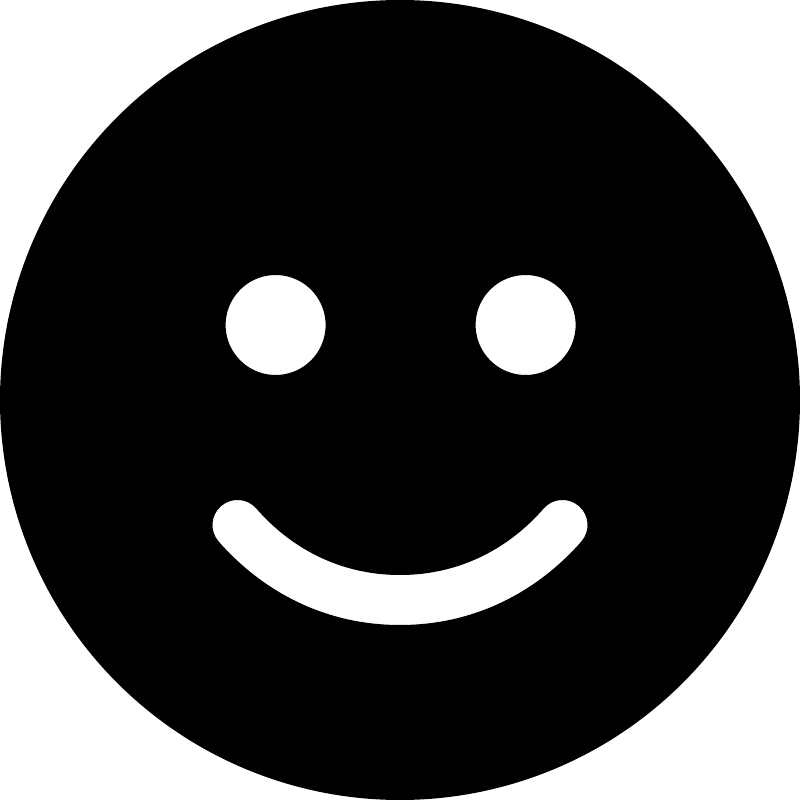}\hspace{1pt}\textsc{Emotion}}}     % \faIcon{masks-theater}, \faIcon{face-smile}, \faIcon{heart}
\newcommand{\taskRace}{\mbox{\taskicon{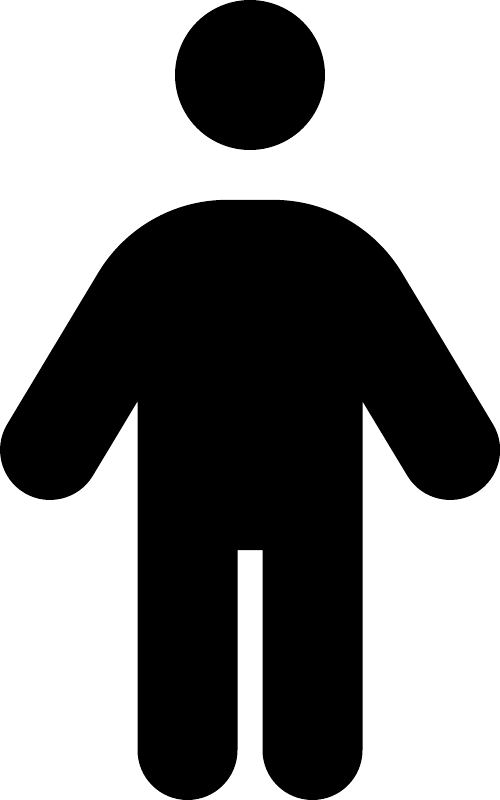}\hspace{0.5pt}\textsc{Race}}}             % \faIcon{people-group}, \faIcon{people-line}, \faIcon{person}
\newcommand{\taskReligion}{\mbox{\taskicon{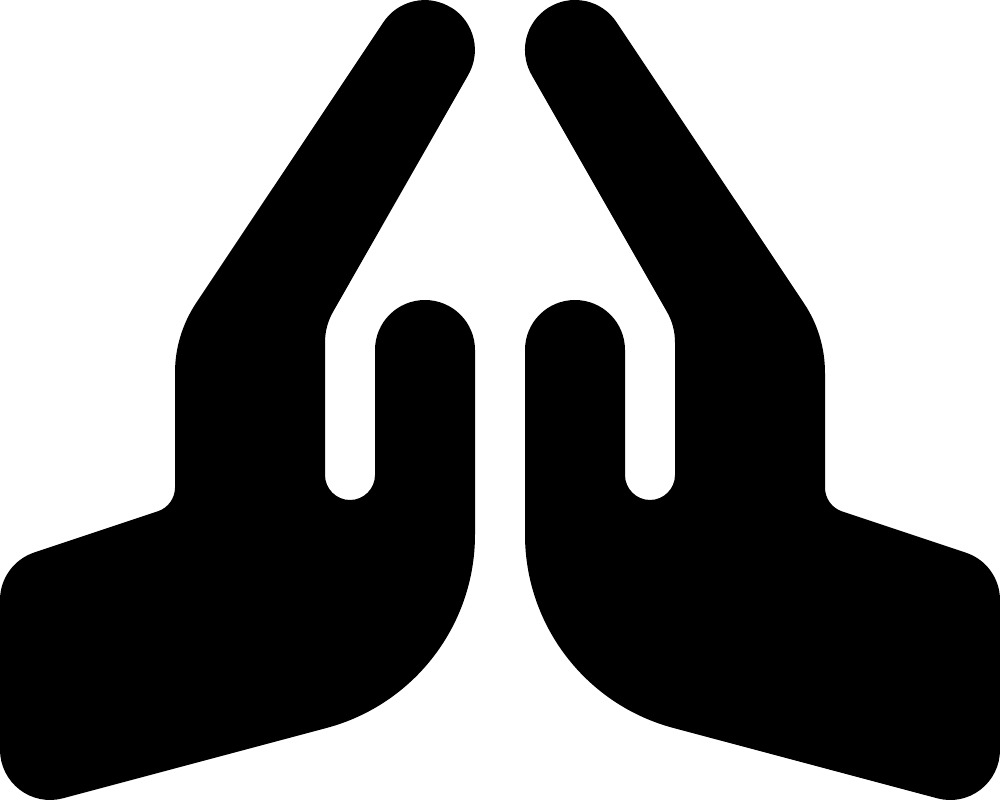}\hspace{1pt}\textsc{Religion}}} % \faIcon{hands-praying}, \faIcon{person-praying}, \faIcon{landmark}

\newcommand{\taskPronNum}{\mbox{\taskicon{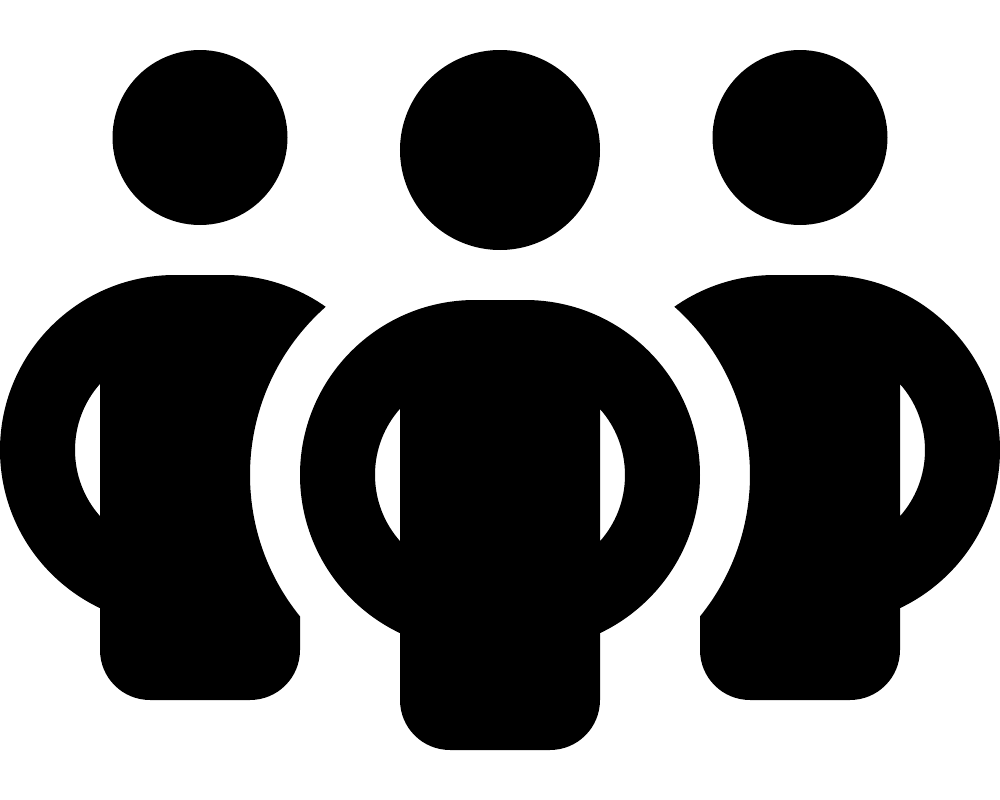}\hspace{1pt}\textsc{PronNum}}}     % \faIcon{people-group}, \faIcon{person}, \faIcon{people-line}
\newcommand{\taskPronPers}{\mbox{\taskicon{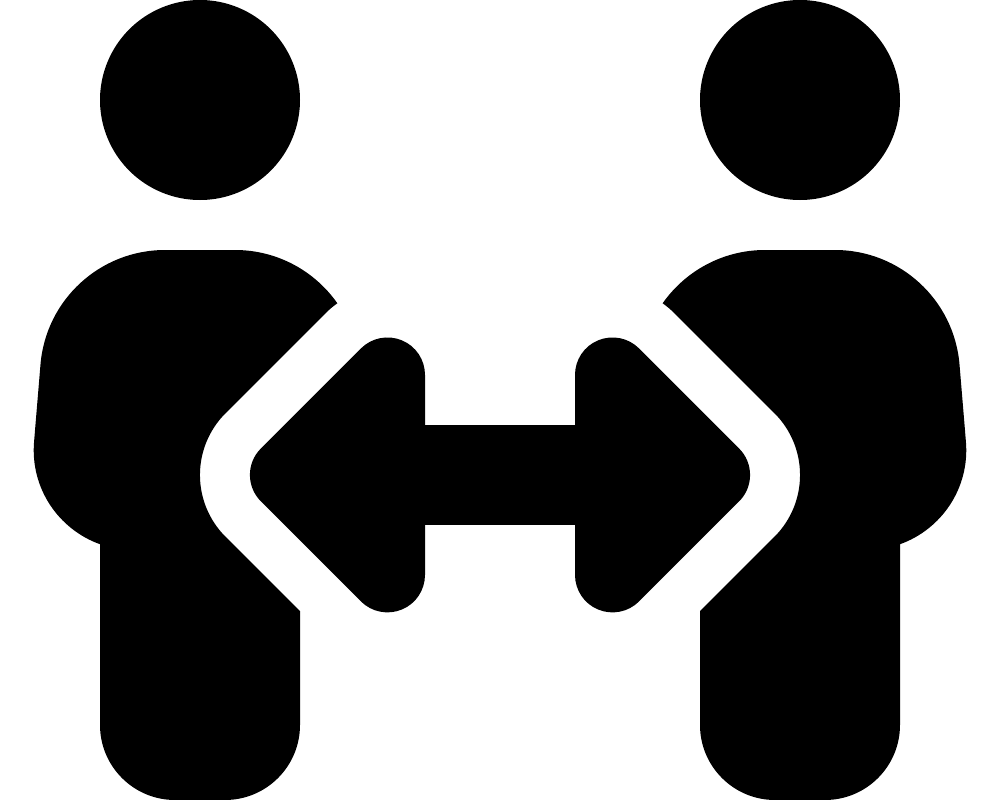}\hspace{1pt}\textsc{PronPers}}} % \faIcon{people-arrows}, \faIcon{id-card}, \faIcon{person-circle-question}

\newcommand{\taskRavelCont}{\mbox{\taskicon{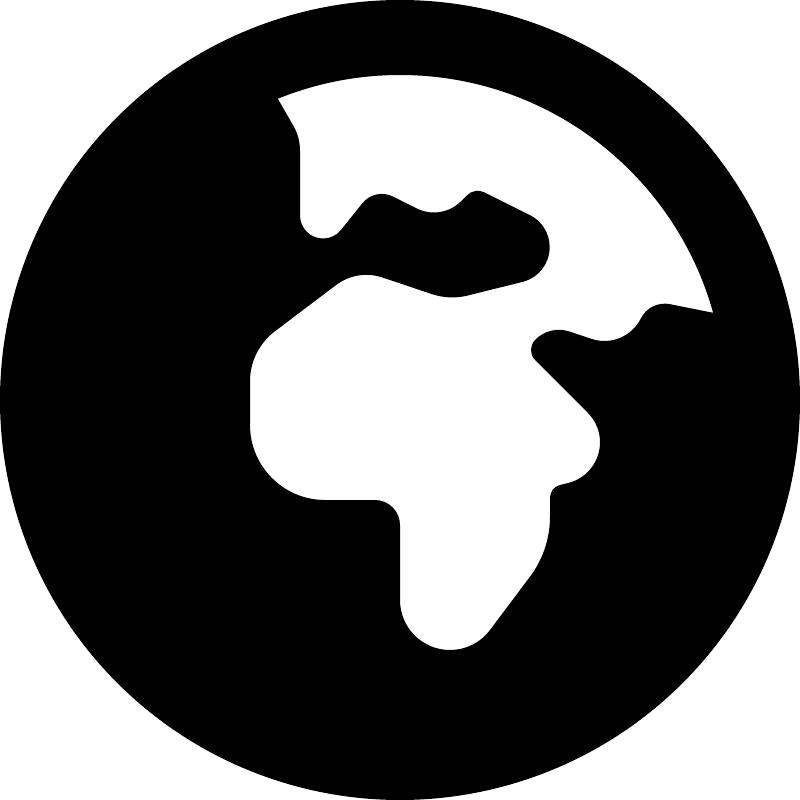}\hspace{1pt}\textsc{RavelCont}}}     % \faIcon{globe}, \faIcon{earth-africa}, \faIcon{earth-europe}
\newcommand{\taskRavelCountry}{\mbox{\taskicon{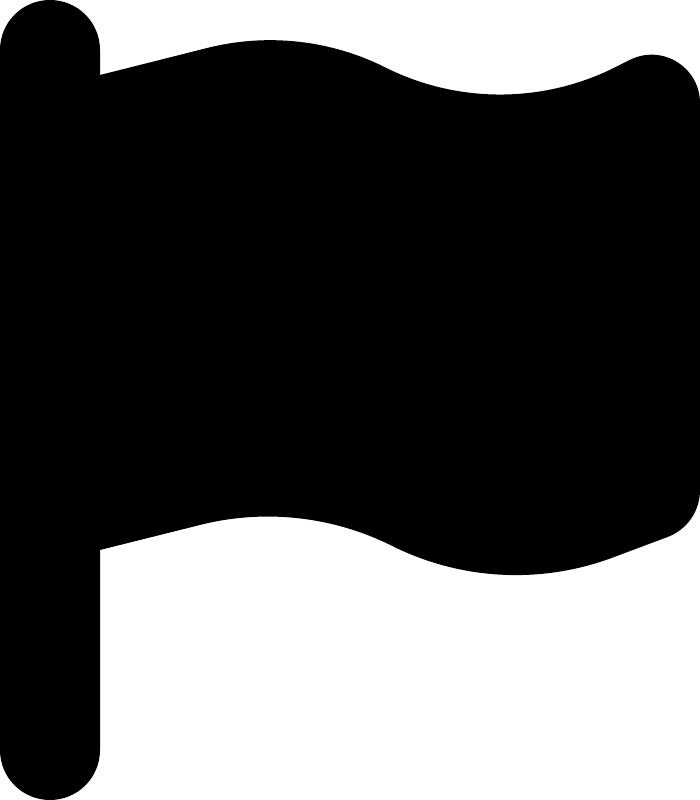}\hspace{1pt}\textsc{RavelCntry}}}         % \faIcon{flag}, \faIcon{map-location-dot}, \faIcon{location-dot}
\newcommand{\taskRavelLang}{\mbox{\taskicon{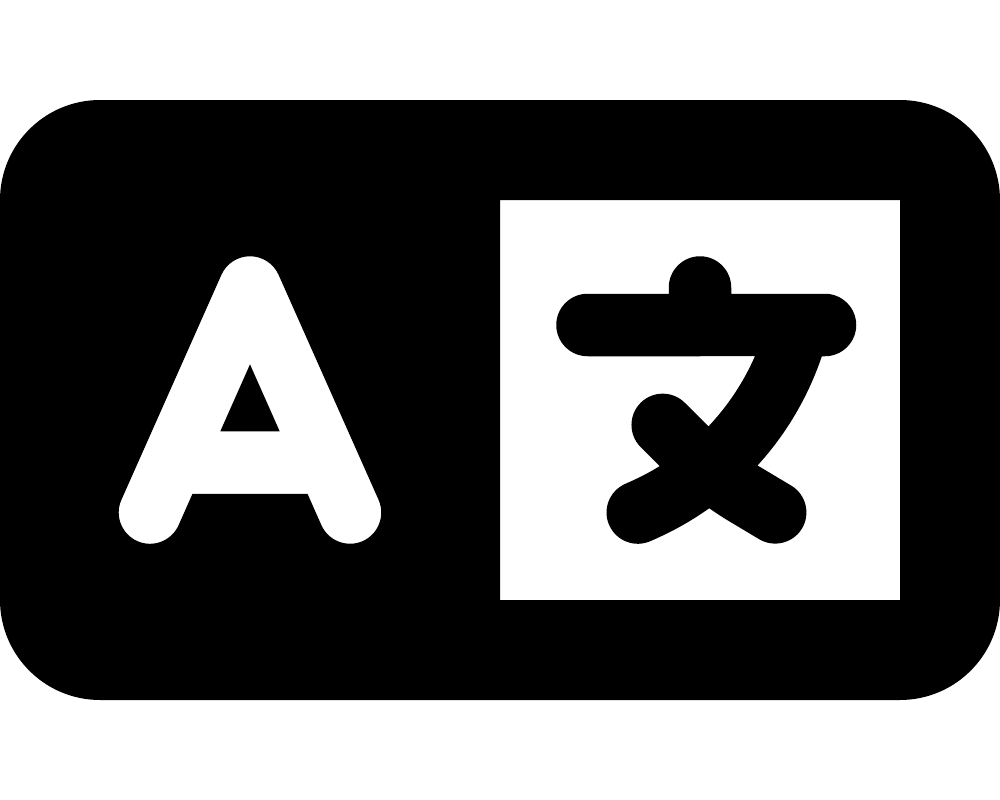}\hspace{1pt}\textsc{RavelLang}}}         % \faIcon{language}, \faIcon{comment-dots}, \faIcon{comments}

\newcommand{\taskLangID}{\mbox{\taskicon{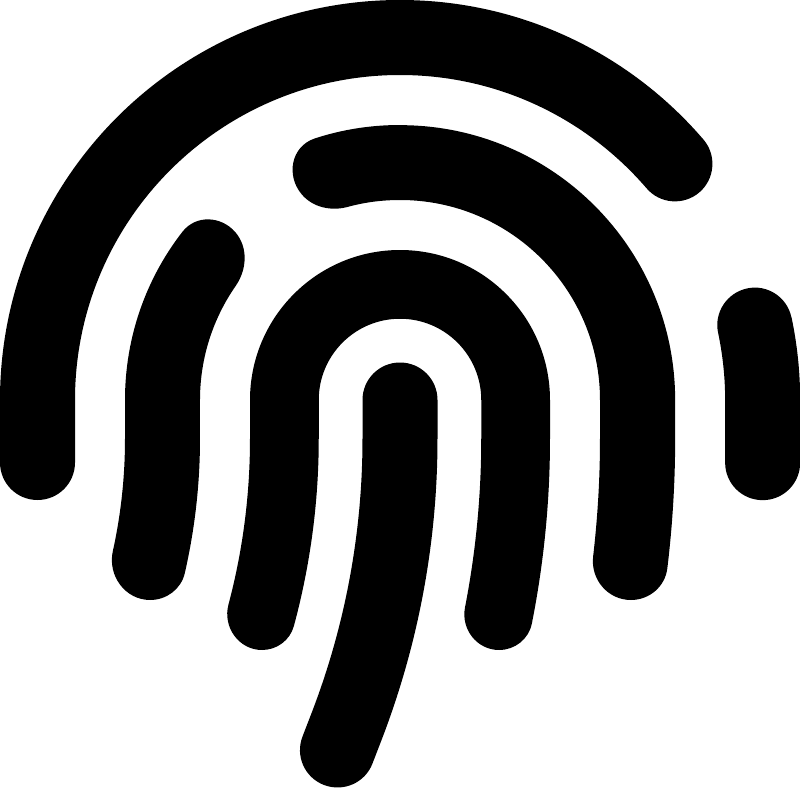}\hspace{1pt}\textsc{LangID}}}       % \faIcon{fingerprint}, \faIcon{magnifying-glass}, \faIcon{font}

\newcommand{\taskMIBIOI}{\mbox{\taskicon{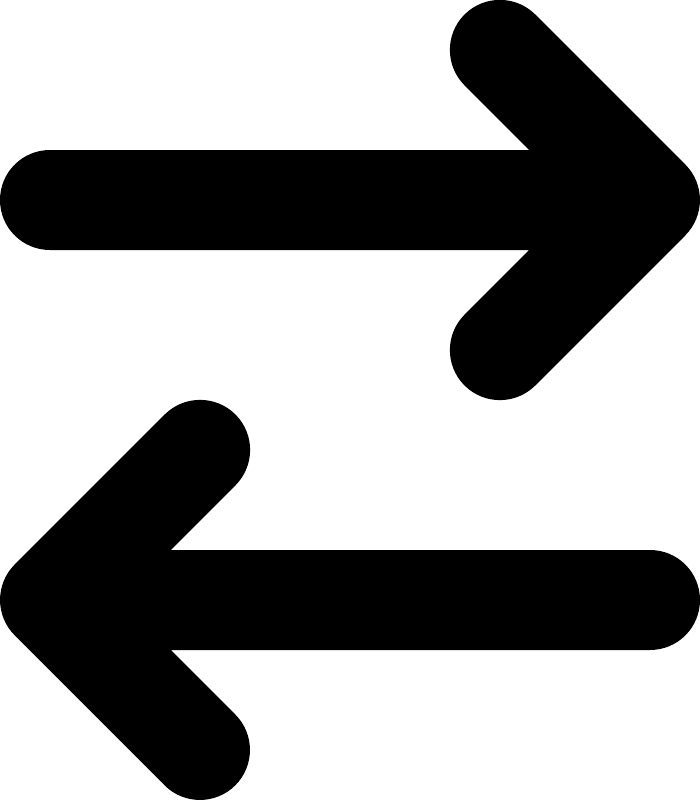}\hspace{1pt}\textsc{IOI}}} % \faIcon{people-arrows}, \faIcon{person-circle-question}, \faIcon{arrow-right-arrow-left}
\newcommand{\taskKeyValue}{\mbox{\taskicon{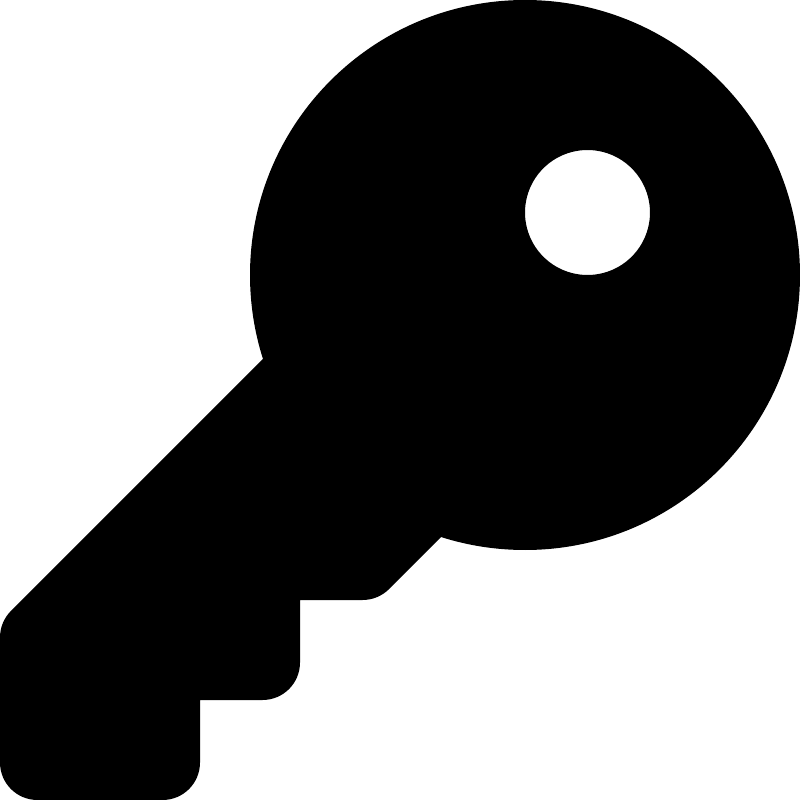}\hspace{1pt}\textsc{KeyValue}}}     % \faIcon{key}, \faIcon{database}, \faIcon{list}
\newcommand{\taskInduction}{\mbox{\taskicon{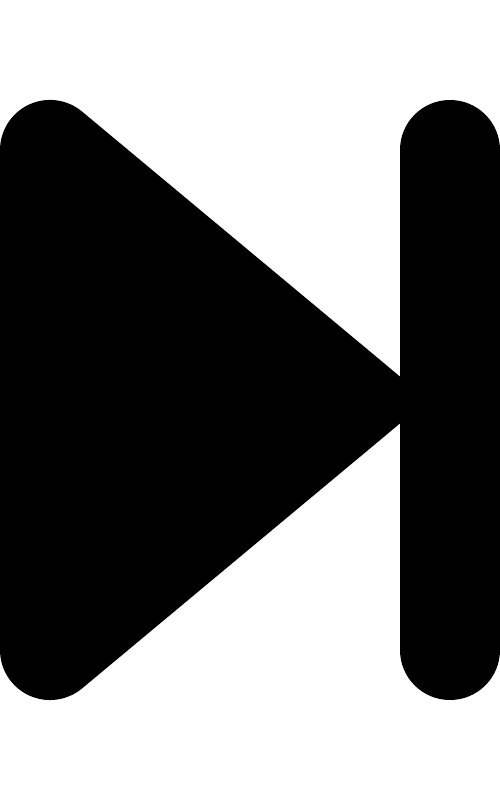}\hspace{1pt}\textsc{Induction}}} % \faIcon{forward-step}, \faIcon{diagram-next}, \faIcon{arrow-right}
\newcommand{\taskRepetition}{\mbox{\taskicon{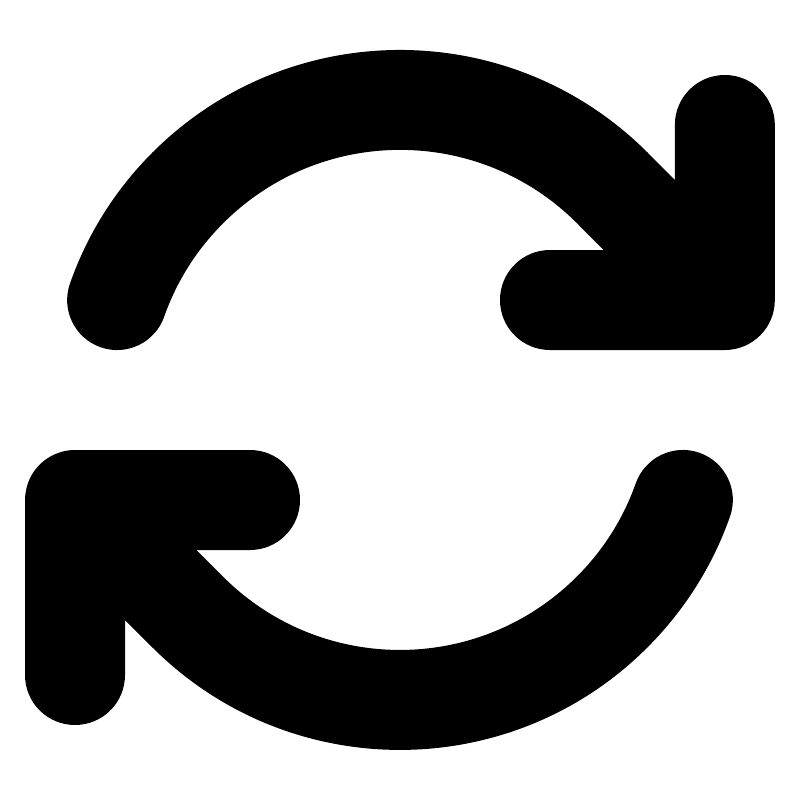}\hspace{1pt}\textsc{Repetition}}} % \faIcon{arrows-rotate}, \faIcon{group-arrows-rotate}, \faIcon{infinity}
\newcommand{\taskFuncComp}{\mbox{\taskicon{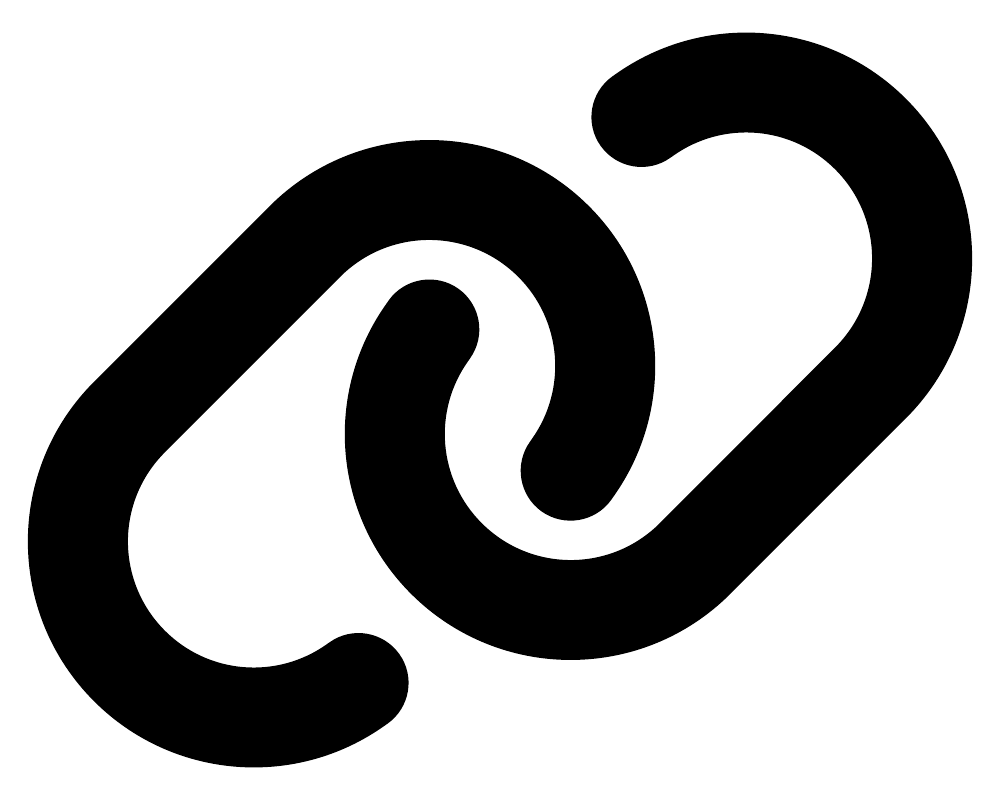}\hspace{1pt}\textsc{FuncComp}}}           % \faIcon{code-branch}, \faIcon{diagram-project}, \faIcon{link}

\usepackage{colortbl}

\providecommand{\MethodSAE}{$\mathrm{SAE}$}
\providecommand{\MethodSAEkOne}{$\mathrm{SAE}^{k=1}$}

\providecommand{\SummarySignalRule}[1]{\midrule}
\providecommand{\SummaryMethodRule}[1]{%
  \arrayrulecolor{gray!30}%
  \specialrule{0.4pt}{0.3pt}{0.3pt}%
  \arrayrulecolor{black}%
}

\definecolor{tfred}{HTML}{B64040}

\colorlet{cAct}{tfgreen}
\colorlet{cActGrad}{tforange}
\colorlet{cParamGrad}{tfred}

\definecolor{tfaccent}{HTML}{008C95}

\colorlet{cAct}{actDark}
\colorlet{cActGrad}{agDark}
\definecolor{tfred}{HTML}{B64040}
\colorlet{cParamGrad}{pgDark}

\definecolor{tfaccent}{HTML}{C08A00} % ochre / gold
\definecolor{tfmean}{HTML}{6A3D9A}   % purple
\definecolor{tfiend}{HTML}{007C91}   % deep cyan

\colorlet{cAccent}{tfaccent}
\colorlet{cMean}{tfmean}
\colorlet{cIEND}{tfiend}

\tikzset{
  stepbadge/.style={
    rounded corners=0.75mm,
    draw=cAccent!55,
    fill=cAccent!14,
    line width=0.35pt,
    minimum width=3.8mm,
    minimum height=2.9mm,
    inner sep=0pt,
    font=\fontsize{3.9pt}{4.2pt}\selectfont\bfseries,
    text=cAccent!95!black,
    align=center
  },
  headericon/.style={
    text=cAccent!95!black
  }
}

\newcommand{\PlaceHeader}[4]{%

  \def\HeaderCut{2mm}

  \begin{scope}
    \clip[rounded corners=2mm]
      ([yshift=-\HeaderCut]#1.north west)
      rectangle
      (#1.south east);

    \fill[white]
      ([yshift=-\HeaderCut]#1.north west)
      rectangle
      (#1.south east);

    \fill[cAccent!9]
      ([yshift=-\HeaderCut]#1.north west)
      rectangle
      ([yshift=-6.2mm]#1.north east);
  \end{scope}

  \draw[
    draw=cAccent!25,
    line width=0.28pt
  ]
    ([yshift=-6.2mm]#1.north west)
    --
    ([yshift=-6.2mm]#1.north east);

  \node[
    rounded corners=1mm,
    fill=cAccent,
    draw=cAccent,
    minimum width=3.2mm,
    minimum height=3.2mm,
    inner sep=0pt,
    font=\fontsize{6pt}{4.4pt}\selectfont\bfseries,
    text=white,
    anchor=north west
  ]
  at ([xshift=0.5mm,yshift=-2.45mm]#1.north west)
  {#2};

  \node[
    anchor=north west,
    font=\scriptsize\bfseries,
    text=cAccent!95!black
  ]
  at ([xshift=4.3mm,yshift=-2.05mm]#1.north west)
  {#3};

  \node[
    anchor=north east,
    text=cAccent!95!black,
    scale=1.0
  ]
  at ([xshift=-0.1mm,yshift=-1.15mm]#1.north east)
  {#4};

  \draw[
    draw=cAccent!55,
    rounded corners=2mm,
    line width=0.4pt
  ]
    ([yshift=-\HeaderCut]#1.north west)
    rectangle
    (#1.south east);
}

\newcommand{\gpttwo}{\mbox{GPT-2}}
\newcommand{\pythia}{\mbox{Pythia-70M}}
\newcommand{\gemmaS}{\mbox{Gemma-3-1B}}
\newcommand{\qwenS}{\mbox{Qwen-3.5-2B}}
\newcommand{\gemma}{\mbox{Gemma-2-2B}}
\newcommand{\gemmaM}{\mbox{Gemma-3-4B}}
\newcommand{\mistral}{\mbox{Mistral-7B}}
\newcommand{\llama}{\mbox{Llama-3.1-8B}}
\newcommand{\deepseek}{\mbox{DeepSeek-R1-8B}}
\newcommand{\qwenL}{\mbox{Qwen-3.5-9B}}

\title{Gradients for Interventions and Activation \rev{Values} for Detection: Targeted Feature Learning in Language Models}
\author{Jonathan Drechsel \& Steffen Herbold \\
Faculty of Computer Science and Mathematics \\
University of Passau\\
Passau, Germany \\
\texttt{\{jonathan.drechsel,steffen.herbold\}@uni-passau.de}
}

\providecommand{\caga}{\textsc{caga}}
\providecommand{\agiend}{\textsc{agiend}}
\iclrfinalcopy % Uncomment for camera-ready version, but NOT for submission. % todo uncomment Published as... again
\begin{document}

\maketitle

\begin{abstract}
Model-internal features can be studied through both their ability to identify a specified concept and their causal effect when manipulated, e.g., through steering or weight editing.
A prominent approach to feature learning is Sparse Autoencoders (SAEs), which learn broad feature dictionaries whose relation to particular concepts is typically identified post hoc.
However, many interpretability questions are instead hypothesis-driven and concern a concept specified in advance.
We study this setting as \emph{targeted feature learning}, where a single feature is constructed for such a predefined concept.
We present a controlled comparison across three model signals (activation values, activation gradients, and parameter gradients) and two estimators (contrastive mean and a learned one-dimensional encoder-decoder),
yielding six targeted methods, with \caa\ and \gradiend\ as existing instances and four new methods covering the remaining combinations.
We compare these methods against pretrained SAEs across 15 tasks and ten language models, evaluating both detection and causal intervention.
Across models, the strongest detection performance is achieved by contrastive activation value methods, whereas the strongest intervention performance is achieved by gradient \rev{signals, most strongly with a contrastive mean}.
\rev{Notably, within the contrastive mean family, activation gradients substantially improve intervention over activation values without requiring parameter space interventions.}
Overall, our results show that targeted feature quality depends jointly on the model signal and estimator, with detection and intervention capturing complementary properties.
\end{abstract}

\begin{figure}[h]
    \centering
    \vspace{-5pt}\includegraphics[width=0.75\linewidth]{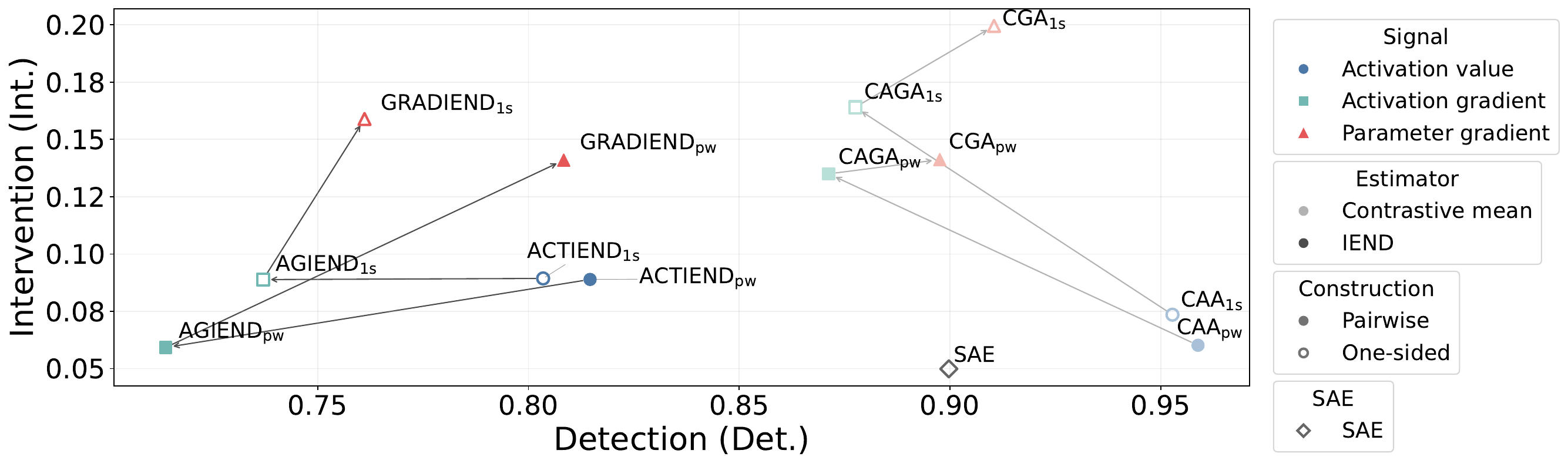}
    \vspace{-5pt}
\caption{\textbf{Detection and intervention capture complementary properties of
targeted features.}
Across 15 tasks and \rev{ten} language models, activation value methods show stronger
detection and most gradient-based stronger intervention overall.
Per-model results are shown in Figure~\ref{fig:detection-vs-intervention}.
}
    \label{fig:detection-intervention-mean}
\end{figure}

\begin{figure*}[!t]
\centering
\resizebox{\textwidth}{!}{%

}
\vspace{-15pt}
\caption{\textbf{Study overview.}
From a predefined feature specification, we cross three model signals and two estimators, yielding six targeted feature learning methods evaluated on detection and intervention.
SAEs provide an untargeted activation value reference closest to \actiend.
}
\label{fig:overview}
\vspace{-5pt}
\end{figure*}

\section{Introduction}

\acrlong{sae}s (\acrshort{sae}\glsunset{sae};
\cite{bricken2023monosemanticity,huben2024sparse}) have become a prominent feature learning approach in mechanistic
interpretability \citep{templeton024}, learning large dictionaries of
activation features whose relation to particular concepts is typically
identified post hoc
\citep{arad-etal-2025-saes}.
Many scientific questions are instead hypothesis-driven, targeting
%Researchers may want to identify features representing particular
linguistic %phenomena
\citep{jing-etal-2025-lingualens,drechsel-etal-2026-understanding},
semantic %relations
\citep{diera-scherp-2026-language},
or demographic %characteristics
\citep{shan-mueller-2026-measuring} features,
and testing whether they  generalize and causally influence model
behavior \citep{zou2023transparency,wu2025axbench}.

This motivates a complementary setting that we call \textbf{targeted learning}: learning a \emph{single} feature for a concept specified in advance.
Besides avoiding post-hoc feature selection, targeting a desired level of abstraction can address \emph{feature splitting} and \emph{feature absorption} in \glspl{sae}, which may divide broader concepts or leave systematic gaps in otherwise interpretable features
\citep{chanin_feature_splitting_absorption,pmlr-v267-bussmann25a}.
A targeted feature should distinguish the specified concept from relevant
alternatives and support interventions that causally change the corresponding
model behavior.

%Among existing targeted approaches, activation steering methods learn directions that can be added to internal representations to control model behavior \citep{actadd,zou2023transparency}.
%In particular, \caa\ \citep{caa} derives such directions from contrastive activation means.
%\gradiend\ \citep{drechsel2026gradiend}, in contrast, uses a one-dimensional encoder--decoder architecture to learn a targeted feature from contrastive parameter gradients and can apply the learned decoder direction as a persistent parameter update.
Existing targeted methods differ along several dimensions.
\caa\ \citep{caa} constructs feature directions from contrastive activation means, whereas \gradiend\ \citep{drechsel2026gradiend} learns targeted features from contrastive parameter gradients.
These representative approaches therefore differ simultaneously in the model \emph{signal} from which the feature is constructed (e.g., activations or gradients), the \emph{estimator} used to extract it (e.g., a contrastive mean or a learned model), and the \emph{space} in which intervention is performed (e.g., activation or parameter space).
A direct comparison cannot determine which of these choices accounts for differences in feature quality.

We disentangle these factors through a controlled $3\times2$ comparison (Figure~\ref{fig:overview}).
We consider \emph{activation values}, \emph{activation gradients}, and \emph{parameter gradients} as signals, and pair each with either a \emph{contrastive mean} estimator or the one-dimensional
\emph{\acrfull{iend}} estimator generalized from \gradiend.
We introduce activation gradients as a bridge between existing approaches by incorporating gradient information while retaining activation space intervention, allowing us to separate the effect of gradient information from parameter space intervention.
%new signal for targeted feature learning, providing a bridge between the two established settings: they incorporate gradient information while still permitting activation-space intervention.
%Activation gradients provide an important bridge between the two established settings: they incorporate gradient information while still permitting activation-space intervention.
%This allows us to test whether the benefits of gradient-based feature construction stem from gradient information itself or specifically from parameter space intervention.

AxBench \citep{wu2025axbench} provides the closest existing benchmark of
feature representations through detection and steering, but its construction is not directly transferable to our gradient-based methods:
it aggregates activations across the entire input sequence, while our targeted methods construct features at the feature-inducing target position.
AxBench further relies on LLMs for feature data generation and steering evaluation.
In contrast, our benchmark avoids this LLM dependence and further evaluates representations throughout model depth rather than at only two fixed layers, an important distinction given the substantial layerwise variation we observe (Appendix~\ref{app:layer-selection}).

To evaluate this target-localized setting broadly, we compare all six targeted
approaches against \glspl{sae} as an untargeted reference across 15 diverse
tasks (\taskGender, \taskMIBIOI, \taskLangID, \dots) and \rev{ten} open-weight
language models.
Following AxBench, we evaluate two complementary dimensions:
\emph{detection}, measuring how well the learned feature distinguishes the
specified concept, and \emph{intervention}, measuring how effectively
manipulating the feature causally changes model behavior.

Across models and tasks, we find a recurring trade-off between detection and intervention (Figure~\ref{fig:detection-intervention-mean}):
%Activation-based methods highlighted by \caa\ have superior detection performance, whereas the highest intervention performance is consistently achieved by gradient-based methods, highlighted by \gradiend.
activation value methods highlighted by \caa\ achieve the strongest detection, whereas gradient-based methods highlighted by \rev{\cga} achieve the strongest intervention.
The intermediate methods help disentangle the effects of signal and estimator.
Replacing the contrastive mean with \iend\ consistently reduces detection \rev{without a consistent intervention gain}.
%and, for gradient signals, increases intervention.
Within a fixed estimator family, \rev{replacing} activation values \rev{with} parameter gradients increases intervention, \rev{along with} a loss in detection.
\rev{For the contrastive mean, activation gradients already capture most of this intervention gain. Thus, much of the gradient advantage does not require parameter space interventions or the higher dimensionality of parameter space.}
%Overall, feature quality depends jointly on signal and estimator. %, and representations best suited to identify a feature need not be those best suited to manipulating it.

\paragraph{Contributions.}
We systematically study targeted feature learning as a framework for hypothesis-driven feature analysis and provide a controlled comparison of %activation values, activation gradients, and parameter gradients under contrastive and learned estimators.
model signal and feature estimator, introducing \actiend, \agiend, \caga, and \cga\ to complete the resulting 
$3\times2$ design.
Across 15 tasks, \rev{ten} language models, and pretrained \glspl{sae} as an untargeted reference, we characterize how these choices differently affect detection and causal intervention.

%the six targeted methods  against pretrained
%\sae s across 15 tasks and \rev{ten} language models.
%Our study identifies a recurring detection-intervention trade-off and disentangles the roles of model signal, feature estimator, and intervention mechanism.

\section{Related work}

\paragraph{Targeted concept representations and activation steering.}
A long line of work studies researcher-specified concepts through
linear directions in model activations.
Concept Activation Vectors \citep{kim2018interpretabilityfeatureattributionquantitative} learn supervised concept directions, while
representation engineering methods \citep{zou2023transparency} and Activation Addition \citep{actadd} use activation directions for analysis and steering.
\gls{caa} \citep{caa} constructs directions from paired contrastive examples, and closely related mean-difference directions have been used for targeted concept detection \citep{marks2024geometry,wu2025axbench}.
More generally, probing establishes decodability without necessarily demonstrating a coherent or causally used feature \citep{hewitt-liang-2019-designing,belinkov-2022-probing}.
\cite{pmlr-v235-park24c} connects linear representations defined by
counterfactual pairs to both probing and steering, closely matching our use of controlled factual-alternative pairs. 
%Our work extends this perspective by systematically varying the signal space and estimator used to construct the feature.
%Beyond directly constructed directions, ReFT \citep{wu2024reft} optimizes representation interventions from supervision, while related work studies the geometry and optimization of activation steering \citep{pmlr-v235-singh24d,im2026unifiedunderstandingevaluationsteering,braun2025understanding}.
Other work optimizes supervised representation interventions or studies activation-steering geometry \citep{wu2024reft,pmlr-v235-singh24d,im2026unifiedunderstandingevaluationsteering,braun2025understanding}.
Most closely, AxBench \citep{wu2025axbench} evaluates both concept detection and steering across representation methods, including mean-difference directions and SAEs, \rev{showing that detection performance does not necessarily predict steering effectiveness}.
Our study instead systematically varies the signal space and estimator used for feature construction, extending the comparison from activation values to
activation gradients and parameter gradients.

\paragraph{Sparse feature learning.}
\acrlong{sae} (\acrshort{sae}\glsunset{sae}; \citet{bricken2023monosemanticity,huben2024sparse,templeton024}) learn large activation feature dictionaries without specifying the concept of interest in advance and thus provide an untargeted reference for our setting.
Identifying a concept post hoc can be complicated by feature splitting and absorption, variation across SAE trainings, and concepts distributed across multiple latents \citep{chanin_feature_splitting_absorption,leask2025sparse,paulo2026sparse,bhalla2026sparseautoencoderscaptureconcept}, making intervention dependent on latent selection and combination \citep{chalnev2024improvingsteeringvectorstargeting,arad-etal-2025-saes,soo2025interpretable,jorgensen2026steeringllmsactuallysparse}.
%This can be complicated by feature splitting and absorption \citep{chanin_feature_splitting_absorption}, variation in the learned feature dictionary across SAE trainings \citep{leask2025sparse,paulo2026sparse}, and concepts that are distributed across multiple latents \citep{bhalla2026sparseautoencoderscaptureconcept}.
%Accordingly, SAE-based intervention depends on how relevant features are selected or combined \citep{chalnev2024improvingsteeringvectorstargeting,arad-etal-2025-saes,soo2025interpretable,jorgensen2026steeringllmsactuallysparse}.
Recent work brings concept supervision into SAE training, moving closer to the targeted setting considered here.
Sparse Conditioned Autoencoders (SCAR; \cite{harle2024scar}) supervises a designated latent, while Guided SAEs \citep{harle2025measuring} encourage specified concepts to localize in the sparse representation.
Both retain a larger SAE dictionary rather than constructing only the specified feature. 

\paragraph{Gradient-based feature construction and intervention.}
While \glspl{sae} learn dictionaries of activation features, \gradiend\
\citep{drechsel2026gradiend} uses a one-dimensional encoder-decoder to learn a single targeted feature from factual-alternative parameter gradients, and uses its decoder direction for persistent parameter intervention.
Parameter space directions are also used in task arithmetic \citep{ilharco2023editing}, while activation gradients have previously supported inference-time control %in PPLM, K-Steering, and COLD-Steer 
\citep{dathathri2020plug,oozeer-etal-2025-beyond,sharma2026coldsteer}.
%More broadly, additive parameter space directions have been used to modify model behavior, for example through task arithmetic based on fine-tuning differences \citep{ilharco2023editing}.
%Prior work has also used gradients with respect to hidden activations directly for inference-time control, including PPLM \citep{dathathri2020plug}, K-Steering \citep{oozeer-etal-2025-beyond}, and COLD-Steer \citep{sharma2026coldsteer}.
In contrast, our activation gradient methods treat such gradients as the signal from which a reusable targeted feature is learned across factual-alternative contrasts.
Notably, Gradient \glspl{sae} \citep{shu-etal-2025-beyond}, despite the name, learn features from activations and use output gradients only to rank their estimated influence.
%Notably, Gradient \glspl{sae} \citep{shu-etal-2025-beyond} are \emph{not} a gradient-based \sae\ counterpart: their SAE features are learned from activations, and output gradients are subsequently used to rank the learned latents by their estimated influence.

\section{Targeted feature learning}\label{sec:method}

\subsection{From feature discovery to targeted feature learning}

We study %a setting in which a feature of interest is specified in advance and ask whether the model supports coherent internal features corresponding to it.
one-dimensional \emph{targeted features} for concepts specified before feature construction.
%We study this question through one-dimensional targeted features, each characterized by a scalar feature score and a model-internal intervention vector, adopting one-dimensionality as an operational abstraction consistent with prior concept vector and representation steering approaches \citep{kim2018interpretabilityfeatureattributionquantitative,zou2023transparency}.
Each feature comprises a scalar score for \emph{detection} and an associated model-internal direction for \emph{intervention}, following the one-dimensional abstraction common to concept-vector and representation-steering methods \citep{kim2018interpretabilityfeatureattributionquantitative,zou2023transparency}.
\emph{Detection} asks whether the score distinguishes the specified
feature realizations, whereas \emph{intervention} asks whether applying the intervention vector causally changes model behavior in the corresponding feature direction
%Evaluating internal representations through both detection and intervention follows prior work
\citep{wu2025axbench}.

We call feature learning \emph{targeted} when the feature specification is used during feature construction, rather than only to identify or interpret a representation post hoc.
\glspl{sae} 
\citep{bricken2023monosemanticity,huben2024sparse,templeton024} 
provide an untargeted reference,  since their features are learned independently of the target concept, after which examples expressing that feature can be used to identify the corresponding latent.
%Because the feature of interest is specified in advance in targeted feature learning, there is no need to learn a large dictionary of candidate features. 
%This motivates the one-dimensional \iend\ architecture used here, which learns the specified feature through a single scalar feature neuron and an associated decoder direction.
Targeted methods instead use these examples during construction.
Here, supervision consists of matched factual and alternative targets that instantiate different feature
realizations while sharing the same context. 
%By controlling which aspect differs between the two targets, these pairs define the feature to be learned. We formalize these feature specifications in Section~\ref{sec:feature-contrasts}.
Section~\ref{sec:feature-contrasts} formalizes these contrasts.

Given the same feature specification, targeted methods can differ in \emph{what} model-derived information they use and \emph{how} they construct a
one-dimensional feature from it. 
\acrlong{caa}\glsunset{caa} (\acrshort{caa}; \citet{caa})  and \gradiend\
\citep{drechsel2026gradiend} provide two useful anchors for these choices  because both construct a single concept-specific feature from contrastive supervision, but make fundamentally different choices for both signal and estimator.
\caa\ constructs a contrastive mean direction from activation values, whereas
\gradiend\ learns a one-dimensional encoder-decoder with a single scalar feature neuron from parameter gradient signals. 
A direct comparison between them therefore changes both the \emph{signal} from which the feature is learned and the \emph{estimator} used to learn it.
We disentangle them by crossing activation
values, activation gradients, and parameter gradients with either a contrastive mean estimator or the \gradiend-generalized one-dimensional \acrfull{iend}.
This yields the six targeted methods summarized in
%Table~\ref{tab:method-grid}: \caa\ and \actiend\ for activation values, \caga\ and \agiend\ for activation gradients, and \cga\ and \gradiend\ for parameter gradients.
Table~\ref{tab:method-grid}.

\begin{table}[t]
    \centering
    \begin{threeparttable}
    \caption{
        Targeted feature learning methods across three
        model-derived \emph{signals} and two \emph{estimators}.
    }
    \label{tab:method-grid}
    \fontsize{6.8}{7.7}\selectfont
    \setlength{\tabcolsep}{2pt}
    \begin{tabular}{@{}lll@{}}
        \toprule
        \textbf{Signal} &
        \textbf{Contrastive mean} &
        \textbf{\acrfull{iend}} \\
        \midrule
        Activation value &
        \acrfull{caa} \citep{caa} &
        \acrfull{actiend}\tnote{\textdagger}\, \textit{(new)} \\
        Activation gradient &
        \acrfull{caga} \textit{(new)} &
        \acrfull{agiend} \textit{(new)} \\
        Parameter gradient &
        \acrfull{cga} \textit{(new)} &
        \acrfull{gradiend}\tnote{*} \citep{drechsel2026gradiend} \\
        \bottomrule
    \end{tabular}

    \begin{tablenotes}[flushleft]
        \scriptsize
        \item[\textdagger] Pretrained \glspl{sae} provide an untargeted
        activation value reference most directly comparable to \actiend.
        \item[*] GRADIEND originally stood for
        \emph{GRADIent ENcoder Decoder}. With IEND as the generalized
        estimator family, we use \emph{GRADient IEND}.
    \end{tablenotes}
    \end{threeparttable}
\end{table}

\subsection{Specifying features through contrasts}
\label{sec:feature-contrasts}

We specify a feature through different realizations of that feature.
For a feature with $K$ classes, let $ \mathcal C \coloneqq \{c_1,\ldots,c_K\}$ and $\mathcal D_k$ denote natural examples whose factual target realizes $c_k$. 
For each example $x_i$, we denote this factual target by $y_i^F$ and construct an alternative target $y_i^A$ for the same input such that the feature realization changes while the surrounding context is held fixed.
We refer to the resulting factual-alternative comparison $(y_i^F,y_i^A)$ as a \emph{contrast}.
By controlling what differs within this contrast, we specify the feature to be learned.
\rev{For example, a gender contrast may pair the factual target ``she'' with the alternative ``he'' in the same context.}
We consider two contrast constructions:
\begin{equation}
\begin{array}{lll}
\textbf{Pairwise }(c_k,c_m):
&
x_i\in\mathcal D_k\cup\mathcal D_m,
&
y_i^A\text{ realizes the other of }c_k,c_m,
\\[1.5mm]
\textbf{One-sided }(c_k):
&
x_i\in\mathcal D_k,
&
y_i^A\text{ realizes a non-target alternative.}
\end{array}
\label{eq:contrast-construction}
\end{equation}
Pairwise learning is symmetric: factual examples are drawn from both  classes, with the factual and alternative roles reversed across them.
It yields a polar axis between $c_k$ and $c_m$, oriented from $c_k$ (lower scores) to $c_m$ (higher scores), with intermediate scores indicating weaker alignment with either class.
This construction is used by \caa\ and \gradiend. %\citep{caa,drechsel2026gradiend}.
%For $x_i\in\mathcal D_k$, the factual target realizes $c_k$ and the alternative realizes $c_m$; for $x_i\in\mathcal D_\ell$, these roles are reversed. 
%The resulting feature represents a polar axis between the two classes. 
%Since the sign of a one-dimensional feature is arbitrary, we orient the axis such that lower feature scores correspond to $c_k$ and higher scores to $c_m$.
%Scores between the two poles indicate weaker relative alignment with either class, without implying a separate feature class. 
%Pairwise contrasts of this form are used by both \caa\ and  \gradiend\ \citep{caa,drechsel2026gradiend,drechsel-etal-2026-understanding}.

One-sided learning instead uses natural examples only from the target class
$c_k$ and learns its presence relative to valid non-target alternatives.
We orient the feature such that higher scores indicate stronger expression of $c_k$.
For categorical features, alternatives are drawn from
$\mathcal C\setminus\{c_k\}$.
This also covers binary concepts for which natural data are available only for the positive class: the alternative represents absence of the feature.

For $K=2$, pairwise learning yields one symmetric polar feature, while one-sided learning yields two independently learned target-presence features.
For $K\ge2$, the constructions yield $\binom{K}{2}$ pairwise and $K$ one-sided features, respectively.
Thus, pairwise learning targets particular class-to-class distinctions, while one-sided learning provides a more compact class-specific representation \rev{that scale linearly with $K$}.

Let $\phi$ map an input-target pair to the model-derived signal used for feature learning. 
Each factual-alternative contrast induces $F_i=\phi(x_i,y_i^F)$, $A_i=\phi(x_i,y_i^A)$, and $D_i=F_i-A_i$, where $F_i$ and $A_i$ are the signals induced by the factual and alternative targets, respectively.
%We next instantiate $\phi$ in three model-derived signal spaces.
%The resulting factual, alternative, and difference signals are subsequently used by either of the two feature estimators.
We next instantiate $\phi$ in three signal spaces and use these signals with either feature estimator.

%The construction of encoder--decoder training pairs depends on the contrast type. In pairwise learning, the factual class alternates across natural examples, so the factual--alternative differences naturally occur in both directions of the polar axis. In one-sided learning, the factual target always realizes the target feature, and $D_i$ therefore has a fixed target--non-target orientation. To expose both feature poles to the encoder, we use both realizations as sources: $F_i$ is paired with $D_i$, while $A_i$ is paired with $-D_i$.

\subsection{Three signal spaces}
\label{sec:signal-spaces}

We instantiate each contrast in three model signal spaces.
Let $f_\theta$ denote the model, $\Omega$ the selected internal components (e.g., residual stream activations or transformer block parameters), $h_\Omega$ their activations, and $\theta_\Omega$ the corresponding
parameters. We define
\begin{equation}
\phi^{h}(x,y)\coloneqq h_\Omega(x,y),\quad
\phi^{\nabla_h}(x,y)\coloneqq \nabla_{h_\Omega}\mathcal L(f_\theta(x),y),\quad
\phi^{\nabla_\theta}(x,y)\coloneqq\nabla_{\theta_\Omega}\mathcal L(f_\theta(x),y).
\label{eq:signal-spaces}
\end{equation}
Activation values provide the internal representation itself, activation gradients characterize how the objective locally changes with that representation, and parameter gradients how it locally changes with the selected parameters.
Model scopes, readouts, and objectives are specified in Section~\ref{sec:experimental-setup}.

The signal spaces also differ in when target information becomes available.
For gradient-based signals, the candidate target enters  through the objective, so gradients can be extracted from the model state preceding the target.
For activation values, factual and alternative targets induce the same pre-target activation. 
Targeted learning methods based on activation values  therefore require the candidate target to be filled in before extracting the learning signal (i.e., ``he'' and ``she'' must be part of the input in Figure~\ref{fig:overview}).
Intervention, however, must act before the corresponding prediction is made.
Pretrained \glspl{sae} are not subject to this restriction because their features are learned independently of the factual-alternative contrast.
Appendix~\ref{app:sae-variants} compares pre-target and filled-target \rev{\gls{sae} features: pre-target selection lowers detection and raises intervention for most models, but remains far below the gradient-based methods.}
%activations and finds no consistent intervention advantage for pre-target activations. 

\subsection{Two targeted feature estimators}
\label{sec:feature-estimators}

%Given the factual-alternative contrasts defined in Section~\ref{sec:feature-contrasts},  we consider two estimators for constructing a targeted feature: a contrastive mean and a learned one-dimensional encoder-decoder.
%Both operate on the same signals but use the factual-alternative differences differently.
%The contrastive mean estimator directly averages the oriented feature transitions, while \iend\ learns to reconstruct these transitions from the corresponding source signals through a single scalar bottleneck.

We compare the two estimator families underlying the established targeted methods \caa\ and \gradiend: a contrastive mean and a learned one-dimensional encoder-decoder.
Both yield scalar features for detection and intervention across all three signal spaces, while differing in whether the feature is
estimated directly from contrastive differences or learned \rev{using a transition prediction} %with a reconstruction 
objective.

\paragraph{Contrastive mean.} 

\rev{
We define the contrastive mean as the average of the
feature transitions $D_i=F_i-A_i$, each oriented toward the positive feature pole:
\begin{equation}
    d_{\mathrm{mean}} \coloneqq
    \frac{1}{N}\sum_{i=1}^{N}o_iD_i,
\end{equation}
where $o_i\in\{-1,+1\}$ ensures consistent orientation.
For pairwise features $(c_k,c_m)$, we define $c_k$ as the negative and $c_m$ as the positive pole, setting $o_i=+1$ when the factual target realizes $c_m$ and $o_i=-1$ otherwise.
For one-sided features, the target class is positive
and $o_i=+1$ throughout.} Under a balanced pairwise construction, this is equivalent to a difference-of-means estimator.
Applied to activation values, activation gradients, and parameter gradients, it yields \caa\ \citep{caa}, \caga, and \cga, respectively.

\paragraph{\gls{iend}.}
We generalize the encoder-decoder construction introduced by \gradiend\ \citep{drechsel2026gradiend} into an estimator that can be applied across different signal spaces, which we call \gls{iend}.
Rather than averaging the transitions into a single vector, \iend\ learns to \rev{construct} them from corresponding source signals through a single scalar bottleneck. 
\rev{Structurally, \iend\ resembles an individual \sae\ feature \citep{bricken2023monosemanticity,huben2024sparse}}: both associate a scalar encoder value with a decoder vector. 
\rev{However, their learning objectives differ:}
an \sae\ learns a large dictionary through untargeted activation reconstruction, \rev{whereas \iend\ learns a single scalar feature from predefined transitions.}
%It therefore acts as a targeted analogue of learning an individual SAE-like feature rather than a dictionary from which the desired feature must later be identified. 

Let $(S,T)$ denote the signal provided to the encoder and the \rev{corresponding target transition for} the decoder.
In pairwise learning, examples from both classes already expose both feature poles as sources: across contrasts, $A_i$ spans both classes and is paired with $D_i$.
In one-sided learning, factual signals all belong to the target side, while alternative signals represent the non-target side. Using only $(A_i,D_i)$ would therefore expose the encoder only to non-target sources.
We instead use both realizations as sources: $(A_i,D_i)$ and $(F_i,-D_i)$. 
Let $\mathcal P$ denote the resulting set of source-target pairs.
%Let $\mathcal P\coloneqq\bigcup_{i=1}^{N}\mathcal P_i$ denote the resulting source-target pairs $(S,T)$.
The encoder maps each source signal to a scalar feature value, \rev{from which} the decoder
constructs the corresponding transition:
\begin{equation}
    z=\operatorname{enc}(S)\coloneqq\tanh(w_e^\top S+b_e),
    \qquad
    \operatorname{dec}(z)\coloneqq b_d+w_d z.
    \label{eq:iend-encoder-decoder}
\end{equation}
Encoder and decoder are jointly optimized to minimize
$\mathcal L=\frac{1}{|\mathcal P|}\sum_{(S,T)\in\mathcal P}
\lVert T-\operatorname{dec}(\operatorname{enc}(S))\rVert_2^2$.
Applied to activation values, activation gradients, and parameter gradients,
this yields \actiend, \agiend, and \gradiend, respectively.

\subsection{Feature detection and feature intervention}
\label{sec:detection-intervention}

We use each learned feature for both \emph{detection} and \emph{intervention}.
For contrastive mean methods, the feature score is %$s_{\mathrm{mean}}(S)=d_{\mathrm{mean}}^\top S$. 
$s_{\mathrm{mean}}(S)
\coloneqq
\frac{d_{\mathrm{mean}}^\top S}{\|d_{\mathrm{mean}}\|_2\,\|S\|_2}$.
For \iend, it is the encoder output $s_{\mathrm{IEND}}(S)\coloneqq\operatorname{enc}(S)$.
For intervention, we orient the learned transition toward a selected class $c_i$.
For contrastive mean methods, this means orienting $d_{\mathrm{mean}}$ toward $c_i$. 
\rev{For \iend, we use $\operatorname{dec}(z_{c_i})$, where $z_{c_i}\in\{-1,+1\}$ denotes the feature pole corresponding to $c_i$.}
%Gradient-derived directions are sign-reversed because a loss gradient points toward increasing the loss, whereas intervention toward $c_i$ requires decreasing it.
Let $d_{c_i}$ denote the resulting class-oriented intervention direction.
We intervene with
$h_\Omega \leftarrow h_\Omega+\alpha d_{c_i}$ in activation space and
$\theta_\Omega \leftarrow \theta_\Omega+\alpha d_{c_i}$ in parameter space, with intervention strength $\alpha\in\mathbb R$.
Positive $\alpha$ moves toward $c_i$, while negative $\alpha$ moves in the
opposite direction.

\paragraph{Sparse autoencoder reference.}
For \sae, we identify the requested feature post hoc by ranking latents by their ability to distinguish it.
The selected latent activation provides the detection score and its decoder vector the activation-space intervention direction.
We treat each selected latent as a one-sided feature, steering toward or away from its decoder direction.

\section{Experimental Setup}
\label{sec:experimental-setup}

%\paragraph{Models and tasks.} 
%We evaluate \gpttwo\ \citep{radford2019language}, \pythia\ \citep{pythia}, %\gemma\ \citep{gemmateam2024gemma2improvingopen}, 
%and \llama\ \citep{grattafiori2024llama} on 15 tasks spanning semantic, grammatical, factual, language, and algorithmic features.
\paragraph{Models and tasks.} We evaluate \rev{ten pretrained language models ranging from 70M to 9B parameters:
\pythia\ \citep{pythia}, \gpttwo\ \citep{radford2019language}, \gemma\ \citep{gemmateam2024gemma2improvingopen},
\gemmaS, \gemmaM\ \citep{gemmateamGemma3Technical2025}, 
\qwenS\ and \qwenL\ \citep{qwen35}, \mistral\ \citep{jiang_mistral_2023},
\llama\ \citep{grattafiori2024llama}, and \deepseek\ \citep{deepseekai2025deepseekr1incentivizingreasoningcapability}.}
\rev{Each model is evaluated} on 15 tasks spanning semantic, grammatical, factual, language, and algorithmic features.
The benchmark includes both pairwise and one-sided feature specifications (Section~\ref{sec:feature-contrasts}).
For the \sae\ baseline, we use pretrained \glspl{sae} from SAELens \citep{saelens}.
The models were selected based on the availability of pretrained SAEs and to cover pretraining from different model providers at different model sizes.
Table~\ref{tab:task-summary} summarizes the tasks. Details including dataset constructions are reported in Appendix~\ref{app:data}.

\begin{table}[t]
    \centering
    \scriptsize
    \caption{
        Summary of the 15 feature learning tasks.
        $pw$ denotes pairwise and $1s$ one-sided tasks.
    }
    \label{tab:task-summary}
    \setlength{\tabcolsep}{3pt}

    \begin{tabularx}{\linewidth}{
        @{}
        l
        c
        >{\raggedright\arraybackslash}X
        @{}
    }
        \toprule
        \textbf{Family} & \textbf{Eval.} & \textbf{Tasks (classes)} \\
        \midrule

        Semantic
        & $pw$/$1s$
        & \taskGender\ (male/female);
          \taskEmotion\ (positive/negative);
          \taskRace\ (Asian/Black/White);
          \taskReligion\ (Christian/Muslim/Jewish)
        \\

        Factual
        & $pw$/$1s$
        & \taskRavelCont\ (Asia/Europe/Africa);
          \taskRavelCountry\ (United States/China/Russia);
          \taskRavelLang\ (English/Portuguese/Spanish)
        \\

        Language
        & $pw$/$1s$
        & \taskLangID\ (English/French/German)
        \\

        Grammatical
        & $pw$/$1s$
        & \taskPronNum\ (singular/plural);
          \taskPronPers\ (first/second/third)
        \\

        Algorithmic
        & $1s$
        & \taskMIBIOI\ (IO/subject);
          \taskKeyValue\ (value/other value);
          \taskInduction\ (match/distractor);
          \taskRepetition\ (yes/no);
          \taskFuncComp\ (result/distractor)
        \\

        \bottomrule
    \end{tabularx}
    \vspace{-3pt}
\end{table}

\paragraph{Protocol and model representations.} 
All methods use the same train-validation-test splits.
Training data constructs the features, validation data selects method-specific choices, and test data is used only for final evaluation. 
Validation selection includes \iend\ checkpoints, \sae\ latents, detection thresholds, and intervention strengths.
Activation value methods use residual stream activations after each selected transformer block, averaged over the positions corresponding to the target tokens.
Gradient methods condition on the first target token: activation gradients are taken with respect to the residual stream at its prediction position and parameter gradients with respect to transformer block parameters.
We exclude embeddings and the final output head. Details are in Appendix~\ref{app:experimental-details}.

%For the \sae\ reference, features are selected using validation data only.
%We report both the highest-ranked individual latent ($k=1$) and a validation-selected set of latents ($k^\star$), allowing the untargeted dictionary to use either a single feature or a feature set to represent the targeted concept.

\paragraph{Detection evaluation.}
Detection measures how well a feature score separates its target class from neutral examples and, where applicable, other classes of the same feature, which we call \emph{rivals}.
For target class $c_k$, let $\mathrm{AUC}_n$ denote target-versus-neutral AUROC and $\mathrm{AUC}_o$ the lowest target-versus-rival AUROC.
At a validation-selected threshold, $\mathrm{Spec}_n$ is specificity on neutral examples and $\mathrm{Excl}$ the lowest specificity across rivals.
We define
\begin{equation}
\mathrm{Det.}=
\begin{cases}
\min\{\mathrm{AUC}_n,\mathrm{AUC}_o,\mathrm{Spec}_n,\mathrm{Excl}\}
& \text{with rivals}\\
\min\{\mathrm{AUC}_n,\mathrm{Spec}_n\}
& \text{without rivals}.
\end{cases}
\end{equation}
Taking the minimum requires a feature to perform well on all applicable detection criteria rather than allowing strong performance on one criterion to compensate for failure on another while still yielding a single scalar score.

\paragraph{Intervention evaluation.}
Intervention measures how manipulating the feature changes the model's probability of the target behavior.
For a feature oriented toward target class $t$, let $I^+$ denote a strengthening intervention toward $t$, $I^-$ a weakening intervention in the opposite direction, and $I_0$ the unmodified model.
Let $\bar P_I(t)$ denote the mean next-token probability assigned to $t$ across
the corresponding held-out examples under intervention $I$.
For independently selected strengthening and weakening interventions,
\begin{equation}
    \Delta P^+\coloneqq\bar P_{I^+}(t)-\bar P_{I_0}(t),
\qquad
\Delta P^-\coloneqq\bar P_{I_0}(t)-\bar P_{I^-}(t),
\qquad
\mathrm{Int.}\coloneqq (\Delta P^++\Delta P^-)/2.
\end{equation}
For pairwise features, strengthening is evaluated on rival examples and weakening on target examples. For one-sided features, both use target examples.
To prevent large effects caused by broad model degradation, strengthening and weakening magnitudes are selected independently on validation data subject to a language modeling constraint~\citep{meng2022locating}.
With $\mathrm{LMS}\coloneqq 1/(1+\overline{\mathrm{NLL}})$, where $\overline{\mathrm{NLL}}$ is the mean token-level negative log-likelihood, we require $\mathrm{LMS}(I)\geq0.99\,\mathrm{LMS}(I_0)$ and independently select strengthening and weakening interventions by grid search over approximately log-spaced magnitudes, maximizing the validation effect.
%Further details are in Appendix~\ref{app:intervention-evaluation}.

\paragraph{Representation selection.}
\caa\ and pretrained \glspl{sae} are conventionally evaluated as layer-local
activation features \citep{lawson2025residual,lieberum-etal-2024-gemma}, and their performance can depend strongly on layer choice \citep{caa,multilingualsteering}.
An individual-layer scope uses the post-block residual stream representation after transformer block $\ell$ for activation-based methods, and the corresponding block parameters for parameter-gradient methods.
For \sae\ and contrastive mean methods, we evaluate all single layer scopes and an all-layer scope, the latter enabling direct comparison with \iend, which learns jointly across selected components.
Validation detection selects among these candidates, and the selected scope is fixed for both test detection and intervention.
For \iend, evaluating every layer separately would require training a distinct encoder-decoder for each scope, so we use fixed joint scopes: all post-block residual stream activations for \actiend\ and \agiend, and the joint parameter gradient scope for \gradiend.

\paragraph{Aggregation.}
For each method, model, and task, detection and intervention scores are first computed for each evaluated class or contrast member and then averaged.
Task scores are further averaged within each model, and cross-model summaries average the resulting model means.

\section{Feature detection and intervention performance}
\label{sec:main-results}

\begin{figure}[!t]
    \centering
    \includegraphics[width=\linewidth]{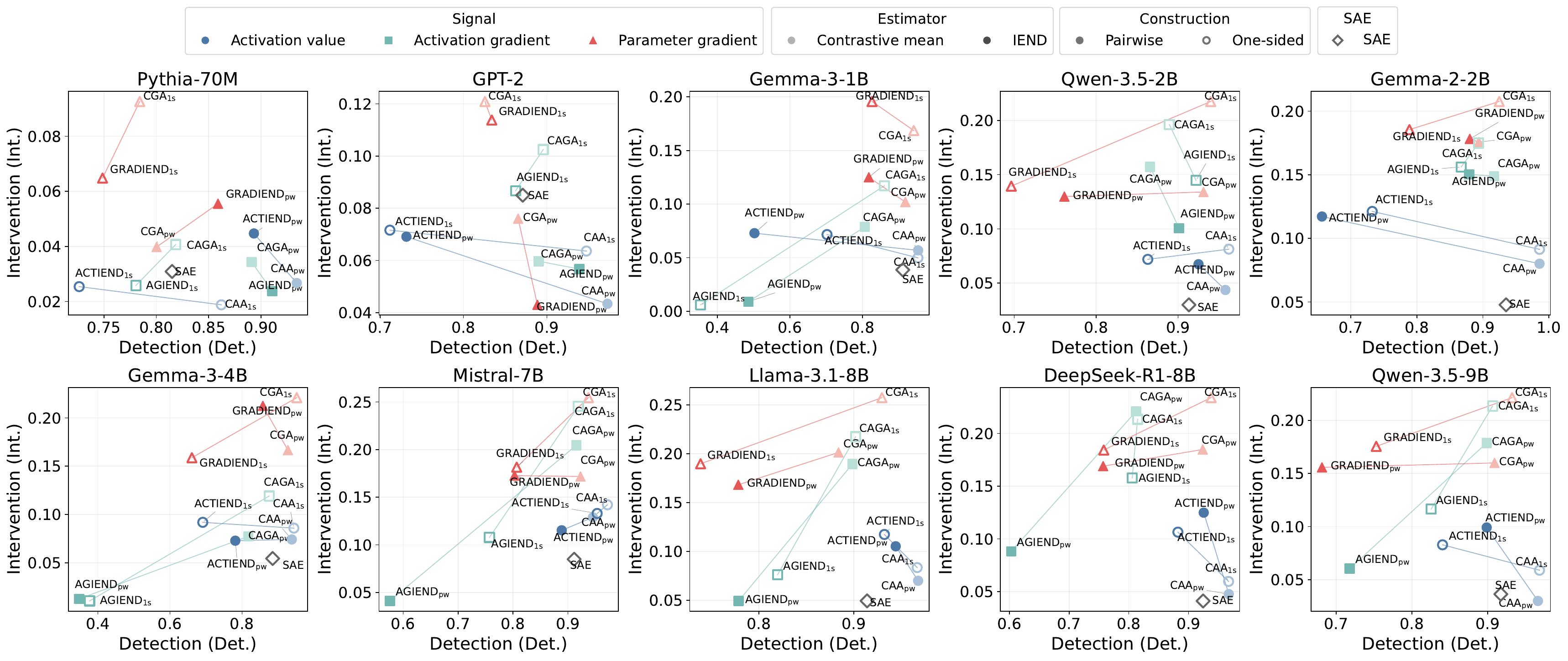}
\caption{Task-averaged detection and intervention for each method.
Lines connect the two estimators using the same signal and construction.
The model-averaged aggregate is shown in Figure~\ref{fig:detection-intervention-mean}.}
    \label{fig:detection-vs-intervention}
\end{figure}

Figures~\ref{fig:detection-intervention-mean} and
\ref{fig:detection-vs-intervention} show the model-averaged and corresponding per-model results.
Activation value methods, particularly \caa, achieve the strongest detection, while gradient-based methods, particularly \rev{\cga}, achieve the strongest intervention overall.
%With the exception of \gemma, methods that move toward stronger detection tend to exhibit weaker intervention and vice versa.
Thus, detection and intervention capture complementary properties. %, while the best signal for intervention can depend on model architecture.
The \sae\ reference \rev{is outperformed by \caa\ on detection and has the lowest mean intervention of all methods.}
Detailed results \rev{and ablations} 
are in Appendix~\ref{app:detailed-results}.

\paragraph{Signal and estimator.}
Within the contrastive mean family, \cga\ \rev{achieves the strongest intervention, but \caga\ already captures much of this advantage: it achieves roughly twice the intervention performance of \caa\ while operating in the same activation space. This shows that the gradient advantage is not restricted to parameter space interventions, although the advantage of \caga\ largely disappears without layer selection (Appendix~\ref{app:layer-selection}).}
For gradient signals, replacing the contrastive mean with \iend\ typically \rev{does not increase intervention and reduces detection}.
For activation values, however, replacing the contrastive mean with \iend\ typically reduces detection \rev{but increases} intervention, moving \actiend\  below \caa\ on \rev{detection and above it on intervention}.
%\actiend\ and the pretrained \sae\ reference, the two learned activation value approaches, occupy a similar overall region.

\paragraph{Feature construction.}
The aggregate figure is not directly comparable along this axis because one-sided methods include additional tasks.
On matched tasks, one-sided construction generally matches pairwise performance %and often yields slightly higher detection, with similar intervention for most methods 
\rev{except for higher detection for \actiend\ and \agiend, as well as lower intervention for \gradiend} (Appendix~\ref{app:pairwise-onesided}).
One-sided construction additionally requires only $K$ rather than $\binom{K}{2}$ feature models and supports features without a natural opposing class.

\begin{figure}[!t]
    \centering

    \begin{subfigure}[t]{0.6\textwidth}
        \centering
        \includegraphics[width=\linewidth]{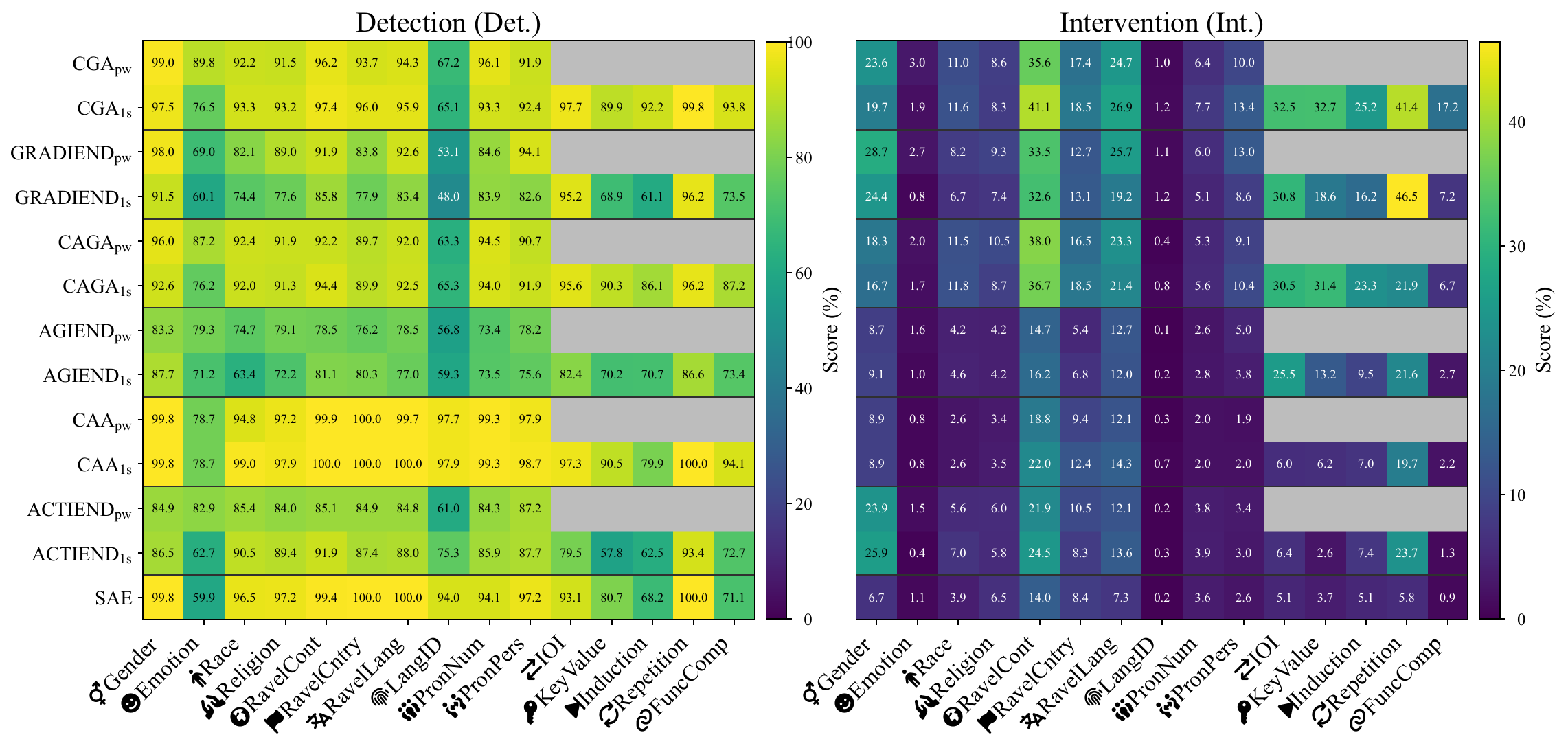}
        \caption{Model-averaged Det. and Int. scores for each task and method.}
        \label{fig:task-heatmap}
    \end{subfigure}
    \hfill
    \begin{subfigure}[t]{0.31\textwidth}
        \centering
        \includegraphics[width=\linewidth]{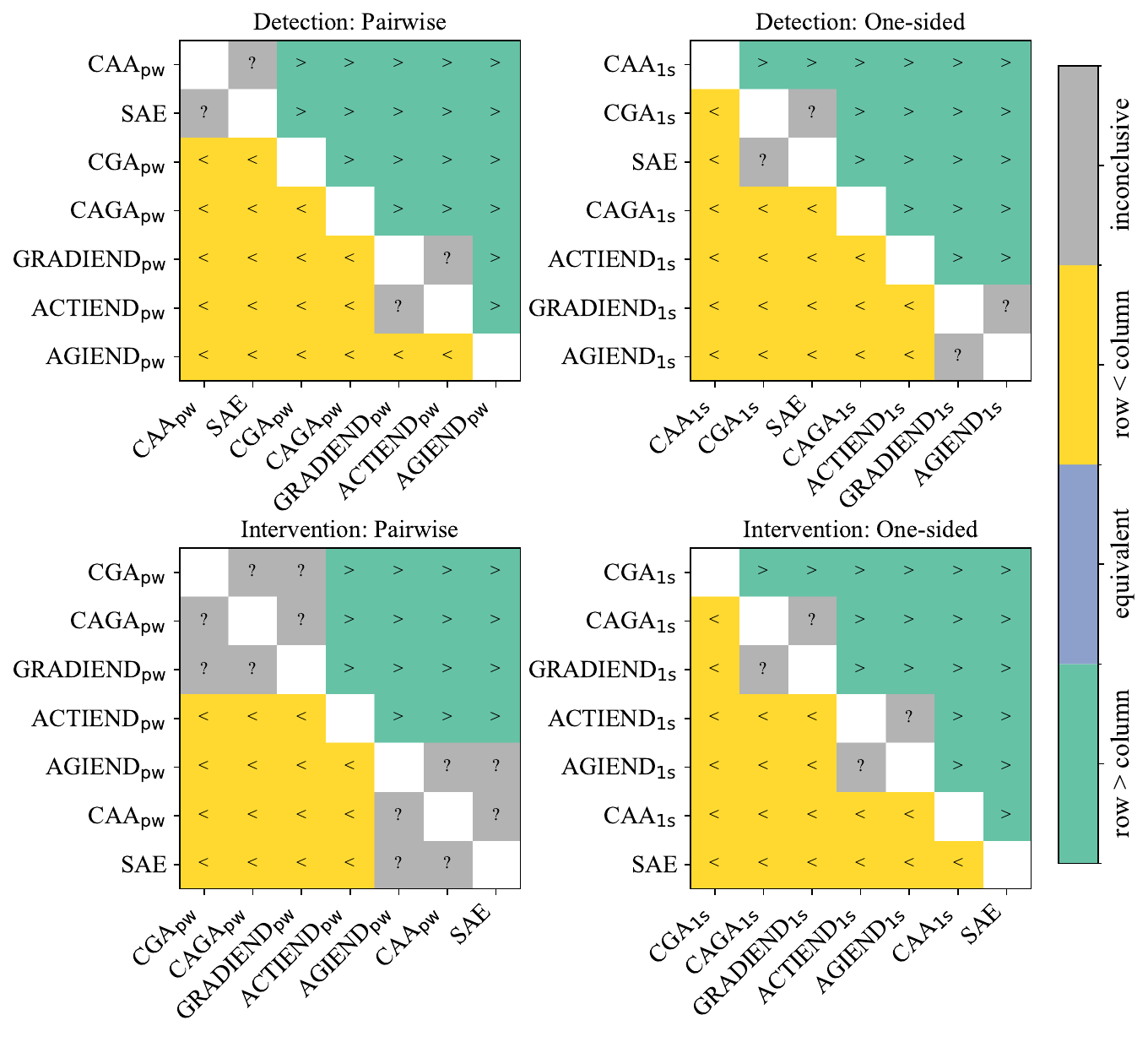}
        \caption{Bayesian method comparison.
        %Posterior decisions across matched model-task observations, shown separately for detection and intervention and for pairwise and one-sided feature construction.
        }
        \label{fig:bayesian-decision-overview}
    \end{subfigure}
%\vspace{-3pt}
    \caption{Overview of task-level performance and Bayesian method comparisons.}
  %  \vspace{-5pt}
    \label{fig:task-and-bayesian-overview}
\end{figure}

\paragraph{Task-level variation.}
Figure~\ref{fig:task-heatmap} shows substantial variation across tasks.
\taskLangID\ and \taskEmotion\ are the most difficult for both detection and intervention. For \taskEmotion, the \rev{evaluated} \glspl{sae} fail to recover a feature latent \rev{for some models}, whereas the targeted methods retain non-trivial detection \rev{(see Appendix~\ref{app:emotion-recovery})}.
%\rev{\taskLangID detection varies strongly across methods: \caa and \sae detect it well, \gradiend and \agiend do not.}
Most remaining pairwise capable tasks are detected reliably across methods, whereas several one-sided tasks are substantially harder, with some methods approaching chance-level detection (e.g., \taskKeyValue).
Intervention varies more strongly across tasks and methods than detection: strong detection does not consistently imply strong intervention. The RAVEL tasks, \taskGender, \taskRepetition\ \rev{and \taskMIBIOI} show some of the largest intervention effects, particularly for gradient-based \mbox{methods}.

\paragraph{Bayesian method comparison.}
To complement the mean scores with a comparison that is less dominated by the magnitude of isolated method-task differences, we also perform a Bayesian pairwise analysis over the matched model-task observations using
\texttt{autorank} \citep{Herbold2020}.
Figure~\ref{fig:bayesian-decision-overview} supports our main conclusion: \rev{\caa\ is} the strongest position for detection, while \rev{\cga,} \gradiend\, \rev{and \caga\ are} strongest for pairwise intervention \rev{and statistically unresolved from one another}.
Under one-sided construction, %the two remain statistically unresolved.
\rev{\cga\ is clearly strongest.}
Details are given in Appendix~\ref{app:statistics} and~\ref{app:autorank}.

\paragraph{Representation scope.}
We next compare features represented at a single transformer layer with those using the corresponding joint all-layer representation.
For \sae\ and contrastive mean methods, validation detection selects a single layer over the all-layer representation in \rev{93.1}\% of class/contrast decisions. The validation-best single layer achieves higher held-out detection than the corresponding all-layer representation in \rev{83.7}\% of comparisons (Appendix~\ref{app:layer-selection}). Intervention differences are smaller and less systematic.
Moreover, repeating the main detection-intervention comparison from Figure~\ref{fig:detection-vs-intervention} with \sae\ and contrastive mean methods restricted to their all-layer representations preserves the main signal-level pattern: activation value methods remain strongest for detection, while \rev{parameter} gradient methods remain strongest for intervention (Figure~\ref{fig:detection-intervention-all-layer}).
Thus, our key observation is not driven by the representation-scope difference of contrastive mean methods relative to \iend.

%Thus, restricting a feature to one layer appears substantially more important for obtaining a clean detection than for effective intervention, motivating approaches that decouple the representations used for detection% and manipulation.

\section{Discussion and Limitations}
\label{sec:discussion}

\paragraph{Detection versus intervention.}
Our results qualify the widespread use of activation value features for both feature identification and model steering.
\rev{While prior work has shown that detection performance does not necessarily predict intervention effectiveness \citep{marks2024geometry,wu2025axbench}, our study identifies a systematic trade-off between model signals:}
the strongest detection methods use activation values, whereas the strongest intervention methods use gradient-based signals.
This finding is especially notable given the prominence of activation-based steering approaches in prior work.
\caa\ was introduced as a targeted activation-steering method \citep{caa}, while \glspl{sae} are widely used for feature discovery and have
also been applied to causal intervention
\citep{bricken2023monosemanticity,huben2024sparse,templeton024}.
Yet in our evaluation, \caa\ is particularly strong for detection, whereas
\rev{\cga}\ and \gradiend\ are particularly strong for intervention.
\rev{Importantly, \caga}\ further shows that this advantage does not require parameter space intervention: \rev{it intervenes about twice as well as \caa, although with lower detection and only with layer selection.}

\paragraph{Positional asymmetry.}
An important consideration concerns where these signals become observable.
In our tasks, activation value contrasts are measured at the feature-related target representation: factual and alternative examples become distinguishable only once the corresponding target token (e.g., a gender pronoun for \taskGender) enters the residual stream.
Intervention, however, must influence that target before or while it is generated.
Gradient signals do not share this restriction, since the candidate target enters through the CLM objective and its gradient signal can be derived at the positions predicting it.
This positional asymmetry is a consequence of our targeted feature learning setup: when factual and alternative examples share the same prefix, their pre-target activations are identical, so no targeted activation contrast can be learned.
Untargeted \sae\ features do not have this restriction because they are learned independently of the factual-alternative contrast.
%We therefore additionally select \sae\ features from pre-target activations as a partial control for positional alignment.
%This does not consistently improve intervention (Appendix~\ref{app:sae-variants}), suggesting that positional alignment may contribute to, but does not by itself explain, the intervention advantage of gradient signals.
Nevertheless, selecting \sae\ features from pre-target activations \rev{improves intervention only moderately (mean 0.075 vs. 0.050).} (Appendix~\ref{app:sae-variants}).
Thus positional alignment may contribute to, but does not fully explain, the intervention advantage of gradient signals.

%More general activation-based constructions can aggregate information across broader token spans \citep{wu2025axbench,pai-etal-2026-billy,ICLR2026_a79237d6}, which we do not investigate here.

\paragraph{Decoupling detection and intervention.}
The separation between detection and intervention suggests that identifying and manipulating a feature need not use the same signal or representation scope.
Activation values directly reflect whether a feature is present in the model state, while gradients encode how the objective changes with respect to that state or its parameters.
Our representation-scope analysis (Figure~\ref{fig:layer-selection-paired-deltas}) points in the same direction: detection benefits strongly and consistently from localizing a feature to an informative single layer, whereas intervention is considerably less sensitive to the same localization.
%A natural extension of \iend\ is therefore to decouple these roles, for example using localized activations as encoder inputs and contrastive gradients as decoder targets, potentially with different input and output scopes.
%Such a hybrid could combine strong activation-based readout with gradient information useful for intervention.
%Whether these advantages can be combined in a single feature learner, or whether their trade-off between detection and intervention is more fundamental, remains an open empirical question.
\rev{
We explore this possibility through \textsc{HYBRIEND}
(Appendix~\ref{app:additional-baselines}), which combines activation value encoding with parameter gradient decoding.
The hybrid generally improves upon the single-signal \iend\ variants, but does not consistently outperform contrastive mean methods.
This suggests that combining different signals is promising, although further work is needed to overcome the trade-off.
}

\paragraph{\rev{Practical recommendations.}}
\rev{
Our results suggest different methods depending on the intended use.
For detecting a predefined feature, \caa\ provides a simple and effective choice, particularly with layer selection.
For activation-space steering, \caga\ generally achieves stronger interventions than \caa\ with layer selection.
For persistent model editing, \cga\ offers strong intervention performance without requiring inference-time hooks.
As no method consistently dominates both detection and intervention, the choice of method should depend on the intended application.
}

\paragraph{Limitations.}
Our evaluation covers \rev{ten} language models and 15 tasks, and generalization to
other architectures, scales, and feature types remains to be established.
The target-localized formulation enables controlled comparison across methods
but differs from broader benchmarks such as AxBench \citep{wu2025axbench},
preventing a direct matched comparison.
Finally, intervention is evaluated at the specified target token and
does not capture effects on longer generations.

\section{Conclusion}
\label{sec:conclusion}

We presented a controlled comparison of targeted feature learning methods that separates three design choices: the model signal, the feature estimator, and the feature specification.
Across \rev{ten} language models and 15 tasks, detection and intervention emerge as complementary measures of feature quality.
Contrastive activation value features provide particularly strong detection, while gradient-based methods achieve the strongest causal intervention, \rev{particularly \cga\ and \gradiend}.
Feature construction and representation scope further affect this trade-off: one-sided formulations broaden the applicability of targeted feature learning, while layer localization benefits detection considerably more consistently than intervention.
Together, these findings suggest that identifying and causally manipulating a feature need not rely on the same signal or representation scope.
%They motivate targeted feature learning methods that explicitly combine these complementary strengths, including hybrid and layer-selective IEND formulations.
\rev{Our exploratory hybrid experiment further suggests that combining activation-based detection with gradient-based intervention is promising, although no single method consistently dominates both objectives.
}

\subsection*{AI use statement}

Generative AI tools were used to discuss and refine aspects of the research
methodology and experimental design, assist with the implementation of several
methods following author-specified designs, and support interpretation of
experimental results.
Additionally, generative AI tools were used for literature discovery,
generation of initial drafts of parts of the manuscript, and editing of the
text. All AI-assisted code was reviewed and tested by the authors, suggested
references were checked against the original sources, and AI-assisted text was
reviewed and substantially revised. Generative AI tools were not used to
generate synthetic datasets.
The authors take responsibility for the final content of this work, including
all claims, results, code, and text produced with the aid of generative AI.

\subsection*{Reproducibility statement}

We release the full study code (Appendix~\ref{app:implementation}) and all redistributable task datasets (Tables~\ref{tab:task-inventory} and~\ref{tab:neutral-inventory}), including the extended \textsc{gradiend} package code, experiment configurations, and analysis scripts.
Synthetic datasets can be regenerated deterministically from fixed seeds, while
tasks based on third-party data are reconstructed from their public sources using the provided scripts.
The repository README documents installation, dataset generation, and the
commands for reproducing the main experiments.
Appendices~\ref{app:data} and \ref{app:experimental-details} provide further details on the tasks, models, signals, training, evaluation, aggregation, layer selection, and
statistical analysis. Representative compute costs for two models on one task are reported in Appendix~\ref{app:runtime}.
All training and evaluation runs use fixed seeds, with model and method selection performed on validation data and test data used only for final evaluation.

\bibliography{iclr2027_conference}
\bibliographystyle{iclr2027_conference}

\appendix

\section{Dataset Documentation}
\label{app:data}

\begin{table}[!tp]
\tiny
    \centering
    \caption{Complete task inventory. Sizes are effective labeled train/validation/test rows after task-level class mapping and before counterfactual expansion. P = both poles occur naturally and both pairwise and one-sided regimes are evaluated. O = only the first listed class is factual and the second is its constructed counterpart. We do not redistribute \taskLangID\ due to licensing constraints and instead provide a script for reconstructing it from the original sources.}
\label{tab:task-inventory}
% Labeled counts are post-adapter base rows, before one-sided CF expansion.
\begin{tabularx}{\linewidth}{@{}>{\raggedright\arraybackslash}p{1.48cm}
                       >{\raggedright\arraybackslash}p{1.75cm}
                       >{\centering\arraybackslash}p{0.1cm}
                       >{\raggedleft\arraybackslash}p{1.5cm}
                       >{\raggedright\arraybackslash}X@{}}
\toprule
Task & Feature classes & Reg. & Train/val/test & Source and Hugging Face ID \\
\midrule

\taskGender & male (M), female (F) & P & 73,860/2,110/8,440 &
%GENTER--Ajibawa name-filled, ;
\href{https://huggingface.co/datasets/aieng-lab/genter-ajibawa-name-filled}{\texttt{aieng-lab/genter-ajibawa-name-filled}} \citep{drechsel2026gradiend} subset \texttt{n1} \\

\taskEmotion & positive, negative & P & 600/200/200 &
%TweetEval contexts + NRC adjective polarity, vocabulary-held-out config split; 
\href{https://huggingface.co/datasets/aieng-lab/en-sentiment-nrc}{\texttt{aieng-lab/en-sentiment-nrc}} \citep{drechsel2026gradiendpythonpackageendtoend} vocabulary-held-out split \\

\taskRace & Asian, Black, White & P & 33,602/9,603/4,799 &
%Wikipedia-10 + bias-attribute lexicon;
\href{https://huggingface.co/datasets/aieng-lab/gradiend_race_data}{\texttt{aieng-lab/gradiend\_race\_data}} \citep{drechsel2026gradiend} \\

\taskReligion & Christian, Muslim, Jewish & P & 20,048/5,730/2,863 &
%Wikipedia-10 + bias-attribute lexicon;
\href{https://huggingface.co/datasets/aieng-lab/gradiend_religion_data}{\texttt{aieng-lab/gradiend\_religion\_data}} \citep{drechsel2026gradiend} \\

\taskRavelCont & Asia, Europe, Africa & P & 1,344/168/168 &
%Pinned RAVEL entities/templates; \path{hij/ravel}@\texttt{2c45eb23}; corrected, entity-disjoint derivative \\
Source: \href{https://huggingface.co/datasets/hij/ravel}{\texttt{hij/ravel}} \citep{huang-etal-2024-ravel}; Processed \iflink \href{https://huggingface.co/datasets/aieng-lab/gradiend-ravel-continent}{\texttt{aieng-lab/gradiend-ravel-continent}} \else \texttt{anonymous} \fi (new) \\

\taskRavelCountry & United States, China, Russia & P & 408/51/51 &
Source: \href{https://huggingface.co/datasets/hij/ravel}{\texttt{hij/ravel}} \citep{huang-etal-2024-ravel}; Processed \iflink \href{https://huggingface.co/datasets/aieng-lab/gradiend-ravel-country}{\texttt{aieng-lab/gradiend-ravel-country}} \else \texttt{anonymous} \fi (new) \\

\taskRavelLang & English, Portuguese, Spanish & P & 528/66/66 &
Source: \href{https://huggingface.co/datasets/hij/ravel}{\texttt{hij/ravel}} \citep{huang-etal-2024-ravel};  Processed \iflink \href{https://huggingface.co/datasets/aieng-lab/gradiend-ravel-language}{\texttt{aieng-lab/gradiend-ravel-language}} \else \texttt{anonymous} \fi (new)
\\
\taskLangID & English (en), French (fr), German (de) & P & 24,000/3,000/3,000 &
Source: \href{https://huggingface.co/datasets/Helsinki-NLP/opus-100}{\texttt{Helsinki-NLP/opus-100}} and \href{https://github.com/facebookresearch/MUSE}{\texttt{https://github.com/facebookresearch/MUSE}}; Processed: \iflink \url{https://github.com/aieng-lab/iend-study/blob/main/study/data/language_parallel.py} \else \texttt{anonymous} \fi (new) \\

\taskPronNum & singular, plural & P & 32,000/4,000/4,000 &
%English Wikipedia pronoun clozes;
\href{https://huggingface.co/datasets/aieng-lab/en-pronouns}{\texttt{aieng-lab/en-pronouns}} \citep{drechsel2026gradiendpythonpackageendtoend} \\

\taskPronPers & first, second, third & P & 40,000/5,000/5,000 &
\href{https://huggingface.co/datasets/aieng-lab/en-pronouns}{\texttt{aieng-lab/en-pronouns}} \citep{drechsel2026gradiendpythonpackageendtoend} \\

\taskMIBIOI & indirect object (IO), subject & O & 10,000/10,000/1,000 &
Source: \path{mib-bench/ioi} \citep{mueller2025mib}; Processed \iflink \href{https://huggingface.co/datasets/aieng-lab/gradiend-mib-ioi}{\texttt{aieng-lab/gradiend-mib-ioi}} \else \texttt{anonymous} \fi (new) \\

\taskKeyValue & value, other value & O & 8,000/1,000/1,000 &
Controlled local generator; Processed \iflink \href{https://huggingface.co/datasets/aieng-lab/gradiend-key-value}{\texttt{aieng-lab/gradiend-key-value}} \else \texttt{anonymous} \fi (new) \\

\taskInduction & match, distractor & O & 8,000/1,000/1,000 &
Controlled local generator; Processed \iflink \href{https://huggingface.co/datasets/aieng-lab/gradiend-induction}{\texttt{aieng-lab/gradiend-induction}} \else \texttt{anonymous} \fi (new) \\

\taskRepetition & yes, no & O & 8,000/1,000/1,000 &
Controlled local generator; Processed \iflink \href{https://huggingface.co/datasets/aieng-lab/gradiend-repetition}{\texttt{aieng-lab/gradiend-repetition}} \else \texttt{anonymous} \fi (new) \\

\taskFuncComp & result, distractor & O & 8,000/1,000/1,000 &
Controlled local generator; Processed \iflink \href{https://huggingface.co/datasets/aieng-lab/gradiend-function-composition}{\texttt{aieng-lab/gradiend-function-composition}} \else \texttt{anonymous} \fi (new) \\
\bottomrule
\end{tabularx}
\end{table}

\begin{table}[!tbp]
\centering
\tiny

\caption{Representative labeled examples. Long natural contexts are shortened
with an ellipsis. The masked span and local wording are retained. The right
column gives the factual target token followed by its feature class. For
one-sided tasks the counterfactual class is still listed in
Table~\ref{tab:task-inventory}, but does not add a second base row.}
\label{tab:task-examples}

\begin{tabularx}{\linewidth}{@{}>{\raggedright\arraybackslash}p{1.48cm}
                       >{\raggedright\arraybackslash}X
                       >{\raggedright\arraybackslash}p{2.90cm}@{}}

\toprule
Task & Masked context & Target / class \\
\midrule

\taskGender & Carl looked around the room, noticing tired expressions and
disheartened faces. \texttt{[MASK]} realized then that something needed to
change. & \emph{he} / M \\

\taskEmotion & I am so unbelievably \texttt{[MASK]} for tomorrow. &
\emph{excited} / positive \\

\taskRace & Lutron alleged that products were made in Mexico and
\texttt{[MASK]} and then imported into the United States. &
\emph{China} / Asian \\

\taskReligion & The inscription should, from a \texttt{[MASK]} point of view,
pinpoint where Jumne was located. & \emph{Christian} / Christian \\

\taskRavelCont & Toronto is a city in the continent of North America.
Pilibhit is a city in the continent of \texttt{[MASK]}. & \emph{Asia} / Asia \\

\taskRavelCountry & St.~Petersburg is in Russia. Johnstown is in
\texttt{[MASK]}. & \emph{United States} / United States \\

\taskRavelLang & People in Mexico City speak Spanish. People in Longreach
speak \texttt{[MASK]}. & \emph{English} / English \\

\taskLangID & \texttt{[en]} Every morning before work he quietly reads about
a famous \texttt{[MASK]} in the paper. & \emph{book} / English \\

\taskPronNum & I was in it and around it. But certainly \texttt{[MASK]}
had no corporeal existence. & \emph{I} / singular \\

\taskPronPers & Later editions may list both authors. \texttt{[MASK]} may
also omit material or add new material. & \emph{They} / third \\

\taskMIBIOI & As Carl and Maria left the consulate, Carl gave a fridge to
\texttt{[MASK]}. & \emph{Maria} / IO \\

\taskKeyValue & \texttt{key5 = white; k20 = green; key5 = [MASK]} &
\emph{white} / value \\

\taskInduction & \texttt{j pear green j [MASK]} & \emph{pear} / match \\

\taskRepetition & The operator's recent replies: yes yes yes yes \texttt{[MASK]} &
\emph{yes} / yes \\

\taskFuncComp & \texttt{m = fork; p = seal; h = m; h = [MASK]} &
\emph{fork} / result \\
\bottomrule
\end{tabularx}

\end{table}
\begin{table}[!tp]
    \centering
    \tiny
    \caption{Task-dependent neutral dataset overview.}
\label{tab:neutral-inventory}
\begin{tabularx}{\linewidth}{@{}>{\raggedright\arraybackslash}p{3.5cm}
  >{\raggedleft\arraybackslash}p{0.8cm}
  >{\raggedright\arraybackslash}X@{}}
\toprule
Task & Size & Source and Hugging Face ID \\
\midrule
\taskGender, \taskRace, \taskReligion, \taskMIBIOI, \taskKeyValue, \taskInduction, \taskRepetition, \taskFuncComp & 4,593,588 & \href{https://huggingface.co/datasets/aieng-lab/biasneutral-ajibawa}{\texttt{aieng-lab/biasneutral-ajibawa}} \citep{drechsel2026gradiend} \\
\taskEmotion & 1,000 & \href{https://huggingface.co/datasets/aieng-lab/en-sentiment-nrc-neutral}{\texttt{aieng-lab/en-sentiment-nrc-neutral}} \citep{drechsel2026gradiendpythonpackageendtoend} \\
\taskPronNum, \taskPronPers & 10,000 & \href{https://huggingface.co/datasets/aieng-lab/en-pronoun-neutral}{\texttt{aieng-lab/en-pronoun-neutral}} \citep{drechsel2026gradiendpythonpackageendtoend} \\
\taskRavelCont & 5,000 & Source: \href{https://huggingface.co/datasets/hij/ravel}{\texttt{hij/ravel}}; Processed: \iflink \href{https://huggingface.co/datasets/aieng-lab/gradiend-ravel-continent-neutral}{\texttt{aieng-lab/gradiend-ravel-continent-neutral}} \else \texttt{anonymous} \fi (new) \\
\taskRavelCountry & 5,000 & Source: \href{https://huggingface.co/datasets/hij/ravel}{\texttt{hij/ravel}}; Processed: \iflink \href{https://huggingface.co/datasets/aieng-lab/gradiend-ravel-country-neutral}{\texttt{aieng-lab/gradiend-ravel-country-neutral}} \else \texttt{anonymous} \fi (new) \\
\taskRavelLang & 5,000 & Source: \href{https://huggingface.co/datasets/hij/ravel}{\texttt{hij/ravel}}; Processed: \iflink \href{https://huggingface.co/datasets/aieng-lab/gradiend-ravel-language-neutral}{\texttt{aieng-lab/gradiend-ravel-language-neutral}} \else \texttt{anonymous} \fi (new) \\
\taskLangID & 10,000 & Source: \href{https://huggingface.co/datasets/Helsinki-NLP/opus-100}{\texttt{Helsinki-NLP/opus-100}} and \href{https://github.com/facebookresearch/MUSE}{\texttt{facebookresearch/MUSE}}; Processed: \iflink \url{https://github.com/aieng-lab/iend-study/blob/main/study/data/language_parallel.py}\else \texttt{anonymous} \fi (new) \\
\bottomrule
\end{tabularx}
\end{table}

Table~\ref{tab:task-inventory} provides an overview of the 15 evaluated tasks, including their feature classes, supported feature constructions, provenance, and available examples per class. 
Table~\ref{tab:task-examples} gives one representative example per task. 
Reported dataset sizes refer to the full available datasets and are given per class. 
The numbers of examples used for training, validation, detection evaluation, and intervention evaluation in this study are reported separately in Appendix~\ref{app:training}.

We distinguish three forms of dataset provenance. \emph{Reused} tasks use existing processed datasets, \emph{adapted} tasks transform existing benchmark or corpus data for our targeted feature learning setting, and \emph{generated} tasks use controlled generators developed for this study. We release the newly generated datasets and their generators together with the study.

\paragraph{Prediction contexts and alternatives.} 
Each task provides a context containing a \texttt{[MASK]} marker, a factual target, and its feature class. 
We retain \texttt{[MASK]} in Table~\ref{tab:task-examples} to make the target position and surrounding context explicit, even though this study uses decoder-only models that rely only on the left context up to the \texttt{[MASK]}.

\paragraph{Splits.}
For datasets whose splits are constructed as part of this study, we use an
80/10/10 train, validation, and test split, with the corresponding underlying entities kept disjoint where applicable.
For reused datasets with predefined splits, such as \taskMIBIOI{}, we retain the original split.

\paragraph{\taskMIBIOI{}.}
This task is adapted from the Indirect Object Identification (IOI) task
distributed with the Mechanistic Interpretability Benchmark (MIB; \cite{mueller2025mib}).
Each example contains an indirect object and a subject name. We use the indirect object as the factual target and the subject as the alternative target, yielding the one-sided IO-versus-subject feature used in our study. We retain the original MIB data split.

\paragraph{RAVEL.}
\taskRavelCont{}, \taskRavelCountry{}, and \taskRavelLang{} are adapted from the RAVEL benchmark \citep{huang-etal-2024-ravel}, as distributed through MIB.
They query the continent, country, or language associated with a city. 
For each attribute, we retain the three most frequent values in the RAVEL training split and sample an equal number of entities from each class. 
The city's true attribute value is the factual target, while a value from one of the other retained classes provides the alternative target. 
We remove city surface forms that map to multiple values of the queried attribute (e.g., \emph{Amsterdam} in the Netherlands and the United States), collapse duplicate entities, and use each retained entity once with entity-disjoint splits. 
Consequently, the dataset sizes reflect distinct, unambiguous entities rather than the repeated query/intervention pairs in the original RAVEL data.

\paragraph{\taskLangID{}.}
This task is a three-way language-identification task with classes English, French, and German. We obtain monolingual sentences from OPUS bitext \citep{zhang-etal-2020-improving,tiedemann-2012-parallel}: 
English and French sentences are drawn from English--French data, and German sentences from English--German data. 
Only one side of each bitext pair is used, so every prediction context contains a single language.
For each sentence, we mask a content word with at least ten words of left context. 
The original word is the factual target, while the alternative is its translation into one of the two rival languages obtained from the corresponding MUSE bilingual lexicon \citep{lample2018word}. 
We exclude stop words, short non-content candidates, words without a MUSE translation, identical cognates after normalization, and duplicate masked contexts. 
The resulting dataset is balanced across the three target languages.

\paragraph{\taskKeyValue{}.}
This task is a symbolic retrieval task in which the model predicts the
value associated with a queried key from two key--value bindings, e.g.,
\[
\texttt{k1 = red; k2 = blue; k1 = [MASK]}.
\]
Keys are sampled without replacement from 30 manually specified identifiers
(\texttt{k1}--\texttt{k20}, \texttt{key1}--\texttt{key5}, and
\texttt{slot1}--\texttt{slot5}).
Values are sampled without replacement from a manually curated vocabulary of
75 single-token colors, animals, and objects, with disjoint value vocabularies
across train, validation, and test splits.
The queried key is chosen from the two bindings: its associated value is the
factual target, while the other value serves as the alternative target.
All examples use the fixed canonical format shown above, with distinct keys
and values and no duplicate masked contexts.

\paragraph{\taskInduction{}.}
This task is a sequence-completion task based on a repeated three-token
unit.
Three distinct tokens are sampled, the unit is repeated once, and its first
token is appended as the query prefix; the model must then predict the second
token of the unit.
For example,
\[
\texttt{j pear green j pear green j [MASK]}
\]
has factual target \emph{pear}.
Tokens are drawn from a manually curated 50-token vocabulary consisting of the
26 lowercase Latin letters and 24 tokenizer-verified short words.
The vocabulary is partitioned into disjoint train, validation, and test pools
of 26, 12, and 12 tokens, respectively.
The alternative is sampled either from another token in the unit or from three
additional tokens outside the unit.

\paragraph{\taskFuncComp{}.}
This task is a symbolic alias-resolution task in which one variable
aliases a variable holding the target value.
Variable names are sampled without replacement from the 26 lowercase Latin
letters, while values are drawn from the same split-disjoint 75-value
vocabulary used for \taskKeyValue{}.
Examples contain one alias step and zero or one unrelated distractor
assignment, using fixed canonical formatting.
For example,
\[
\texttt{m = fork; p = seal; h = m; h = [MASK]}
\]
has factual target \emph{fork}.
If an unrelated assignment is present, its value serves as the alternative;
otherwise, a different value is sampled from the split-local vocabulary.
Alias and distractor variables are distinct, distractor values never equal the factual target, and duplicate masked contexts are removed.

\paragraph{Neutral data.}
In addition to labeled task examples, the study uses neutral examples that do
not express any of the feature classes of the corresponding task.
They support both detection evaluation and intervention evaluation, including
language modeling preservation measurements.
Table~\ref{tab:neutral-inventory} summarizes the neutral sources and their
task assignments.
For the semantic and grammatical tasks, we use existing neutral corpora matched
to the corresponding feature domain.
For the algorithmic tasks, general natural-language text serves as an
out-of-task reference: it does not instantiate the symbolic computation being
tested and therefore supports evaluation of feature specificity, while also
providing ordinary language contexts for measuring preservation under
intervention.
For the RAVEL tasks, we instead construct domain-matched neutral examples from city prompts that query a different attribute, preserving the entity domain and prompt format without expressing the evaluated feature.
For \taskLangID{}, neutral examples use Spanish, Italian, Dutch, and Portuguese sentences in the same input format as the labeled data; these languages are not among the task's three target classes.

\section{Experimental details}
\label{app:experimental-details}

\subsection{Implementation}\label{app:implementation}
We build on the \texttt{gradiend} Python package \citep{drechsel2026gradiendpythonpackageendtoend}, extending its codebase with
the generalized IEND implementations introduced in this work, and release this extended codebase.\footnote{\iflink \url{https://github.com/aieng-lab/gradiend-iend} \else\url{anonymous}\fi}
The experiments are implemented in a separate study repository, which depends on this extended \texttt{gradiend} package and contains the task definitions, implementations of \caa, \cga, \caga, and the \sae\ baselines, as well as the evaluation and analysis code.\footnote{\iflink \url{https://github.com/aieng-lab/iend-study}\else\url{anonymous}\fi}
Both repositories include documentation and configuration files for reproducing the experiments reported in this paper.

\subsection{Models}\label{app:models}

We evaluate \rev{ten} pretrained decoder-only transformer language models \citep{attention} spanning different model families and scales.
Table~\ref{tab:model-sae-config} lists the exact Hugging Face checkpoints and the pretrained SAELens resources used for the \sae\ reference.

\begin{table*}[!t]
    \centering
    \tiny
    \caption{Language-model checkpoints and pretrained SAE resources used in
    the study.}
    \label{tab:model-sae-config}

    \setlength{\tabcolsep}{2.8pt}
    \renewcommand{\arraystretch}{1.05}

\begin{tabularx}{0.95\textwidth}{
    @{}
    p{1.45cm}
    p{3.6cm}
    p{4.0cm}
    p{3.5cm}
    @{}
}
        \toprule
        \textbf{Model} &
        \textbf{HF checkpoint} &
        \textbf{SAELens release} &
        \textbf{SAE ID / hook} \\
        \midrule

        \pythia
        & \href{https://huggingface.co/EleutherAI/pythia-70m-deduped}
               {\nolinkurl{EleutherAI/pythia-70m-deduped}}
        & \href{https://huggingface.co/ctigges/pythia-70m-deduped__res-sm_processed}
               {\texttt{pythia-70m-deduped-res-sm}}
        & \texttt{blocks.\{L\}.hook\_resid\_post} \\

        \gpttwo
        & \href{https://huggingface.co/openai-community/gpt2}
               {\texttt{gpt2}}
        & \href{https://huggingface.co/jbloom/GPT2-Small-OAI-v5-32k-resid-post-SAEs}
               {\texttt{gpt2-small-resid-post-v5-32k}}
        & \texttt{blocks.\{L\}.hook\_resid\_post} \\

       \rev{\gemmaS}
        & \href{https://huggingface.co/google/gemma-3-1b-pt}
               {\texttt{google/gemma-3-1b-pt}}
        & \href{https://huggingface.co/google/gemma-scope-2-1b-pt}
               {\nolinkurl{gemma-scope-2-1b-pt-res-all}}
        & \texttt{layer\_\{L\}\_width\_16k\_l0\_small} \\

        \rev{\qwenS}
        & \href{https://huggingface.co/Qwen/Qwen3.5-2B-Base}
               {\texttt{Qwen/Qwen3.5-2B-Base}}
        & \href{https://huggingface.co/Qwen/SAE-Res-Qwen3.5-2B-Base-W32K-L0_50}
               {\nolinkurl{qwen-scope-3.5-2b-base-w32k-l50}}
        & \texttt{layer\{L\}} \\

        \rev{\gemma}
        & \href{https://huggingface.co/google/gemma-2-2b}
               {\texttt{google/gemma-2-2b}}
        & \href{https://huggingface.co/google/gemma-scope-2b-pt-res}
               {\nolinkurl{gemma-scope-2b-pt-res-canonical}}
        & \texttt{layer\_\{L\}/width\_16k/canonical} \\

        \rev{\gemmaM}
        & \href{https://huggingface.co/google/gemma-3-4b-pt}
               {\texttt{google/gemma-3-4b-pt}}
        & \href{https://huggingface.co/google/gemma-scope-2-4b-pt}
               {\nolinkurl{gemma-scope-2-4b-pt-res-all}}
        & \texttt{layer\_\{L\}\_width\_16k\_l0\_small} \\

        \rev{\mistral}
           & \href{https://huggingface.co/mistralai/Mistral-7B-v0.1}
               {\texttt{mistralai/Mistral-7B-v0.1}}
        & \href{https://huggingface.co/JoshEngels/Mistral-7B-Residual-Stream-SAEs}
               {\texttt{mistral-7b-res-wg}}
        & \texttt{blocks.\{L+1\}.hook\_resid\_pre} \\

        \llama
        & \href{https://huggingface.co/meta-llama/Llama-3.1-8B}
               {\texttt{meta-llama/Llama-3.1-8B}}
        & \href{https://huggingface.co/fnlp/Llama3_1-8B-Base-LXR-32x}
               {\texttt{llama\_scope\_lxr\_32x}}
        & \texttt{l\{L\}r\_32x} \\

       \rev{\deepseek}
       & \href{https://huggingface.co/deepseek-ai/DeepSeek-R1-Distill-Llama-8B}
               {\nolinkurl{deepseek-ai/DeepSeek-R1-Distill-Llama-8B}}
        & \href{https://huggingface.co/OpenMOSS-Team/Llama-Scope-R1-Distill}
               {\texttt{llama\_scope\_r1\_distill}}
        & \texttt{l\{L\}r\_800m\_slimpajama} \\

               \rev{\qwenL}
        & \href{https://huggingface.co/Qwen/Qwen3.5-9B-Base}
               {\texttt{Qwen/Qwen3.5-9B-Base}}
        & \href{https://huggingface.co/Qwen/SAE-Res-Qwen3.5-9B-Base-W64K-L0_50}
               {\nolinkurl{qwen-scope-3.5-9b-base-w64k-l50}}
        & \texttt{layer\{L\}} \\

        \bottomrule
    \end{tabularx}
\end{table*}

\subsection{Computational cost and memory footprint}
\label{app:runtime}

\begin{figure}[!t]
    \centering
    \includegraphics[width=\linewidth]{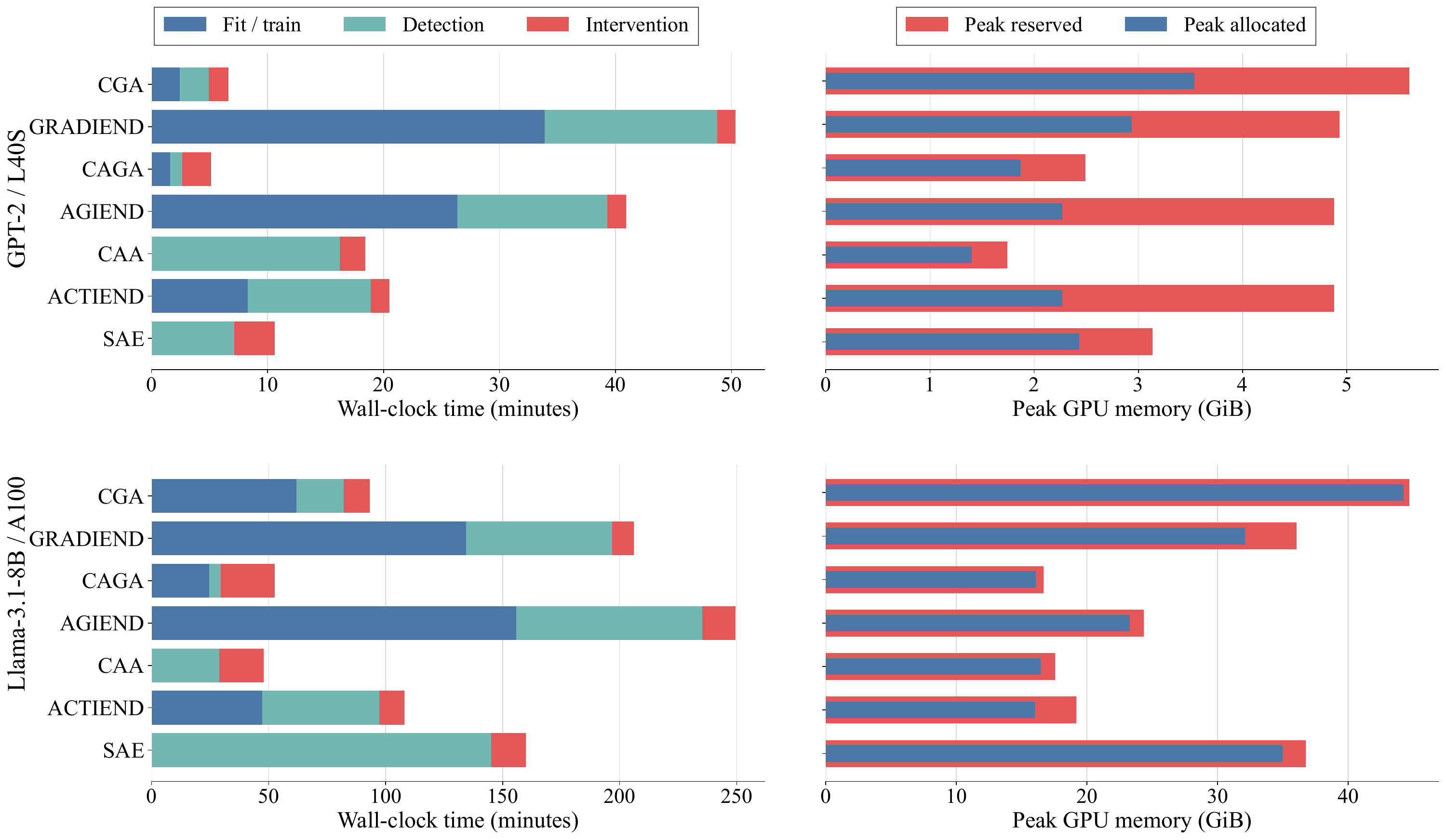}
    \caption{Computational cost and peak GPU memory on \taskGender.
    Left: wall-clock time separated into fitting or training, detection, and
    intervention evaluation.
    Right: peak allocated and reserved CUDA memory across these stages.
    Reserved memory includes allocated memory and is therefore shown as an
    overlaid rather than additive quantity.
    \sae\ pretraining is excluded because we use pretrained dictionaries.}
    \label{fig:runtime-hardware}
\end{figure}

Figure~\ref{fig:runtime-hardware} reports the computational cost of one complete model-task-method evaluation cell for \taskGender.
For each method and model, this comprises task-specific feature fitting or selection, held-out detection evaluation, and intervention evaluation.
For \sae, pretrained dictionary learning is excluded, while task-specific feature selection is included.

Importantly, these measurements characterize the evaluated implementations under the observed execution conditions rather than intrinsic algorithmic complexity, and should therefore be interpreted as rough runtime
indicators.
\gpttwo\ was run on an L40S and \llama\ on an A100, so absolute runtime should only be compared within each model.
The clearest runtime difference is between estimators:
the \iend\ variants are consistently more expensive than their corresponding contrastive-mean methods, with particularly large increases for the gradient-based \agiend\ and \gradiend.
For \llama, \sae\ also incurs substantial runtime despite requiring no feature training, as its post-hoc procedure sweeps all layers, each with a substantially larger latent dictionary on a wider model.
Peak memory follows the expected pattern, with parameter-gradient methods requiring substantially more memory than activation-space methods.

\subsection{Signal extraction}
\label{app:signals}

Let $h_\ell$ denote the residual stream output after transformer block $\ell$ and $\theta_\ell$ the parameters of that block.
For activation value and activation gradient signals, the single-layer and all-layer scopes are
\begin{equation} % remove eq line
    \Omega_\ell^{h}=h_\ell,
    \qquad
    \Omega_{\mathrm{all}}^{h}=(h_1,\ldots,h_L),
\end{equation}
while for parameter-gradient signals, they are
\begin{equation}  % remove eq line
    \Omega_\ell^{\theta}=\theta_\ell,
    \qquad
    \Omega_{\mathrm{all}}^{\theta}=(\theta_1,\ldots,\theta_L).
\end{equation}
Embeddings and the final language-model head are excluded.
For all-layer scopes, the corresponding layer-specific signals are concatenated in transformer-layer order.

\paragraph{Target spans.}
For each labeled example, let $x$ denote the context preceding the feature-related target and let $y=(y_0,\ldots,y_{k-1})$ denote the target token span whose realization determines the corresponding feature class (e.g., a target ``Asia'' could be tokenized as $y_0=$``Asi'' and $y_1=$``a'').
Let $p,\ldots,p+k-1$, with $p\geq1$, denote the sequence positions occupied by $y$.
In a pairwise contrast, the same context $x$ is paired with a factual target $y^F$ and an alternative target $y^A$ realizing different classes. In the definitions below, $y$ denotes either candidate target.

\paragraph{Activation values.}
Activation value methods use the mean residual stream representation over the filled target span,
\begin{equation}
    \phi^{h}_{\ell}(x,y)
    =
    \frac{1}{k}
    \sum_{j=0}^{k-1}
    h_{\ell,p+j}(x,y).
\end{equation}
Thus, the activation value signal is extracted after the tokens realizing the feature have entered the causal context.

\paragraph{Gradient signals.}
Following the original \gradiend\ formulation
\citep{drechsel2026gradiend}, gradient-based methods condition on the first
token $y_0$ of the target through the causal-language-modeling loss
\begin{equation}
    \mathcal{L}_{\mathrm{CLM}}(x,y)
    =
    -\log p_\theta(y_0 \mid x).
\end{equation}
The activation gradient signal at layer $\ell$ is the gradient of this loss with respect to the residual stream at the preceding prediction position,
\begin{equation}
    \phi^{\nabla h}_{\ell}(x,y)
    =
    \nabla_{h_{\ell,p-1}}
    \mathcal{L}_{\mathrm{CLM}}(x,y).
\end{equation}
Parameter gradient methods use the same objective and compute
\begin{equation}
    \phi^{\nabla\theta}_{\ell}(x,y)
    =
    \nabla_{\theta_\ell}
    \mathcal{L}_{\mathrm{CLM}}(x,y).
\end{equation}

\paragraph{Activation-steering positions.}
Following common activation-steering practice \citep{actadd,caa}, all
activation-space interventions are applied across all prefix token positions, rather than only at the position immediately preceding the target.

\paragraph{Target-position asymmetry.}
Because factual and alternative targets in a contrast share the same preceding context, their residual stream representations at the prediction position $p-1$ are identical.
activation value contrasts are therefore extracted from the filled target span, whereas gradient signals condition on the first target token through
the next-token loss at $p-1$.

\subsection{Training and selection}
\label{app:training}

\begin{table*}[t]
    \centering
    \scriptsize
    \caption{Training, selection, and evaluation hyperparameters.}
    \label{tab:training-config}
    \setlength{\tabcolsep}{3pt}
    \renewcommand{\arraystretch}{1.0}
    \begin{tabularx}{\textwidth}{@{}p{4.4cm}X@{}}
        \toprule
        \textbf{Setting} & \textbf{Value} \\
        \midrule

        \multicolumn{2}{@{}l}{\textit{IEND optimization}} \\

%\gradiend\ learning rate & $10^{-4}$ (\gpttwo), $10^{-6}$ (\pythia), %, \qwen),
  %$1{\times}10^{-7}$ (\gemma),  $2{\times}10^{-6}$ (\llama) \\

%\actiend\ encoder learning rate & $5{\times}10^{-5}$ (\pythia, \gpttwo),
  %$3{\times}10^{-5}$ (\gemma),   $10^{-5}$ (\llama) \\ % \qwen, 

\actiend\ decoder learning rate
& $1.15{\times}10^{-2}$ \\

%\agiend\ learning rate & $3{\times}10^{-3}$ (\pythia, \gpttwo, %\gemma,  \llama)\\%; $10^{-3}$ for \qwen \\

        Training steps
        & 500 (\gradiend, \agiend); 100 (\actiend) \\

        Evaluation interval
        & 50 steps (\gradiend, \agiend); 20 steps (\actiend) \\

        Training batch size
        & 8 \\

        Maximum sequence length
        & 128 tokens \\

        Random seeds
        & Up to 3, starting from seed 0; the first converged run is retained \\

        \midrule
        \multicolumn{2}{@{}l}{\textit{Parameter-gradient methods}} \\

\gradiend\ pre-pruning
& 16 examples;  \rev{retain the top-$k$ coordinates (Table~\ref{tab:model-hparams})} \\

\gradiend\ post-pruning
& Retain top $1\%$ of decoder weights \\

\cga\ coordinate scope
& Full parameter gradient (\gpttwo, \pythia);
  same pre-pruned coordinates as \gradiend\ %(\gemma, 
  (other models) \\%\qwen, 

        \midrule
        \multicolumn{2}{@{}l}{\textit{Sampling and selection}} \\

        Neutral-example caps
        & 1,000 training / 300 validation / 1,000 test \\

        Labeled encoder-evaluation caps
        & 50 training / 200 validation examples per class \\

        Intervention-evaluation caps
        & 200 validation / 1,000 test examples per class \\

        \sae\ latent eligibility
        & Activation on at least $1\%$ of labeled validation examples \\

        \midrule
        \multicolumn{2}{@{}l}{\textit{Intervention selection}} \\

Candidate magnitudes
& $10^{-5},\,2{\times}10^{-5},\,5{\times}10^{-5},\,
10^{-4},\,\ldots,\,10^{5}$ (31 values) \\
   
        Preservation constraint
        & $\mathrm{LMS}(I)\geq0.99\,\mathrm{LMS}(I_0)$ \\

        \bottomrule
    \end{tabularx}
\end{table*}

\begin{table}[t]
\centering\scriptsize
\caption{\rev{Model-specific \iend\ and pruning settings. Models with different pairwise and one-sided learning rates list both (pairwise / one-sided).}}
\label{tab:model-hparams}
\begin{tabular}{lcccc}
\toprule
& \multicolumn{3}{c}{Learning rate} & \\
\cmidrule(lr){2-4}
Model & \gradiend & \actiend & \agiend & Pre-pruning top-$k$ \\
\midrule
\pythia  & $10^{-6}$ & $5{\times}10^{-5}$ & $3{\times}10^{-3}$ & 10\% \\
\gpttwo  & $10^{-4}$ & $5{\times}10^{-5}$ & $3{\times}10^{-3}$ & 10\% \\
\rev{\gemmaS}  & $2{\times}10^{-6}$ & $5{\times}10^{-5}$ & $10^{-2}$ / $10^{-3}$ & 1\% \\
\rev{\qwenS}   & $10^{-6}$ & $10^{-5}$ & $10^{-3}$ & 1\% \\
\rev{\gemma}  & $2{\times}10^{-6}$ & $3{\times}10^{-5}$ & $5{\times}10^{-3}$ & 1\% \\
\rev{\gemmaM}  & $2{\times}10^{-6}$ & $10^{-5}$ & $10^{-3}$ & 1\% \\
\rev{\mistral} & $2{\times}10^{-6}$ & $10^{-5}$ & $3{\times}10^{-3}$ & 1\% \\
\llama   & $2{\times}10^{-6}$ & $10^{-5}$ & $3{\times}10^{-3}$ & 1\% \\
\rev{\deepseek} & $3{\times}10^{-5}$ / $2{\times}10^{-6}$ & $10^{-5}$ & $3{\times}10^{-3}$ & 1\% \\
\rev{\qwenL}   & $10^{-5}$ & $10^{-5}$ & $3{\times}10^{-3}$ & 1\% \\
\bottomrule
\end{tabular}
\end{table}

Table\rev{s}~\ref{tab:training-config} \rev{and~\ref{tab:model-hparams}} summarize the training, sampling, and selection hyperparameters.
The following sections provide the corresponding method-specific details.

\paragraph{IEND convergence.}
For pairwise \iend, convergence is assessed using the correlation between the
encoder output and the desired labels, assigning the two contrasted classes
$-1$ and $+1$ and neutral examples $0$.
This follows the established \gradiend\ workflow, which has so far considered
pairwise tasks only \citep{drechsel2026gradiendpythonpackageendtoend}.
For the newly introduced one-sided setting, we instead use AUROC for distinguishing the target class from the pooled non-target examples.
This avoids rewarding separation between neutral examples and alternative target classes, which is irrelevant to the one-sided objective.

\paragraph{Representation selection.}
For \sae, \caa, \cga, and \caga, the candidate set comprises all available individual layer scopes and the corresponding all-layer scope defined in Section~\ref{app:signals}.
For each evaluated feature, we select the scope $\Omega^\star$ with the highest validation $\mathrm{Det.}$.
The selected scope is then fixed for both held-out detection and intervention.

We do not perform the analogous layer sweep for the IEND methods.
For the contrastive mean methods, once the layerwise signals have been
extracted, additional scopes can be evaluated by applying the same
mean-difference estimator to the corresponding signals.
For IEND, each candidate scope instead requires training a separate encoder-decoder representation.
We therefore use the all-layer scope throughout:
$\Omega_{\mathrm{all}}^\theta$ for \gradiend\ and
$\Omega_{\mathrm{all}}^h$ for \actiend\ and \agiend.

\paragraph{Evaluation samples.}
Sampling limits are applied per feature class and are summarized in Table~\ref{tab:training-config}.
Neutral examples are sampled independently from the corresponding training, validation, and test pools.
Neutral training examples are used only by the IEND methods (\gradiend, \actiend, and \agiend), where neutral texts form identity transitions with zero desired change and thereby regularize the learned
representation toward neutral behavior.

Validation neutrals are used for detection-based selection and threshold estimation and, for \sae, additionally enter latent ranking through the neutral-specificity term.
They also provide the preservation set used when selecting intervention
strengths.
Test neutrals are reserved for final evaluation

\paragraph{Threshold selection.}
Classification thresholds are selected on validation data by maximizing
Youden's $J=\mathrm{TPR}+\mathrm{TNR}-1$.
For $\mathrm{Spec}_n$, we fit a target-versus-neutral decision rule.
For $\mathrm{Excl}$, we fit a separate target-versus-rival decision rule for
each rival class.
For each rule, both score orientation and threshold are determined on validation data and then applied unchanged to the held-out test split.

\paragraph{SAE selection.}
Following prior supervised approaches that identify concept-relevant SAE
latents from class-conditional activations
\citep{karvonen2025saebench,gallifant-etal-2025-sparse},
we rank eligible latents on the validation split.
For target class $c$, layer $\ell$, and \sae\ latent $j$, the ranking score is
\begin{equation}
    s_{\ell cj}
    =
    \bar z_{\ell cj}
    -
    \bar z_{\ell,\neg c,j}
    -
    \left|\bar z_{\ell,N,j}\right|,
\end{equation}
where $\bar z_{\ell cj}$ is the mean latent activation on examples of class
$c$, $\bar z_{\ell,\neg c,j}$ the mean over labeled non-$c$ examples, and
$\bar z_{\ell,N,j}$ the mean over neutral examples.
Latents firing on fewer than $1\%$ of labeled validation examples are ineligible.

Eligible latents are ranked separately for each feature class and layer.
An individual-layer candidate uses the top-ranked eligible latent at that layer.
For the all-layer candidate, we independently select the top-ranked eligible latent at every available layer and sum their activations without weighting.
These candidates then enter the representation-selection procedure described above.

\paragraph{Parameter-gradient pruning.}
\gradiend\ uses the pre- and post-pruning procedure introduced with the \texttt{gradiend} package
\citep{drechsel2026gradiendpythonpackageendtoend}, with the settings reported
in Table~\ref{tab:training-config}.
\cga\ uses the full parameter-gradient signal for \gpttwo\ and \pythia, and
the same pre-pruned coordinate set as \gradiend\ for %\gemma\ and 
\llama.

\subsection{Intervention evaluation}
\label{app:intervention-evaluation}

%For gradient-based methods, intervention directions are sign-reversed relative to the raw loss gradient, since moving toward the target follows the negative-loss direction.

The candidate intervention magnitudes and preservation constraint used for
validation-based intervention selection are summarized in
Table~\ref{tab:training-config}.

For \sae, intervention follows the decoder directions corresponding to the latents used by the selected representation.
For a single layer representation, we use the decoder direction of the
selected latent at that layer.
For an all-layer representation, the selected decoder directions are applied simultaneously at their corresponding  residual stream layers.

\subsection{Aggregation}
\label{app:aggregation}

For each method-model-task combination, we compute Det. and Int. for every evaluated target-class and average these scores with equal weight.
For pairwise constructions with $K$ classes, both endpoints of each unordered class pair are evaluated, yielding $2\binom{K}{2}$ scores. 
One-sided constructions and \sae\ yield one score per target class.

For higher-level summaries, task scores are averaged equally within each
model, and the resulting model-level means are averaged equally across models.

\subsection{Statistical analysis}
\label{app:statistics}

We compare methods using the Bayesian analysis implemented in
\textsc{Autorank} \citep{Herbold2020}.
For detection and intervention separately, each matched model--task combination is considered one observation, using the corresponding aggregated task-level score defined in Section~\ref{app:aggregation}.
Thus, individual examples, feature classes, and pairwise contrasts are not
treated as independent observations.

We use an absolute region of practical equivalence (ROPE) of $0.01$ in the respective metric scale.
A pairwise outcome is assigned when the posterior probability of one of the three decisions (the first method performing better, the second performing better, or practical equivalence) exceeds $1-\alpha=0.95$; otherwise, the comparison remains unresolved.
Posterior probabilities are estimated from 50,000 samples.

\section{Detailed results}
\label{app:detailed-results}

This section provides additional analyses supporting and extending the results in Section~\ref{sec:main-results}.

\subsection{Statistical analysis}\label{app:autorank}

\begin{figure}[!t]
    \centering
    \includegraphics[width=\linewidth]{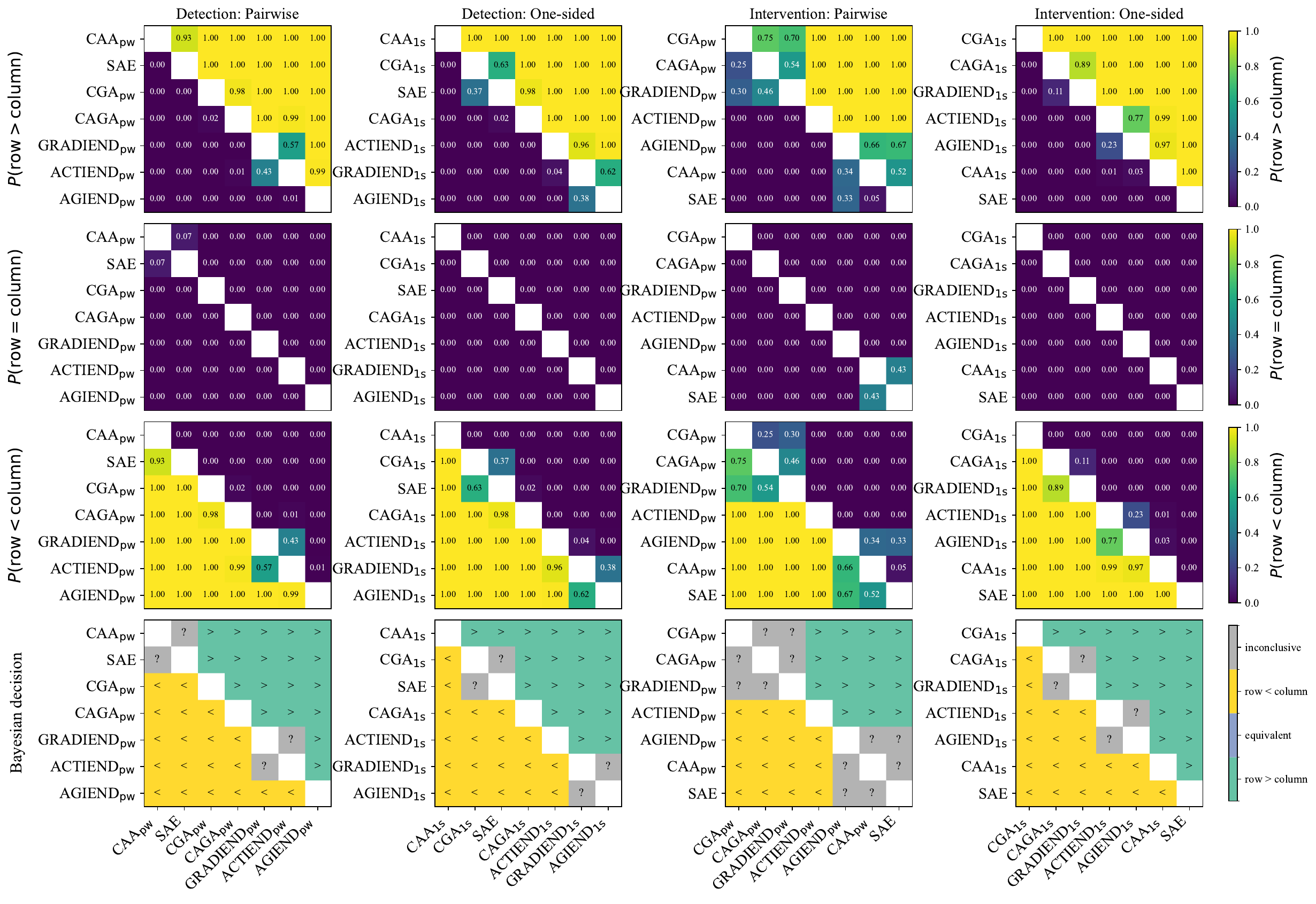}
    \caption{Bayesian pairwise comparison of methods.}
    \label{fig:autorank-4x4}
\end{figure}

Figure~\ref{fig:autorank-4x4} reports the full Bayesian pairwise comparison
underlying the summary shown in Figure~\ref{fig:bayesian-decision-overview}, including the posterior probabilities that the row method performs better, performs worse, or is practically equivalent to the column method, together with the resulting Bayesian decision.

\subsection{Layer localization and representation scope}
\label{app:layer-selection}

\begin{table}[t]
    \centering
    \small
    \caption{Effect of representation scope. Single-layer representations are selected by validation Detection. $\Delta$ denotes single layer minus all-layer performance on the held-out test split. Win rates report the fraction of class/contrast comparisons with $\Delta>0$,
i.e., where the selected single layer outperforms the all-layer representation
on the held-out test split.}
    \label{tab:layer-selection-summary}
    \begin{tabular}{lrrrrr}
        \toprule
        & \multicolumn{1}{c}{Selected}
        & \multicolumn{2}{c}{Detection}
        & \multicolumn{2}{c}{Intervention} \\
        \cmidrule(lr){2-2}
        \cmidrule(lr){3-4}
        \cmidrule(lr){5-6}
        Method
        & Single layer
        & Win
        & Mean $\Delta$
        & Win
        & Mean $\Delta$ \\
        \midrule
     \sae & 99.1\% & 56.9\% & $+0.053$ & 30.9\% & $-0.020$ \\
        \caa & 88.0\% & 87.8\% & $+0.095$ & 32.1\% & $-0.028$ \\
        \cga & 91.4\% & 84.0\% & $+0.143$ & 56.7\% & $+0.013$ \\
        \caga & 97.5\% & 90.0\% & $+0.207$ & 78.9\% & $+0.072$ \\
        \midrule
        Overall & 93.1\% & 83.7\% & $+0.137$ & 52.6\% & $+0.014$ \\
        \bottomrule
    \end{tabular}
\end{table}

For \sae, \caa, \cga, and \caga, the main experiment selection procedure considers both the predefined all-layer representation and every available single best layer representation, and selects among them by validation $\mathrm{Det.}$.
To isolate the effect of representation scope, we additionally compare the all-layer candidate with the single layer achieving the highest validation
$\mathrm{Det.}$, separately for each target class or pairwise contrast.

\paragraph{Single-layer versus all-layer representations.}
Table~\ref{tab:layer-selection-summary} summarizes the comparison.
Validation detection strongly favors single layers over the all-layer representation.
For the targeted contrastive mean methods, this preference also generalizes to held-out detection, with particularly large gains for \cga\ and \caga.
For \sae, in contrast, the strong validation preference for a single layer does not translate into a consistent held-out advantage.

\begin{figure}[!pt]
    \centering
    \includegraphics[width=\linewidth]{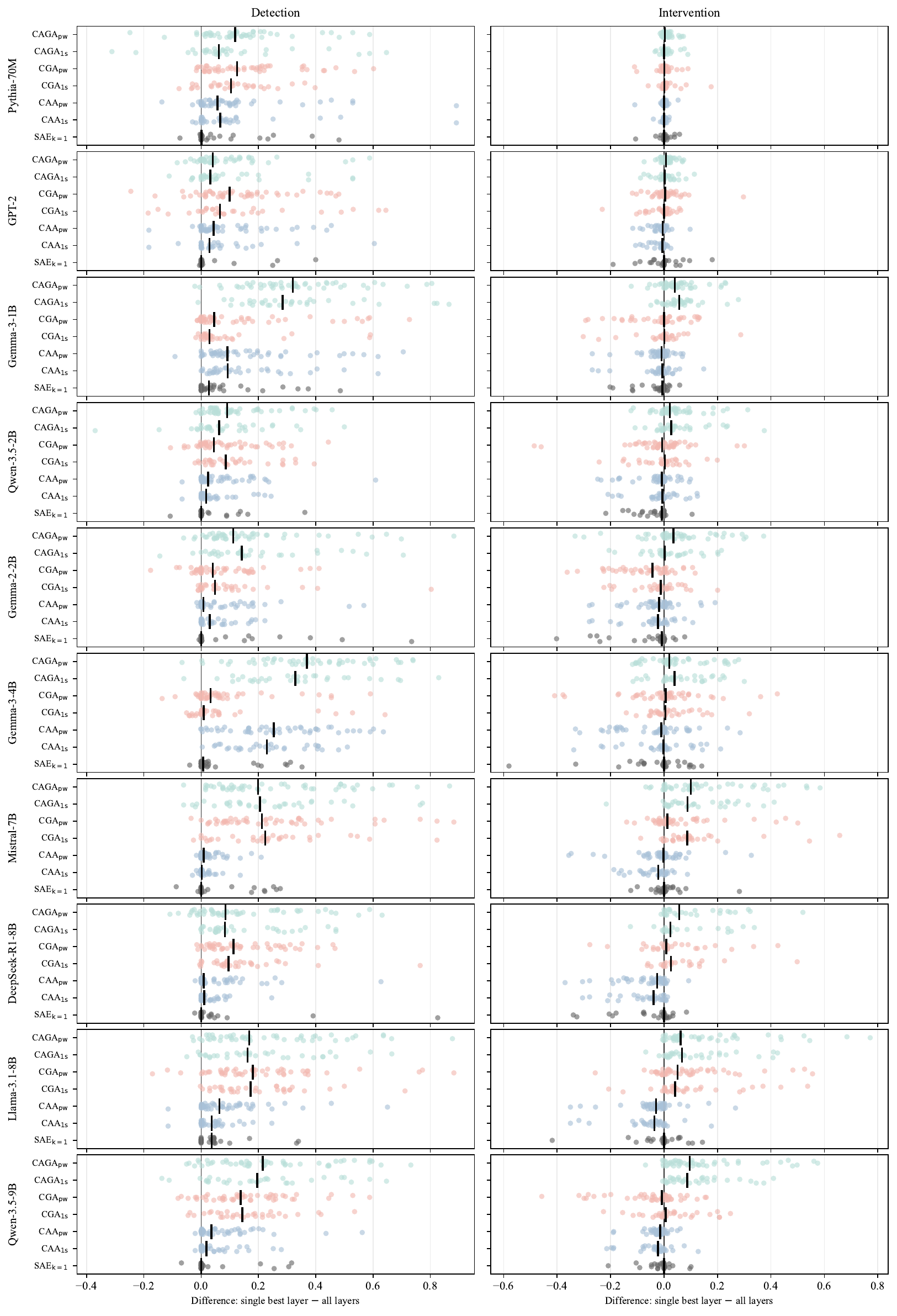}
    \caption{Difference between the validation-selected best single-layer and
    all-layer representations for detection and intervention.
    Positive values indicate higher performance for the selected single layer.}
    \label{fig:layer-selection-paired-deltas}
    % analysis/layer_selection_appendix.py
\end{figure}

Figure~\ref{fig:layer-selection-paired-deltas} shows the corresponding distribution of single-layer minus all-layer differences for detection and intervention.
Intervention does not exhibit the same localization advantage.
Thus, the layer best suited to detecting a feature is not generally the scope best suited to intervening on it.

\begin{figure}[!pt]
    \centering
    \includegraphics[width=0.8\linewidth]{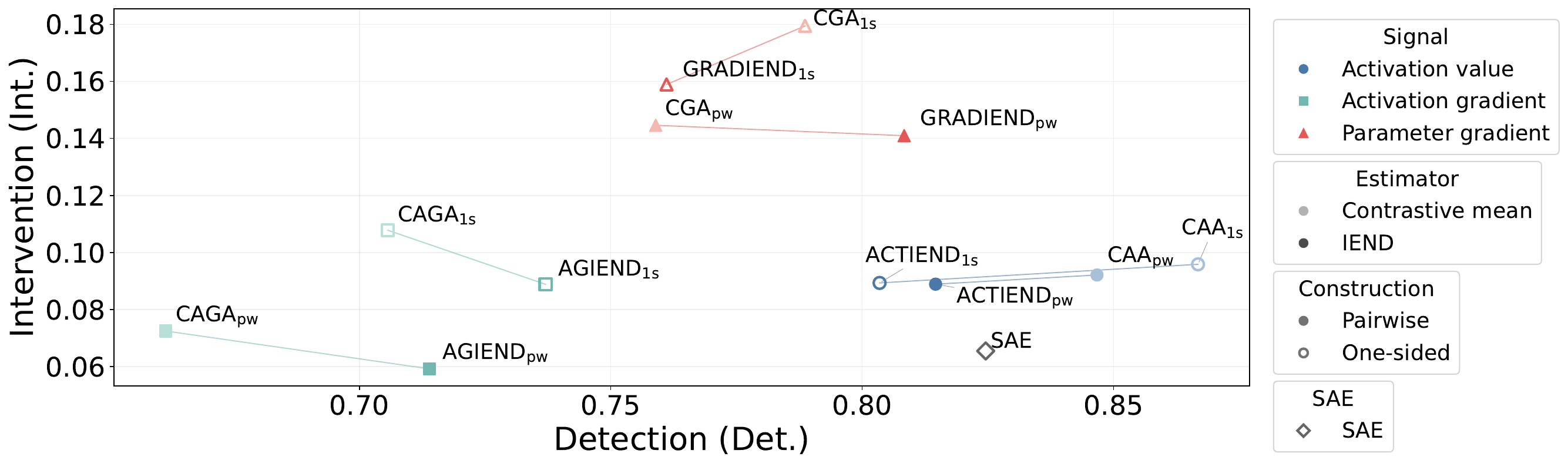}
    \caption{All-layer robustness analysis.
    Detection-intervention comparison corresponding to
    Figure~\ref{fig:detection-vs-intervention}, with \sae\ and
    contrastive mean methods restricted to their all-layer representations rather than using validation-selected scopes.}
    \label{fig:detection-intervention-all-layer}
\end{figure}

\paragraph{Robustness of the main comparison.}
To assess whether representation-scope selection affects the main comparison, Figure~\ref{fig:detection-intervention-all-layer} repeats
Figure~\ref{fig:detection-vs-intervention} with \sae\ and contrastive mean methods restricted to their all-layer representations. The qualitative conclusions remain unchanged, \rev{except that the intervention advantage of \caga\ over \caa\ largely disappears, indicating that \caga\ requires layer selection.}

\begin{figure*}[!tp]
     \centering
     \includegraphics[width=\linewidth]{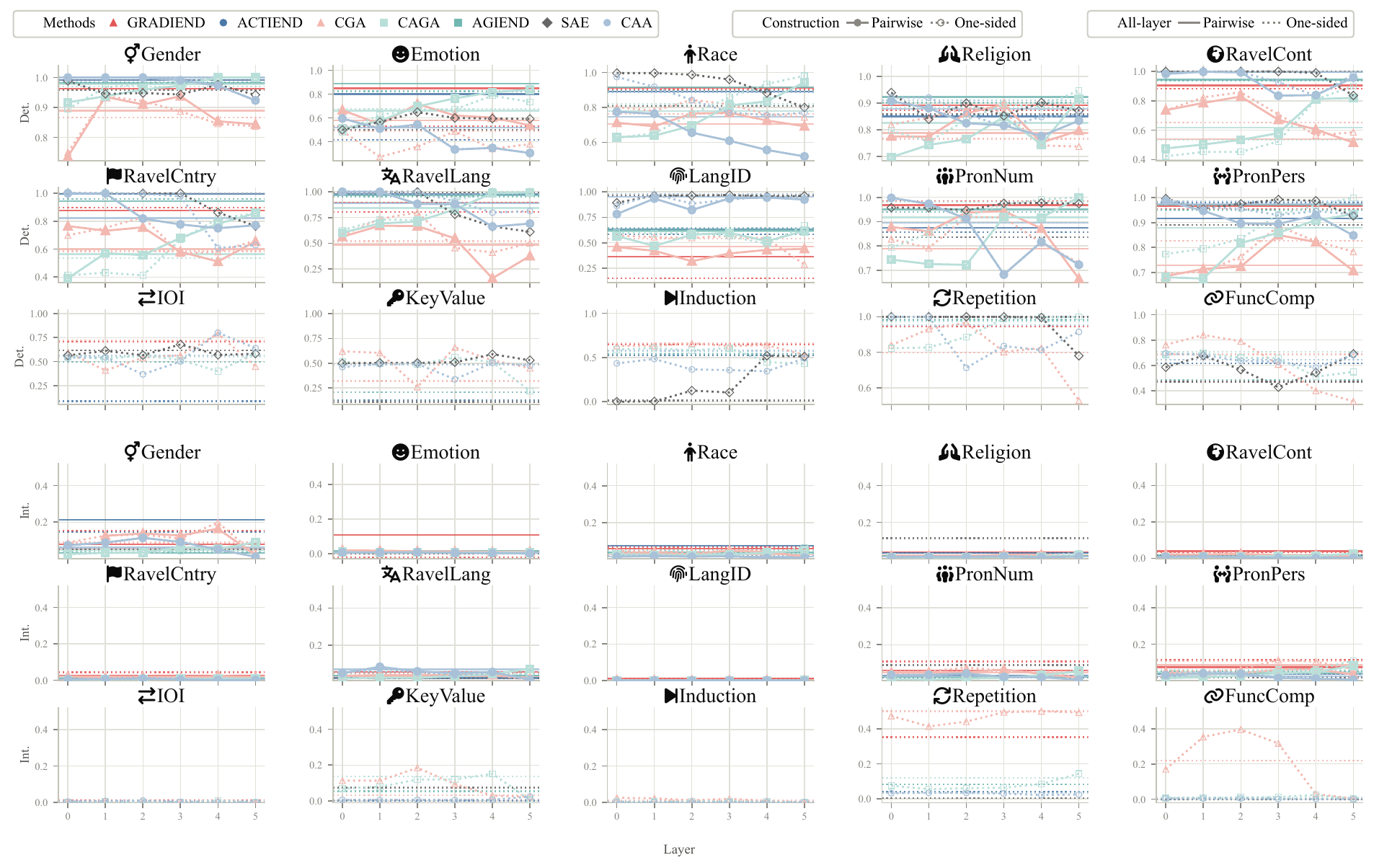}
     \caption{
     Layerwise detection and intervention for \pythia.
     %Each task occupies two vertically aligned panels showing $\mathrm{Det.}$ and $\mathrm{Int.}$ across transformer layers.
     }
     \label{fig:layerwise-pythia}
 \end{figure*}

\begin{figure*}[!tp]
     \centering
     \includegraphics[width=\linewidth]{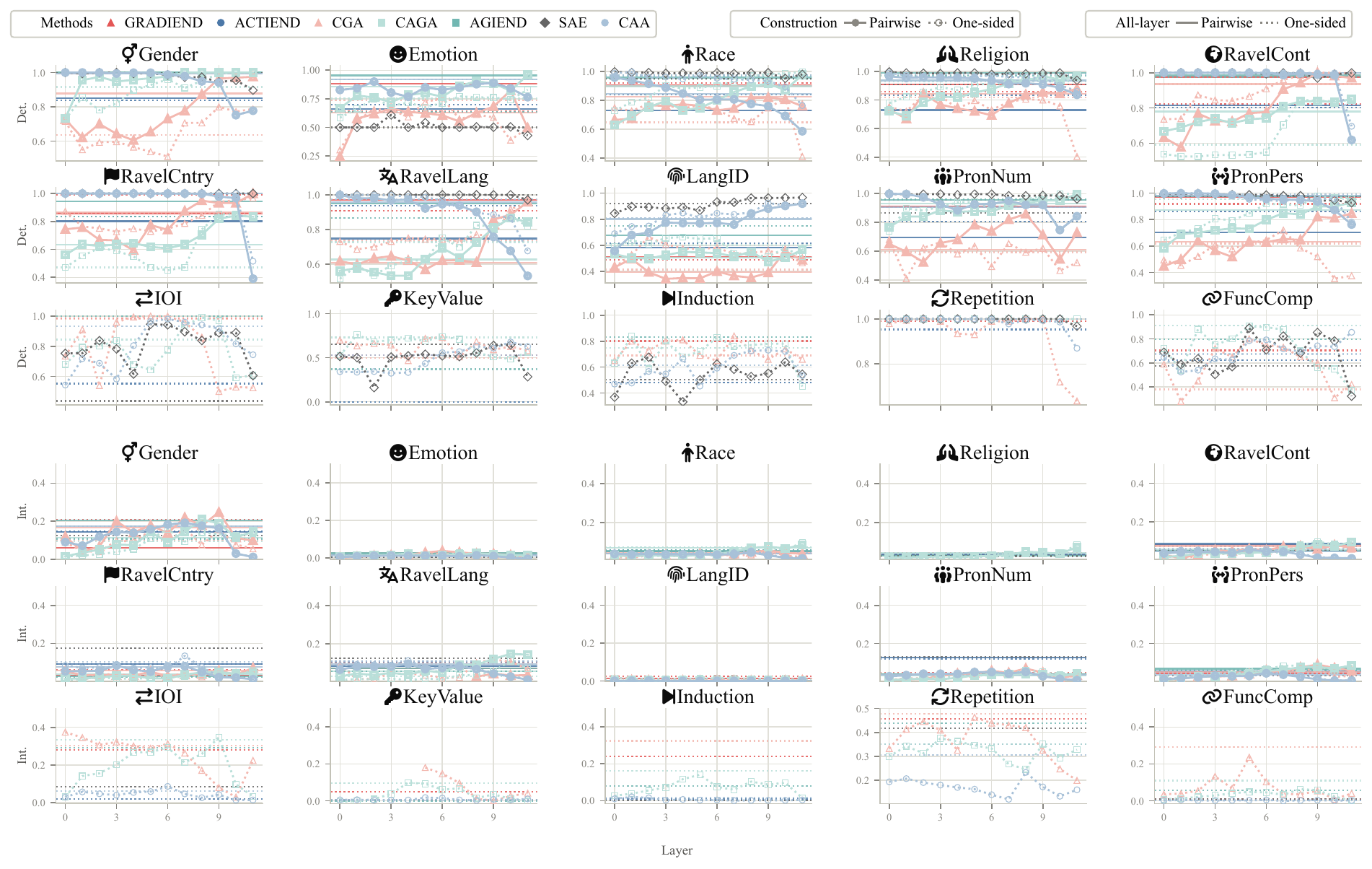}
     \caption{
     Layerwise detection and intervention for \gpttwo.
     %Each task occupies two vertically aligned panels showing $\mathrm{Det.}$ and $\mathrm{Int.}$ across transformer layers.
     }
     \label{fig:layerwise-gpt2}
 \end{figure*}

\paragraph{Layerwise profiles.}
Figures~\ref{fig:layerwise-pythia}--\ref{fig:layerwise-gpt2} report validation
detection and intervention across single layers and the all-layer
representation for each task on the two smaller models, \pythia\ and \gpttwo.
We restrict this exhaustive layerwise analysis to these models for computational tractability: \rev{the larger models expose} substantially more layers, making a full per-task layer sweep considerably more expensive.
The profiles vary substantially across tasks and methods, and layers with strong detection do not necessarily yield strong intervention.

% analysis/layer_performance_plots.py --model gpt2-small --subdir full_suite2

\subsection{Sparse autoencoder variants}
\label{app:sae-variants}

Activation value methods face a positional asymmetry between detection and intervention.
Remember that distinguishing the factual and alternative targets requires observing the filled target representation, whereas an intervention intended to affect that target must act before it is generated.
Because the pretrained \sae\ is learned independently of the requested feature, we can additionally select the relevant latent from the predictive representation immediately preceding the target.
We denote this variant by $\mathrm{SAE}_{\mathrm{pre}}$.

A feature may also be distributed across multiple \sae\ latents.
We therefore compare the single latent representation ($k=1$) used in the main experiments with a
validation selected latent set $k^\star$.
At the selected layer, we sum the activations of the top-$k$ ranked eligible latents for
\[
k\in\{1,2,4,8,16,32,64,128\}.
\]
We select $k^\star$ on validation data by maximizing $\mathrm{AUC}_n+\mathrm{AUC}_o$ when a rival class is defined and $\mathrm{AUC}_n$ otherwise, resolving ties in favor of smaller $k$.
The same procedure is applied to $\mathrm{SAE}_{\mathrm{pre}}$.

\begin{figure}[!t]
    \centering
    \includegraphics[width=\linewidth]{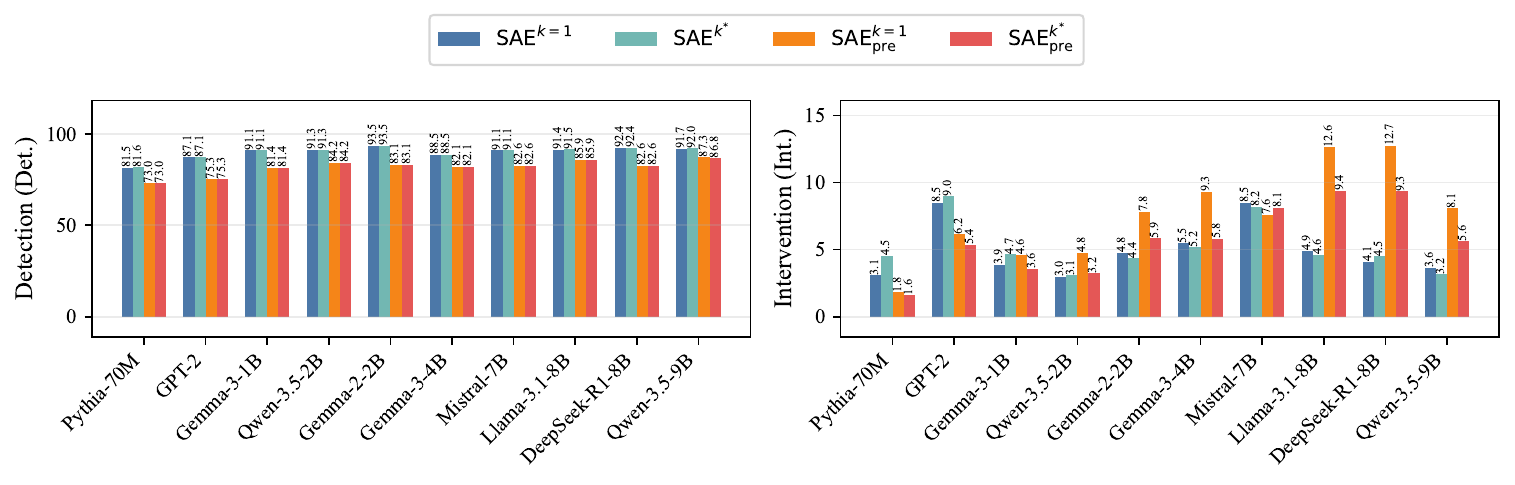}
    \caption{Detection and Intervention for the four \sae\ variants obtained
    by varying latent-selection position (filled-target vs.\ pre-target) and
    the number of selected latents ($k=1$ vs.\ $k^\star$).}
    \label{fig:sae-ablation}
\end{figure}

Together, the positional and latent-count choices yield four variants:
$\mathrm{SAE}_{k=1}$,
$\mathrm{SAE}_{k^\star}$,
$\mathrm{SAE}_{\mathrm{pre},k=1}$, and
$\mathrm{SAE}_{\mathrm{pre},k^\star}$.
Figure~\ref{fig:sae-ablation} compares their detection and intervention
performance.

% python analysis\appendix_evidence.py 
% analysis/tables/appendix_evidence/appendix_sae_variants.pdf

Increasing the number of selected latents from $k=1$ to $k^\star$ changes both detection and intervention only modestly across models.
The representation position has a substantially larger effect:
$\mathrm{SAE}_{\mathrm{pre}}$ yields lower detection than the corresponding filled-target variant on all \rev{ten} models.
For intervention, the effect is model-dependent; pre-target  selection provides no advantage on \pythia, \gpttwo, \rev{or \mistral}
but produces a pronounced increase on \llama\ \rev{and \deepseek}.
Thus, selecting features before the target \rev{trades lower detection for a model-dependent intervention gain.
We use the filled-target \MethodSAEkOne\ as the main reference, because it uses a single latent selected from the same representation as \caa.
}

\subsection{Task characteristics}
\label{app:task-characteristics}

\begin{figure}[!t]
    \centering
    \includegraphics[width=\linewidth]{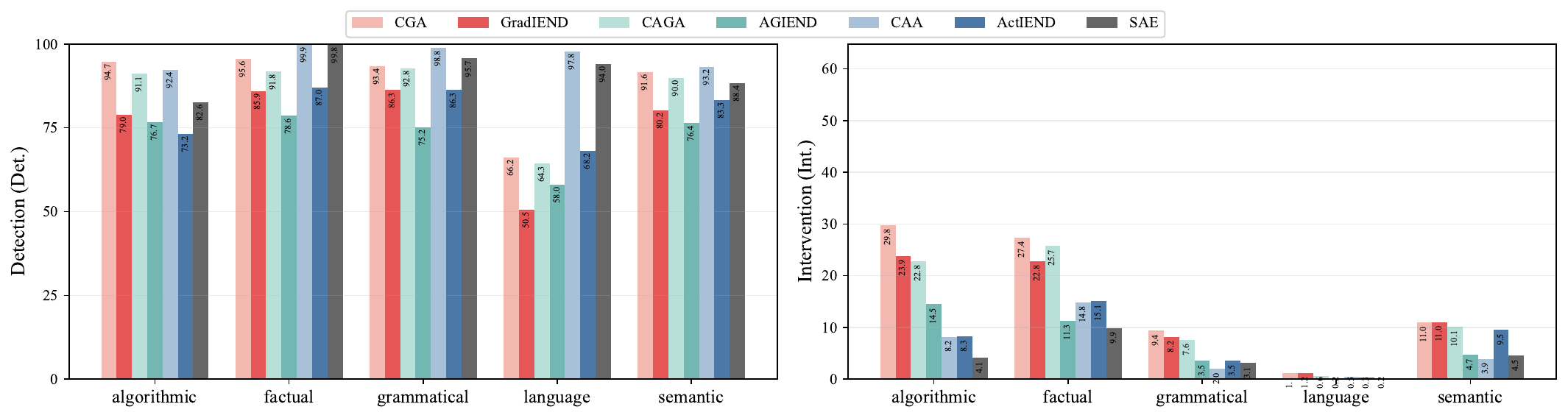}
    \caption{Performance by task family averaged across models.}
    \label{fig:task-family}
\end{figure}

Figure~\ref{fig:task-family} aggregates performance by task family (as introduced in Table~\ref{tab:task-summary}).
Detection is generally high for factual, grammatical, and semantic tasks, whereas the language task reveals substantially greater differences between
methods: \caa\ and \sae\ retain high detection, while several targeted methods drop considerably.
Intervention varies substantially across task families.
Factual \rev{and algorithmic} tasks produce the largest effects for most methods, whereas language tasks consistently yield only small interventions.
Semantic and \rev{grammatical} tasks generally fall in between.

\subsection{Pairwise and one-sided feature specifications}
\label{app:pairwise-onesided}

To compare pairwise and one-sided construction directly, we restrict the
analysis to tasks for which both specifications are available (excluding \taskMIBIOI, \taskRepetition, \taskFuncComp, \taskKeyValue, \taskInduction).
\begin{figure}[!t]
    \centering
    \includegraphics[width=\linewidth]{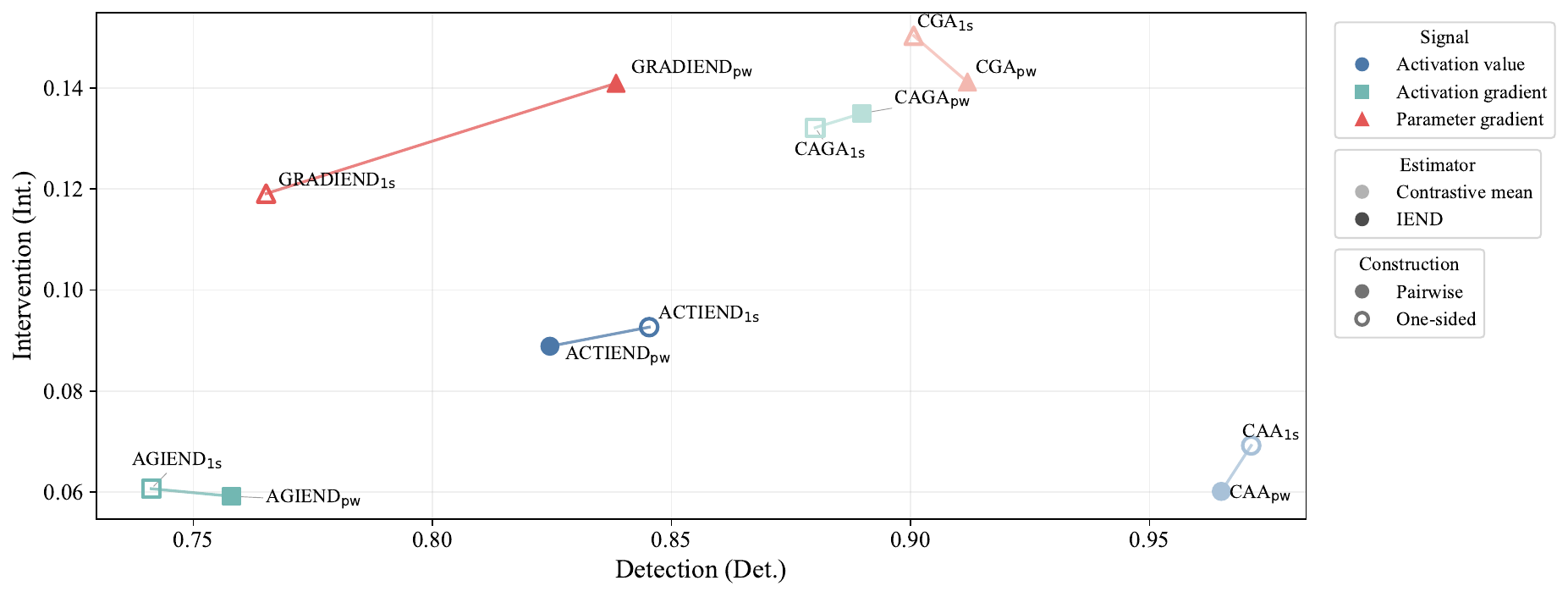}
    \caption{\textbf{Pairwise versus one-sided feature construction.}
    Model-balanced mean Detection and Intervention on tasks supporting both
    constructions.}
    \label{fig:construction-regime}
\end{figure}
Figure~\ref{fig:construction-regime} shows that one-sided construction %often achieves higher detection than pairwise construction, although the differences are modest for most methods.
\rev{achieves clearly higher detection only for \actiend\ and \agiend, and is otherwise similar.}
Intervention is often similar except for \gradiend, \rev{where it is lower}.
Overall, one-sided construction usually at least matches the performance of pairwise construction while requiring only $K$ rather than $\binom{K}{2}$ feature
models and remaining applicable when natural examples are available only for the target class.

\subsection{IEND optimization across signal spaces}
\label{app:iend-optimization}

Unlike the contrastive mean estimators, \iend\ requires optimization choices such as learning rate and number of training steps.
For \gradiend\ and \actiend, we ran a short learning-rate screen on \taskGender\ for each model and \rev{selected a rate}
%retained the largest rate 
that converged under the respective fixed step budget. 
The resulting rates were then fixed across all tasks.
%Because the three \iend\ variants operate on different signal spaces, we also experimented with signal-specific optimization schedules rather than assuming that the original \gradiend\ configuration transfers unchanged.

\paragraph{Decoder learning rate.}
The original \gradiend\ formulation \citep{drechsel2026gradiend} jointly optimizes encoder and decoder with a shared learning rate, which we initially also used for the other \iend\ variants.
For \actiend, however, increasing only the decoder learning rate while keeping the encoder rate fixed substantially improved intervention in experiments.
Analogous decoder-rate increases did not yield a clear benefit for \gradiend\ or \agiend.
We therefore use a larger decoder than encoder learning rate for \actiend, while \gradiend\ and \agiend\ retain shared rates.

\paragraph{Optimization steps.}
We also experimented with shorter optimization schedules together with corresponding learning-rate adjustments.
For \actiend, reducing the budget from 500 to 100 steps while increasing the decoder learning rate by a factor of five produced similar detection and intervention in development experiments.
Analogous shorter schedules for \gradiend\ and \agiend\ led to weaker intervention.
We therefore use 100 steps for \actiend\ and 500 for the gradient-based variants.

\paragraph{Scope.}
The experiments described in this section are small, single-model, single-task ablations intended to motivate the optimization choices rather than to establish general laws.
They support treating the encoder and decoder learning rates as signal-dependent \iend\ parameters, instead of assuming that the shared schedule of \gradiend\ transfers unchanged to other signal spaces.

\iffalse
\begin{table}[t]
\centering
\caption{Effect of a separate decoder learning rate (gpt2-small,
\texttt{gender\_en}, 500 steps). $\Delta P^{+}$ is the selected intervention effect for the female (F) and
male (M) class; $\|d_T\|$ is the final decoder norm. \actiend\ values are means
over three seeds; \gradiend\ and \agiend\ values come from the selected
checkpoint of a small ablation (\agiend\ was run for the male class only; its
``auto lift'' is an automatic decoder-rate increase). The separate rates differ
across methods (each is the value that was tested for that method), so the point
is the direction of the effect, not a tuned optimum.}
\label{tab:decoder-lr}
\small
\begin{tabular}{@{}llrrr@{}}
\toprule
method & decoder rate & $\Delta P^{+}$ (F) & $\Delta P^{+}$ (M) & $\|d_T\|$ \\
\midrule
\actiend & shared ($10^{-5}$) & 0.126 & 0.152 & 0.66 \\
\actiend & separate ($2.3\times10^{-3}$) & 0.215 & 0.275 & 32.82 \\
\midrule
\gradiend & shared ($10^{-4}$) & 0.294 & 0.336 & 4.29 \\
\gradiend & separate ($1.15\times10^{-2}$) & 0.279 & 0.342 & 21.94 \\
\midrule
\agiend & shared ($3\times10^{-3}$) & -- & 0.348 & 1.89 \\
\agiend & separate (auto lift) & -- & 0.348 & 1.90 \\
\bottomrule
\end{tabular}
\end{table}
\fi

\subsection{\rev{\taskEmotion\ feature recovery across models}}\label{app:emotion-recovery}

\begin{figure}[!t]
    \centering
    \includegraphics[width=\linewidth]{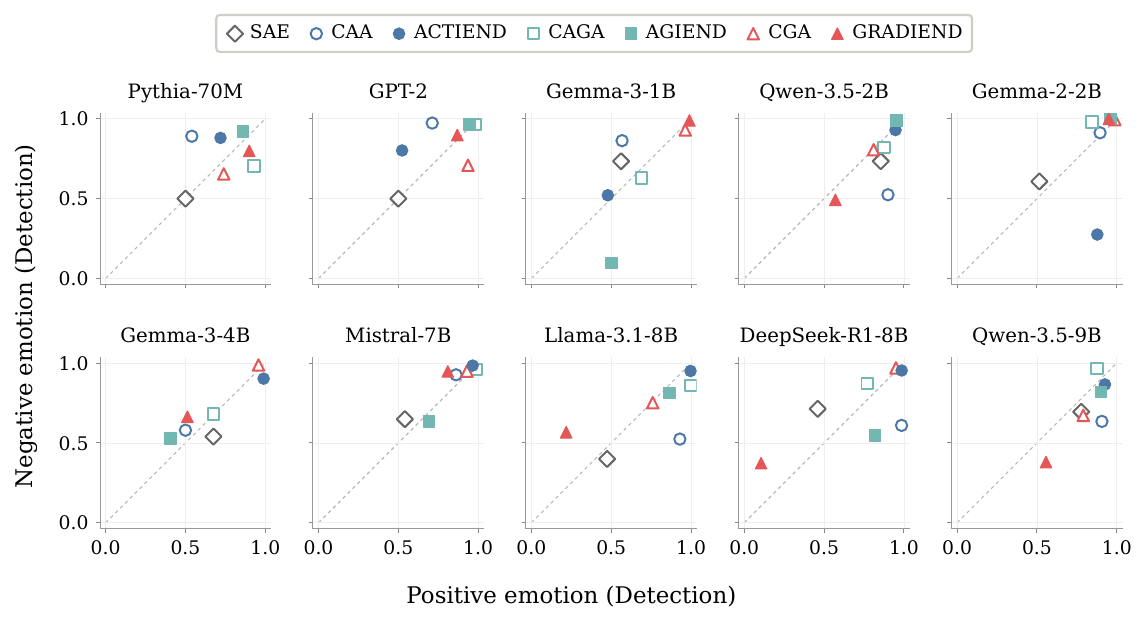}
    \caption{\rev{Detection of positive versus negative
    \taskEmotion\ across models and methods.
    The diagonal indicates equal performance for both polarities.}}
    \label{fig:emotion-polarity}
\end{figure}

\rev{
We examine \taskEmotion\ as a case study of a challenging semantic feature, analyzing detection of its positive and negative poles separately across all ten models.
Figure~\ref{fig:emotion-polarity} reveals substantial variation across models and methods, with some achieving strong detection for both polarities, while others show pronounced asymmetries.
Notably, even targeted methods can detect one polarity substantially better than the other.
Pretrained \glspl{sae} also vary considerably:
some recover features that detect both polarities well (e.g., \qwenS), whereas others perform substantially worse (e.g., \llama).
Overall, targeted methods generally achieve stronger detection than \glspl{sae} across models.
This supports our motivation for targeted feature learning: explicitly learning features for predefined concepts is more reliable than relying on their presence in an untargeted SAE dictionary.
}

\subsection{\rev{Additional baselines and hybrid estimator}}
\label{app:additional-baselines}

\rev{The main experiments systematically compare three model signals and two estimators across ten models and fifteen tasks.
%We complement this comparison with focused experiments addressing two additional questions: how the targeted methods compare with supervised baselines and whether combining different encoder and decoder IEND signals improves the detection-intervention trade-off.
We complement our controlled 3$\times$2 comparison
with two established supervised methods, logistic
probes and ReFT-r1, as well as a hybrid extension of \iend.
}
\rev{We restrict these ablations to a subset of five models (\pythia, \gpttwo, \gemma, \qwenS, and \mistral) and eight tasks (\taskGender, \taskEmotion,
\taskReligion, \taskPronNum, \taskRavelCountry, \taskLangID,
\taskKeyValue, and \taskMIBIOI).
The subset spans different model families and scales, as well as semantic, linguistic, factual, and algorithmic tasks.}

\paragraph{\rev{Logistic probe.}}
\rev{We include an $L_2$-regularized logistic regression classifier trained on activation values at the target position as a standard supervised detection baseline.
For intervention, we use its learned weight vector as an activation-steering direction, following \citet{wu2025axbench}.}

\paragraph{\rev{ReFT-r1.}}
\rev{We additionally evaluate ReFT-r1 \citep{wu2025axbench}, which learns a rank-one representation intervention by optimizing the language modeling objective toward target completions. 
We train separate directions for each feature class (i.e., one-sided configuration) and transformer layers, using validation detection to select the best representation.
%We apply the intervention at the target prediction position across all transformer layers.
%For detection, we use the projection onto its learned subspace.
As for \iend, we select a model-specific learning rate based on validation performance on \taskGender\ and keep it fixed across tasks.
%Intervention strengths are selected under the same language modeling preservation constraint as in the main experiments.
}

\rev{We adapt the training to our target-token prediction setting, supervising one target token per example for both feature and all valid next-token positions for neutral texts.
We also tested using single-token for neutral texts, but this resulted in worse detection performance.
For detection, we use the ReLU-activated feature score at the completed target position, rather than the sequence-level maximum used in AxBench, following our benchmark setting.
Interventions are evaluated under the same language modeling preservation constraint as our main experiments.
}

\paragraph{\rev{HYBRIEND.}}
\rev{
Motivated by the strong detection performance of activation values and intervention performance of parameter gradients, we introduce \textsc{HYBRIEND}, a hybrid extension of \iend\ that combines both signals.
Unlike the methods in our main experiments, which use the same signal space for IEND encoder and decoder, HYBRIEND uses activation values as encoder inputs and parameter gradient transitions as decoder targets.
This allows us to investigate whether the detection advantage of activation values and the intervention advantage of parameter gradients can be combined within a single feature representation.
}

\begin{figure}[!t]
    \centering
    \includegraphics[width=\linewidth]{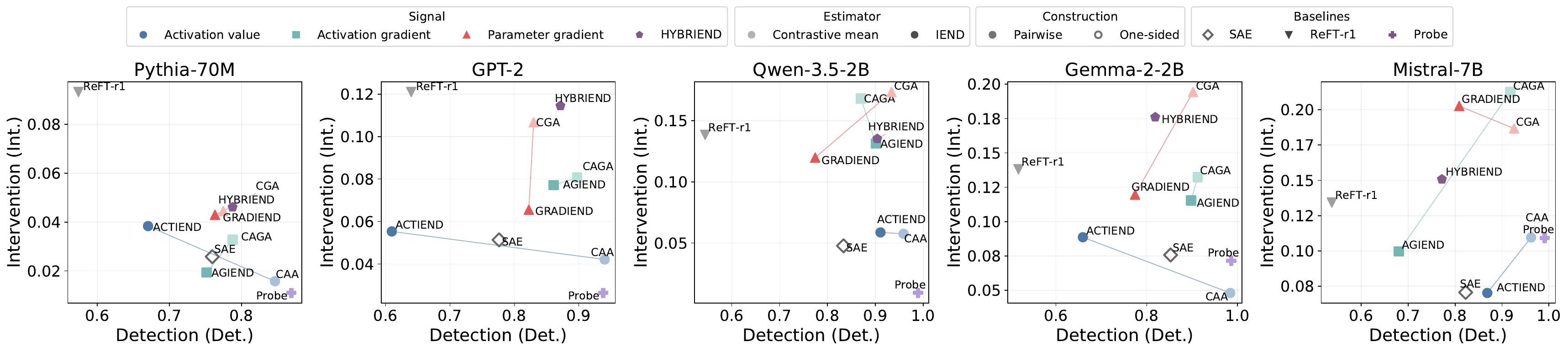}
\caption{\rev{Detection versus intervention performance with additional baselines and HYBRIEND.}}
    \label{fig:hybrid_headline_scatter_by_model}
\end{figure}

\paragraph{\rev{Results.}}
\rev{Figure~\ref{fig:hybrid_headline_scatter_by_model}
compares detection and intervention performance across the five models, while Table~\ref{tab:hybrid_ablation_grid} provides detailed per-task results.}
\rev{The logistic probe performs slightly better in detection than \caa, but its intervention scores vary a lot across tasks and models. Steering effects are negligible on many tasks.
ReFT-r1 also shows mixed performance across models and tasks, with strong interventions on some algorithmic tasks but substantially weaker results elsewhere.
Compared to AxBench, ReFT-r1 generally performs poorly on our more restrictive detection metric, potentially because its training objective does not explicitly account for rival classes.
Incorporating rival-class supervision could be a promising extension of the method.
Notably, \cga\ achieves stronger intervention performance overall than ReFT-r1, despite using a simple contrastive mean estimator.
}

\rev{
\textsc{HYBRIEND} generally outperforms the other \iend\ variants on both detection and intervention, but does not consistently outperform the contrastive mean methods.
These results suggest that combining different encoder and decoder signals is promising, although it does not fully resolve the detection-intervention trade-off.
However, this is only an initial hybrid formulation:
using localized activation inputs, as suggested by our representation-scope analysis (Section~\ref{sec:main-results}), could further improve detection.
}

\begin{table}[!pt]
\centering
\scriptsize
\setlength{\tabcolsep}{3pt}
\caption{\rev{Detection and intervention scores
(both multiplied by 100) by model, method, and task.}}
\label{tab:hybrid_ablation_grid}
\begin{tabular}{lcccccccc}
\toprule
Method & \taskGender & \taskEmotion & \taskPronNum & \taskRavelCountry & \taskLangID & \taskKeyValue & \taskReligion & \taskMIBIOI \\
\midrule
\multicolumn{9}{c}{\textbf{Pythia-70M}} \\
\midrule
CAA$_{\mathrm{pw}}$ & 100.0/6.5 & 71.6/0.3 & 99.8/3.4 & 100.0/0.5 & 96.7/0.0 & 45.0/0.9 & 91.0/1.1 & 73.3/0.0 \\
ACTIEND$_{\mathrm{pw}}$ & 99.3/21.3 & 80.0/1.0 & 87.4/3.3 & 99.6/1.1 & 63.2/0.1 & 13.0/0.0 & 85.0/3.3 & 8.8/0.5 \\
CAGA$_{\mathrm{pw}}$ & 100.0/5.4 & 81.7/0.8 & 100.0/3.8 & 84.7/1.0 & 58.7/0.1 & 49.0/11.9 & 91.7/2.2 & 64.5/1.0 \\
AGIEND$_{\mathrm{pw}}$ & 98.2/3.3 & 88.9/1.6 & 95.0/2.4 & 94.2/0.9 & 62.2/0.0 & 21.0/5.4 & 92.3/1.8 & 49.6/0.0 \\
CGA$_{\mathrm{pw}}$ & 94.2/13.4 & 69.8/1.4 & 94.9/6.5 & 82.5/1.8 & 56.6/0.1 & 54.0/9.5 & 84.7/1.9 & 83.0/1.1 \\
GRADIEND$_{\mathrm{pw}}$ & 96.2/8.0 & 84.9/10.7 & 96.8/5.9 & 87.6/2.7 & 36.3/1.3 & 49.1/0.8 & 89.2/3.8 & 70.5/1.1 \\
\MethodSAEkOne & 89.6/13.1 & 49.1/0.4 & 58.1/1.4 & 50.5/0.0 & 89.7/0.0 & 100.0/2.0 & 71.1/3.3 & 99.5/0.5 \\
Probe & 100.0/6.2 & 48.0/0.0 & 100.0/1.3 & 100.0/0.6 & 98.5/0.0 & 57.0/0.3 & 99.9/0.4 & 92.0/0.0 \\
ReFT-r1$_{\mathrm{1s}}$ & 77.5/9.9 & 27.0/-0.1 & 75.0/5.9 & 61.5/1.3 & 29.8/0.1 & 53.7/33.1 & 81.0/4.8 & 53.6/19.5 \\
HYBRIEND$_{\mathrm{pw}}$ & 100.0/17.9 & 87.5/3.1 & 94.0/7.9 & 100.0/1.7 & 69.0/0.2 & 12.0/0.4 & 94.8/4.3 & 72.9/1.3 \\
\midrule
\multicolumn{9}{c}{\textbf{GPT-2}} \\
\midrule
CAA$_{\mathrm{pw}}$ & 100.0/9.0 & 84.3/0.8 & 99.9/2.8 & 100.0/6.1 & 96.0/0.1 & 75.7/0.8 & 96.5/7.9 & 99.9/6.3 \\
ACTIEND$_{\mathrm{pw}}$ & 85.1/14.5 & 66.3/2.0 & 69.7/12.7 & 80.2/9.1 & 58.3/0.0 & 0.0/0.5 & 72.9/3.5 & 55.3/1.9 \\
CAGA$_{\mathrm{pw}}$ & 100.0/11.2 & 97.2/1.5 & 99.4/4.2 & 83.2/5.5 & 57.1/0.0 & 85.3/9.9 & 96.8/6.3 & 99.0/25.9 \\
AGIEND$_{\mathrm{pw}}$ & 100.0/20.5 & 95.5/2.4 & 95.6/3.9 & 94.5/2.8 & 67.6/0.0 & 37.0/0.0 & 98.8/3.0 & 100.0/29.1 \\
CGA$_{\mathrm{pw}}$ & 97.2/14.6 & 82.3/1.5 & 85.9/6.9 & 94.8/6.3 & 51.9/0.3 & 64.4/10.5 & 89.4/15.1 & 98.0/30.2 \\
GRADIEND$_{\mathrm{pw}}$ & 99.5/5.9 & 88.4/2.4 & 90.8/3.9 & 85.9/3.0 & 51.0/1.3 & 53.0/4.9 & 90.8/3.3 & 98.5/27.9 \\
\MethodSAEkOne & 92.3/19.4 & 43.8/0.5 & 62.1/5.6 & 58.5/7.1 & 95.7/0.0 & 100.0/1.0 & 70.1/3.5 & 98.7/4.0 \\
Probe & 100.0/4.4 & 89.4/0.1 & 100.0/3.3 & 100.0/3.4 & 96.3/0.0 & 65.0/0.8 & 100.0/1.7 & 99.7/7.4 \\
ReFT-r1$_{\mathrm{1s}}$ & 78.8/13.9 & 51.6/0.2 & 84.1/9.7 & 57.4/4.3 & 50.9/0.1 & 61.8/23.9 & 65.4/11.1 & 62.0/33.7 \\
HYBRIEND$_{\mathrm{pw}}$ & 100.0/16.2 & 98.0/3.9 & 63.5/1.2 & 99.5/9.9 & 82.3/0.4 & 58.9/27.3 & 98.5/3.3 & 96.5/29.2 \\
\midrule
\multicolumn{9}{c}{\textbf{Qwen-3.5-2B}} \\
\midrule
CAA$_{\mathrm{pw}}$ & 100.0/2.0 & 71.2/0.8 & 100.0/1.1 & 100.0/8.0 & 98.3/0.3 & 99.8/12.1 & 97.9/1.8 & 100.0/19.8 \\
ACTIEND$_{\mathrm{pw}}$ & 99.1/20.7 & 93.9/1.9 & 94.1/5.4 & 100.0/4.3 & 57.7/0.1 & 92.5/2.7 & 91.7/3.6 & 99.7/8.2 \\
CAGA$_{\mathrm{pw}}$ & 88.7/23.4 & 84.8/2.5 & 92.0/8.0 & 82.0/17.4 & 59.8/0.7 & 97.5/32.3 & 92.2/10.3 & 99.0/40.0 \\
AGIEND$_{\mathrm{pw}}$ & 98.8/20.9 & 97.2/2.6 & 91.0/4.7 & 93.6/9.2 & 58.0/0.3 & 97.0/22.3 & 86.5/3.2 & 99.0/41.9 \\
CGA$_{\mathrm{pw}}$ & 100.0/12.8 & 80.8/3.4 & 97.3/5.5 & 96.1/21.3 & 92.0/1.4 & 95.3/39.0 & 86.1/6.4 & 99.5/49.0 \\
GRADIEND$_{\mathrm{pw}}$ & 100.0/34.0 & 53.2/0.0 & 69.1/3.2 & 97.9/14.3 & 41.3/0.1 & 84.5/5.1 & 74.1/2.8 & 99.5/36.4 \\
\MethodSAEkOne & 97.9/20.1 & 43.6/0.2 & 65.4/2.9 & 80.4/5.0 & 98.2/0.2 & 100.0/0.8 & 81.5/4.8 & 99.9/4.2 \\
Probe & 100.0/0.9 & 92.1/0.2 & 100.0/0.7 & 100.0/0.3 & 99.4/0.0 & 100.0/4.4 & 100.0/0.2 & 100.0/0.5 \\
ReFT-r1$_{\mathrm{1s}}$ & 69.4/5.4 & 36.5/-1.0 & 73.4/7.8 & 59.2/1.9 & 51.8/1.1 & 30.0/45.2 & 71.8/21.2 & 43.5/29.1 \\
HYBRIEND$_{\mathrm{pw}}$ & 98.5/30.1 & 89.0/2.5 & 92.8/5.6 & 100.0/17.6 & 47.5/0.0 & 98.0/18.6 & 98.0/5.8 & 99.5/28.0 \\
\midrule
\multicolumn{9}{c}{\textbf{Gemma-2-2B}} \\
\midrule
CAA$_{\mathrm{pw}}$ & 100.0/2.2 & 90.4/0.5 & 100.0/2.0 & 100.0/15.8 & 98.7/0.4 & 99.6/0.3 & 98.5/3.7 & 100.0/13.8 \\
ACTIEND$_{\mathrm{pw}}$ & 74.3/39.6 & 57.7/1.4 & 60.3/2.7 & 74.1/15.4 & 37.9/0.2 & 75.0/0.3 & 66.3/4.5 & 81.9/6.9 \\
CAGA$_{\mathrm{pw}}$ & 100.0/7.7 & 91.2/2.6 & 89.3/6.9 & 100.0/19.9 & 65.2/0.4 & 96.0/34.7 & 89.9/15.1 & 98.1/18.5 \\
AGIEND$_{\mathrm{pw}}$ & 99.5/7.5 & 98.0/4.6 & 99.4/6.6 & 80.1/19.1 & 46.3/0.2 & 98.0/2.7 & 97.3/12.5 & 99.5/39.1 \\
CGA$_{\mathrm{pw}}$ & 99.8/43.9 & 99.1/3.5 & 96.3/4.5 & 81.2/20.5 & 53.7/1.0 & 99.9/42.1 & 91.6/8.8 & 100.0/31.2 \\
GRADIEND$_{\mathrm{pw}}$ & 84.4/0.1 & 97.5/5.1 & 96.3/11.1 & 80.2/25.4 & 44.9/0.1 & 25.0/1.4 & 91.1/19.5 & 100.0/33.1 \\
\MethodSAEkOne & 98.4/28.0 & 59.5/0.9 & 71.5/7.6 & 74.6/9.5 & 97.9/0.2 & 100.0/1.6 & 80.3/9.1 & 99.8/3.6 \\
Probe & 100.0/1.2 & 90.7/0.0 & 100.0/0.6 & 100.0/10.5 & 98.5/0.0 & 100.0/43.6 & 100.0/0.8 & 100.0/0.6 \\
ReFT-r1$_{\mathrm{1s}}$ & 69.2/5.5 & 19.9/-1.2 & 74.7/8.6 & 37.5/1.1 & 56.2/2.7 & 27.0/40.6 & 46.5/24.4 & 83.0/28.7 \\
HYBRIEND$_{\mathrm{pw}}$ & 15.5/8.4 & 98.0/5.1 & 98.8/14.0 & 99.0/15.0 & 47.2/0.3 & 97.0/42.3 & 99.5/21.8 & 100.0/34.0 \\
\midrule
\multicolumn{9}{c}{\textbf{Mistral-7B}} \\
\midrule
CAA$_{\mathrm{pw}}$ & 99.7/37.6 & 89.4/1.2 & 94.3/4.3 & 99.9/20.1 & 91.4/0.4 & 100.0/17.5 & 94.9/5.9 & 100.0/0.6 \\
ACTIEND$_{\mathrm{pw}}$ & 46.6/2.4 & 97.5/1.1 & 97.4/2.8 & 98.0/18.0 & 62.3/0.3 & 96.5/7.9 & 96.8/13.4 & 100.0/10.3 \\
CAGA$_{\mathrm{pw}}$ & 100.0/25.4 & 97.5/3.0 & 98.2/5.4 & 94.8/33.3 & 52.0/0.5 & 97.0/46.3 & 95.6/15.4 & 99.0/40.6 \\
AGIEND$_{\mathrm{pw}}$ & 81.8/3.5 & 66.4/0.9 & 56.3/0.5 & 54.7/2.4 & 41.6/0.1 & 98.0/32.3 & 54.4/5.9 & 91.0/34.1 \\
CGA$_{\mathrm{pw}}$ & 100.0/24.9 & 94.0/2.3 & 98.6/5.8 & 97.9/16.0 & 58.7/1.2 & 100.0/45.6 & 91.5/11.2 & 100.0/42.3 \\
GRADIEND$_{\mathrm{pw}}$ & 100.0/43.3 & 87.9/1.8 & 57.2/0.7 & 77.5/17.3 & 42.5/0.3 & 98.9/46.3 & 85.6/12.3 & 97.5/40.1 \\
\MethodSAEkOne & 97.4/29.7 & 51.6/0.7 & 64.9/5.1 & 83.2/10.5 & 91.7/0.3 & 100.0/2.3 & 69.7/7.0 & 99.7/1.0 \\
Probe & 100.0/23.6 & 93.9/0.0 & 100.0/2.5 & 99.9/43.0 & 99.2/0.0 & 100.0/4.6 & 100.0/11.6 & 99.9/2.1 \\
ReFT-r1$_{\mathrm{1s}}$ & 61.3/4.6 & 32.0/-0.3 & 63.2/6.5 & 37.7/0.4 & 50.7/2.6 & 73.0/42.0 & 44.1/11.9 & 67.9/39.7 \\
HYBRIEND$_{\mathrm{pw}}$ & 16.5/4.9 & 82.0/1.7 & 88.0/6.0 & 99.5/8.3 & 40.2/0.6 & 97.0/46.3 & 95.3/12.9 & 99.0/39.9 \\

\bottomrule
\end{tabular}
\end{table}

\subsection{\rev{Alignment between contrastive mean and IEND}}
\label{app:direction-alignment}

\rev{We additionally investigate whether \iend\ learns directions similar to their contrastive mean counterparts and how this similarity relates to detection and intervention.
We conduct this analysis on the same subset of models and tasks.
For each model-task-feature combination, we compute the cosine similarity between the \iend\ decoder direction and the corresponding contrastive-mean direction in the same signal space
.}

\begin{figure}[!t]
    \centering
    \includegraphics[width=\linewidth]{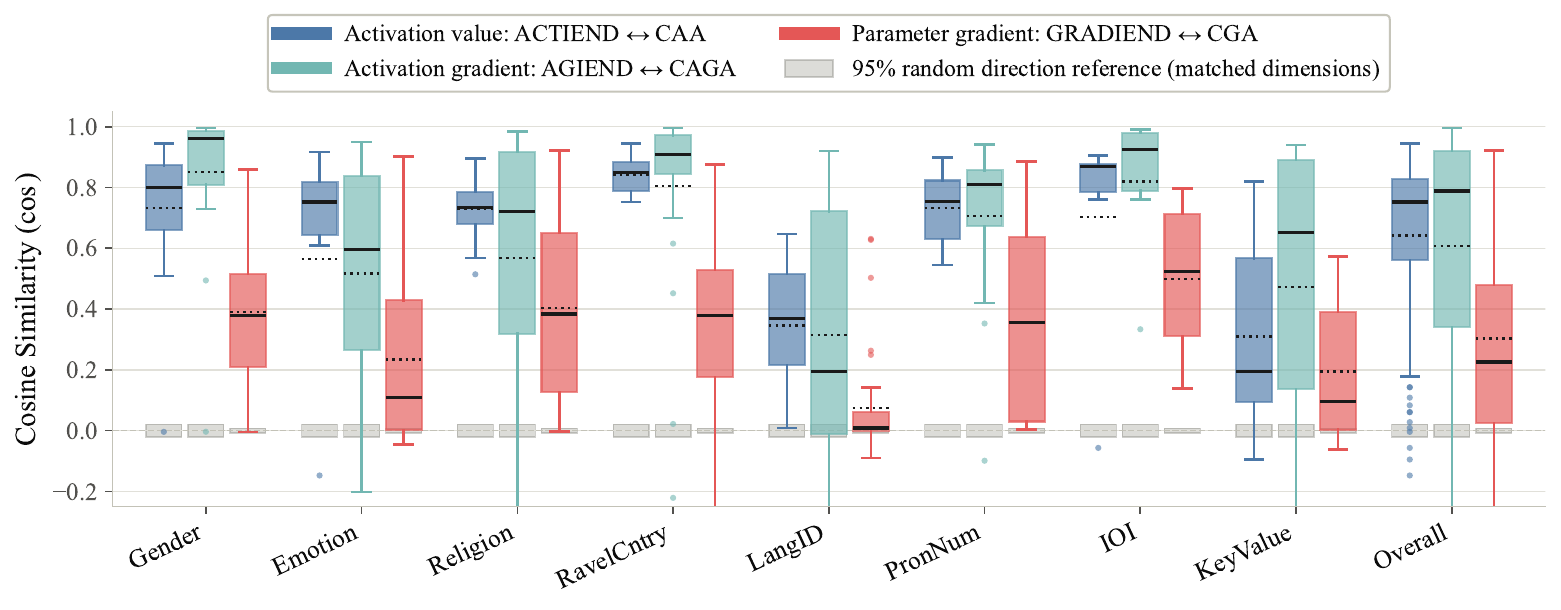}
   \caption{\rev{Cosine similarity between \iend\ decoder directions and their contrastive mean counterparts. 
   Gray bands indicate the central 95\% range expected for independent random directions with matched dimensionality.}}
    \label{fig:ablation_cosine_by_task}
\end{figure}

\begin{figure}[!t]
    \centering
    \includegraphics[width=\linewidth]{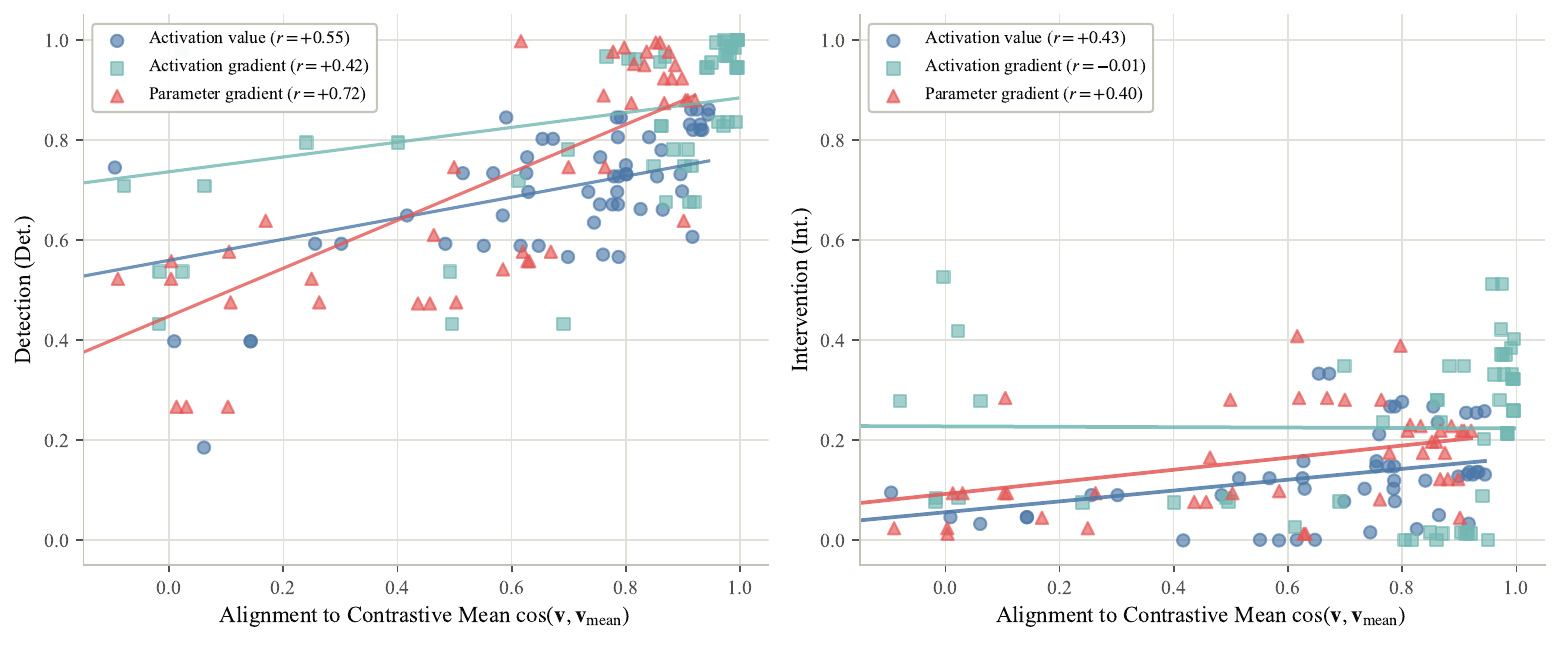}
    \caption{\rev{Relationship between direction alignment and \iend\ detection (left) and intervention (right) performance across signal spaces.}}
    \label{fig:ablation_cosine_performance_correlation}
\end{figure}

\rev{Figure~\ref{fig:ablation_cosine_by_task} shows substantial variation across tasks and signals.
Activation value and activation gradient directions generally exhibit higher alignment than parameter gradient directions.
However, the interpretation of these values depends on dimensionality, as random directions become increasingly orthogonal in higher-dimensional spaces. 
Copared to the dimension-matched random direction reference (gray bands), all three signals generally exhibit substantial alignment, except for parameter gradients on \taskLangID.
}

\rev{Figure~\ref{fig:ablation_cosine_performance_correlation} further examines the relationship between direction alignment and \iend\ performance.
Higher alignment is generally associated with better detection, while the relationship with intervention is positive for activation values and parameter gradients but close to zero for activation gradients.
Hence, alignment with the contrastive mean alone does not determine \iend\ performance.}

\subsection{Complete task-level results}
\label{app:task-level-results}

\begin{figure}[!tp]
    \centering
    \includegraphics[width=\linewidth]{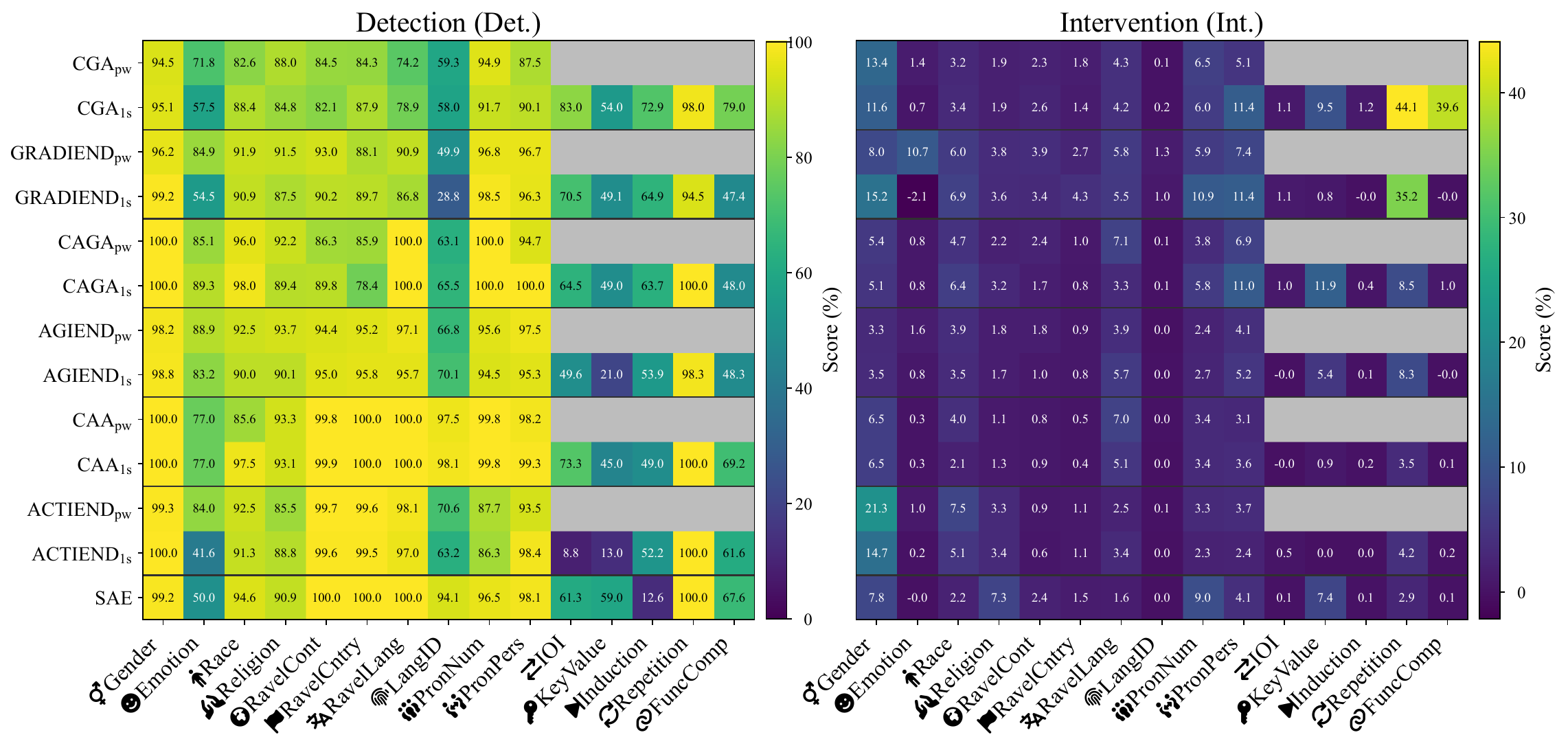}
    \caption{\pythia\ detection and intervention scores per task.}
    \label{fig:task-heatmap-pythia}
\end{figure}

\begin{figure}[!tp]
    \centering
    \includegraphics[width=\linewidth]{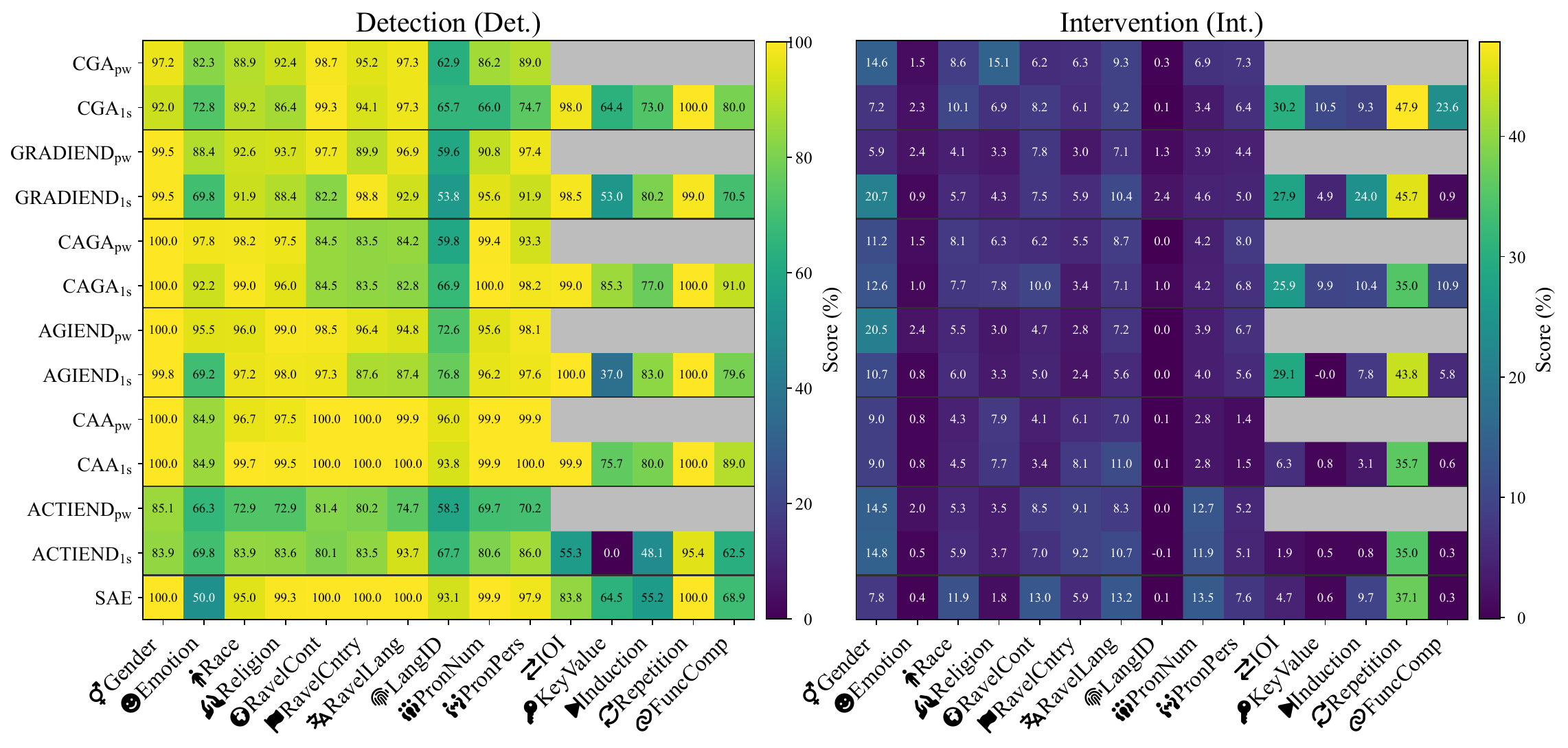}
    \caption{\gpttwo\ detection and intervention scores per task.}
    \label{fig:task-heatmap-gpt2}
\end{figure}

\begin{figure}[!tp]
    \centering
    \includegraphics[width=\linewidth]{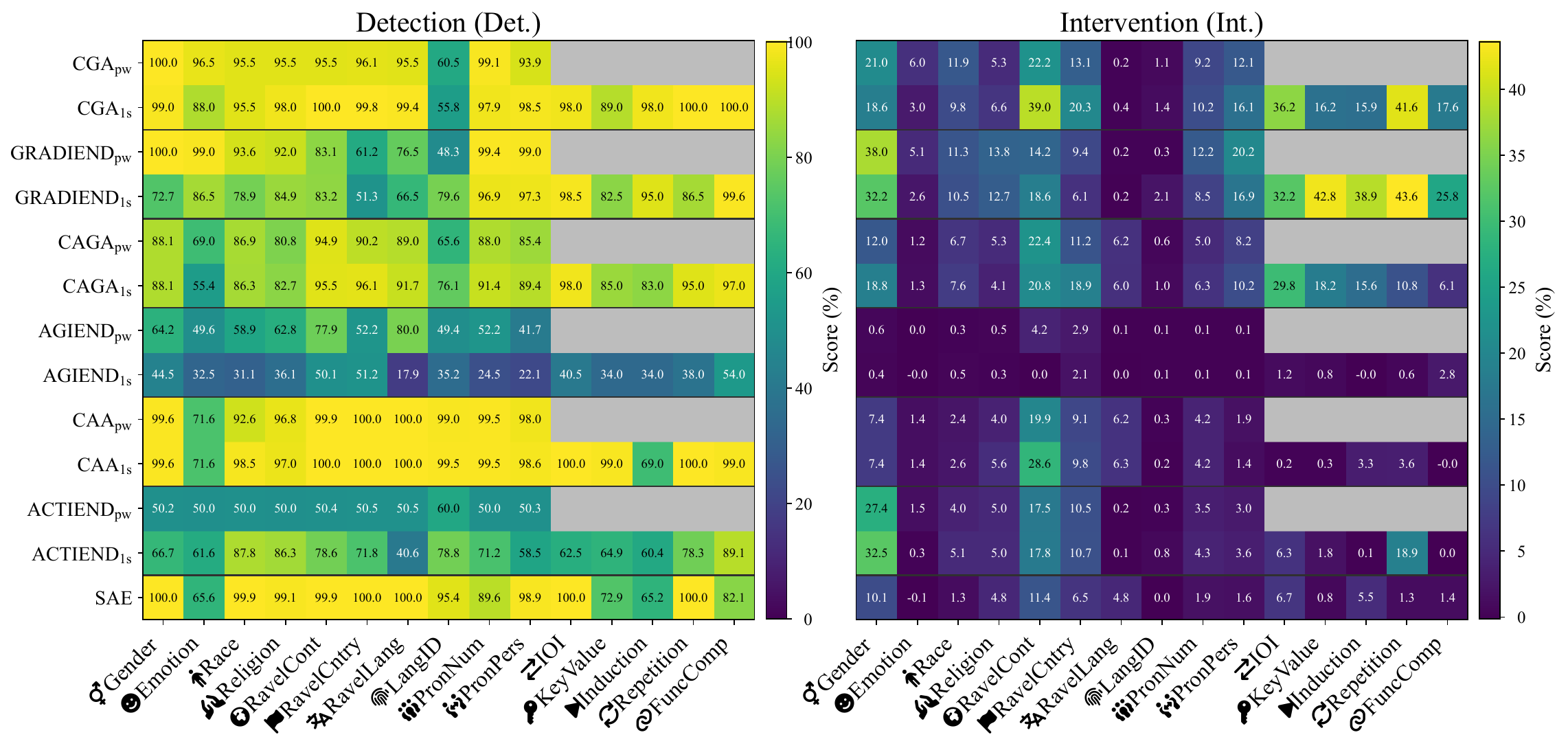}
    \caption{\rev{\gemmaS\ detection and intervention scores per task.}}
    \label{fig:task-heatmap-gemma3}
\end{figure}

\begin{figure}[!tp]
    \centering
    \includegraphics[width=\linewidth]{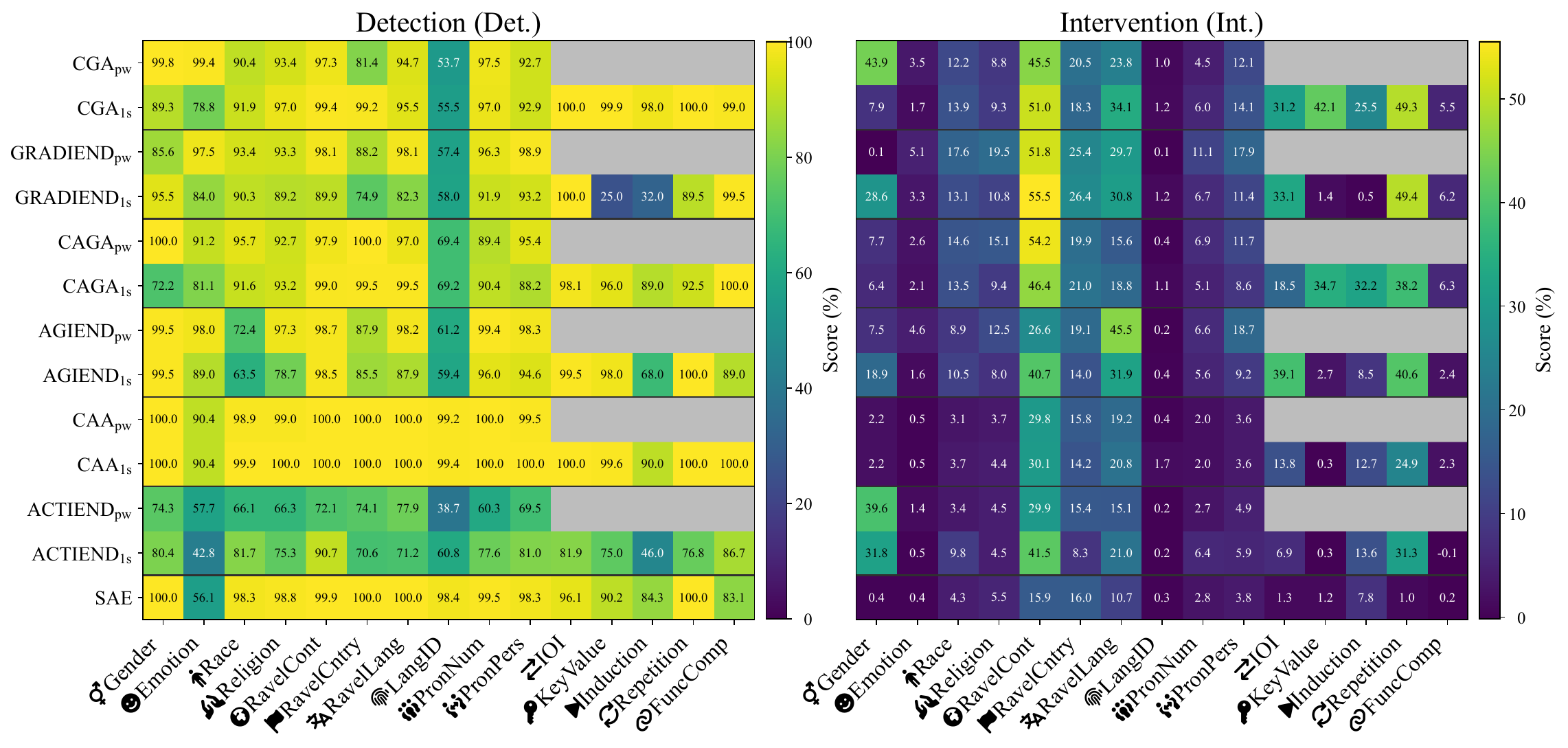}
    \caption{\rev{\gemma\ detection and intervention scores per task.}}
    \label{fig:task-heatmap-gemma}
\end{figure}

\begin{figure}[!tp]
    \centering
    \includegraphics[width=\linewidth]{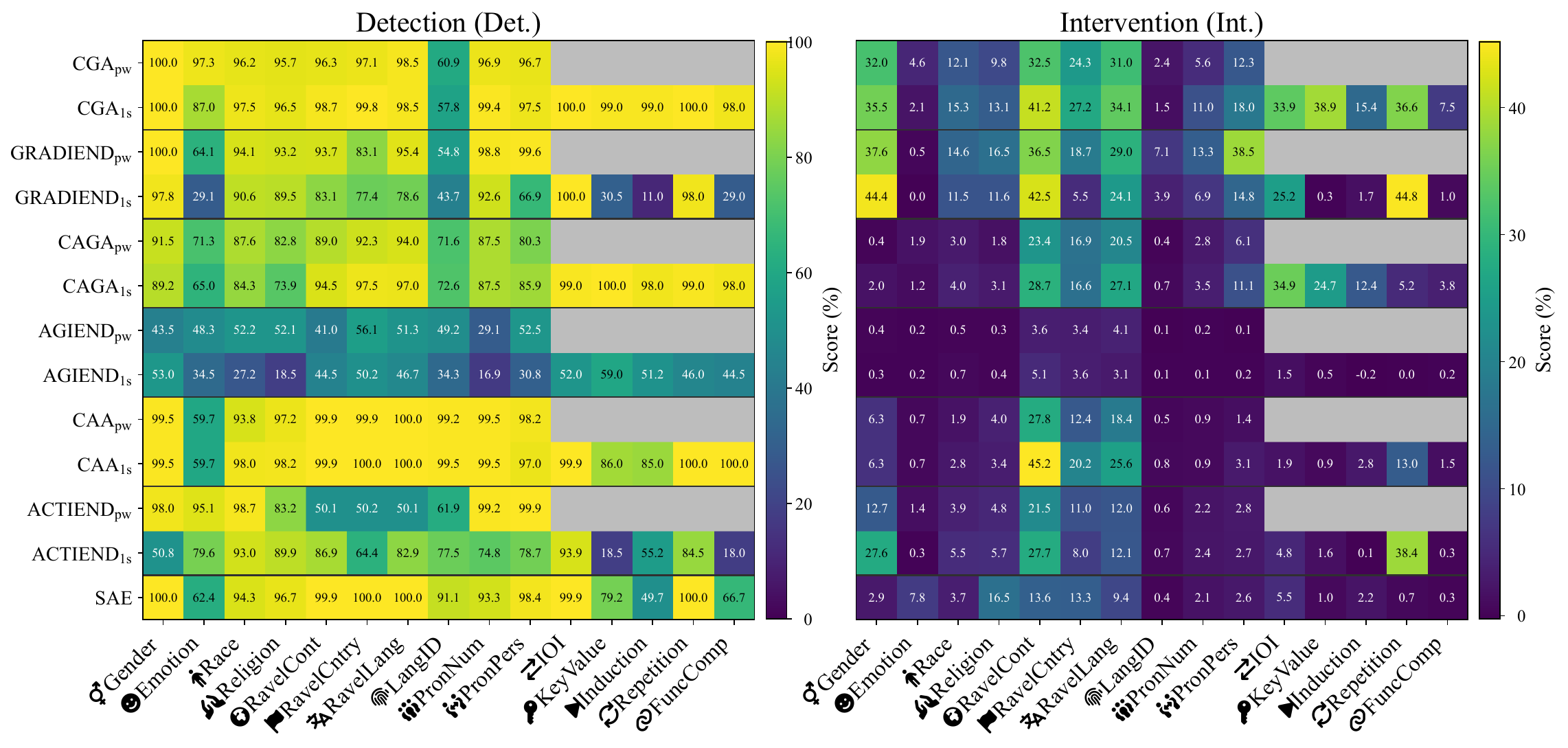}
    \caption{\rev{\gemmaM\ detection and intervention scores per task.}}
    \label{fig:task-heatmap-gemma3-4b}
\end{figure}

\begin{figure}[!tp]
    \centering
    \includegraphics[width=\linewidth]{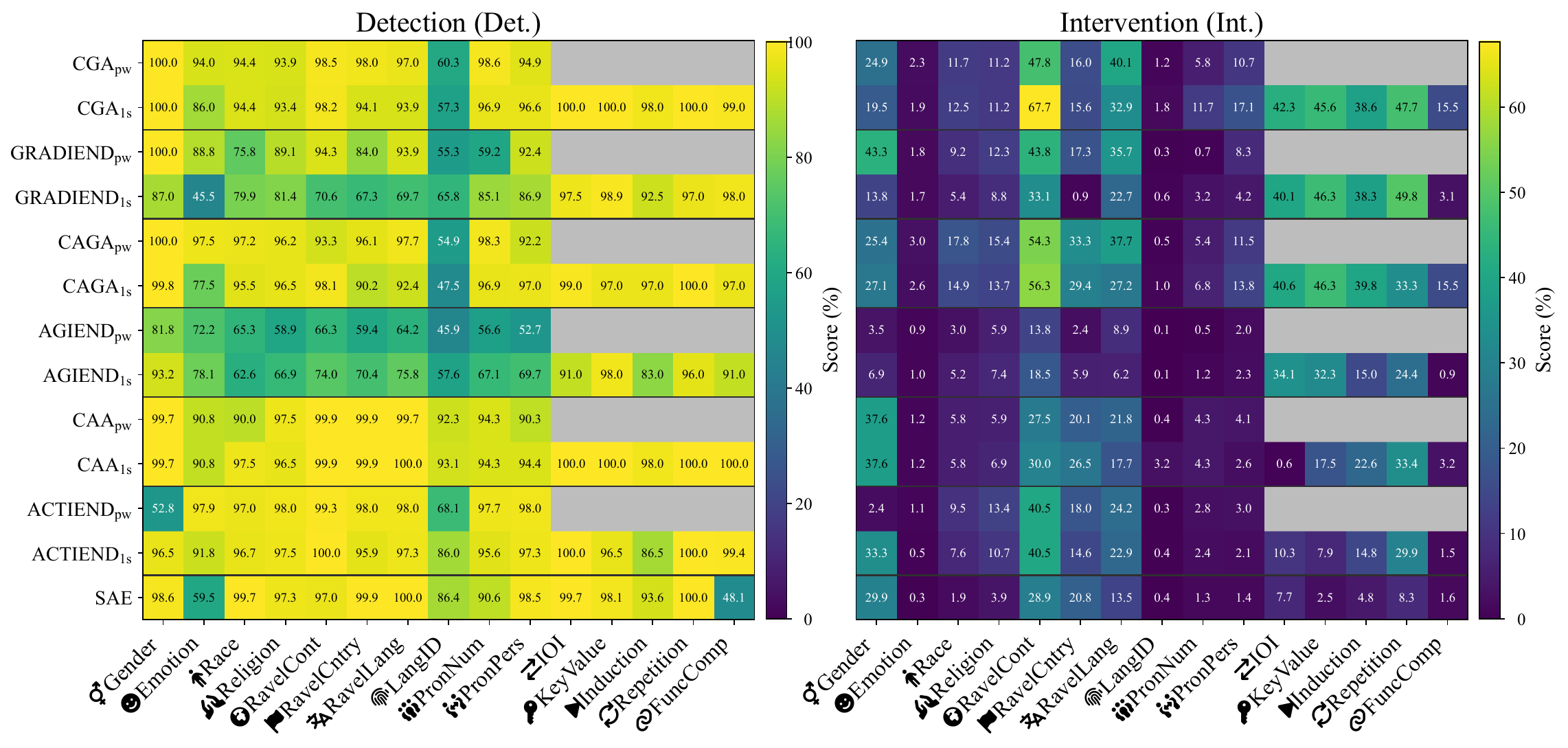}
    \caption{\rev{\mistral\ detection and intervention scores per task.}}
    \label{fig:task-heatmap-mistral}
\end{figure}

\begin{figure}[!tp]
    \centering
    \includegraphics[width=\linewidth]{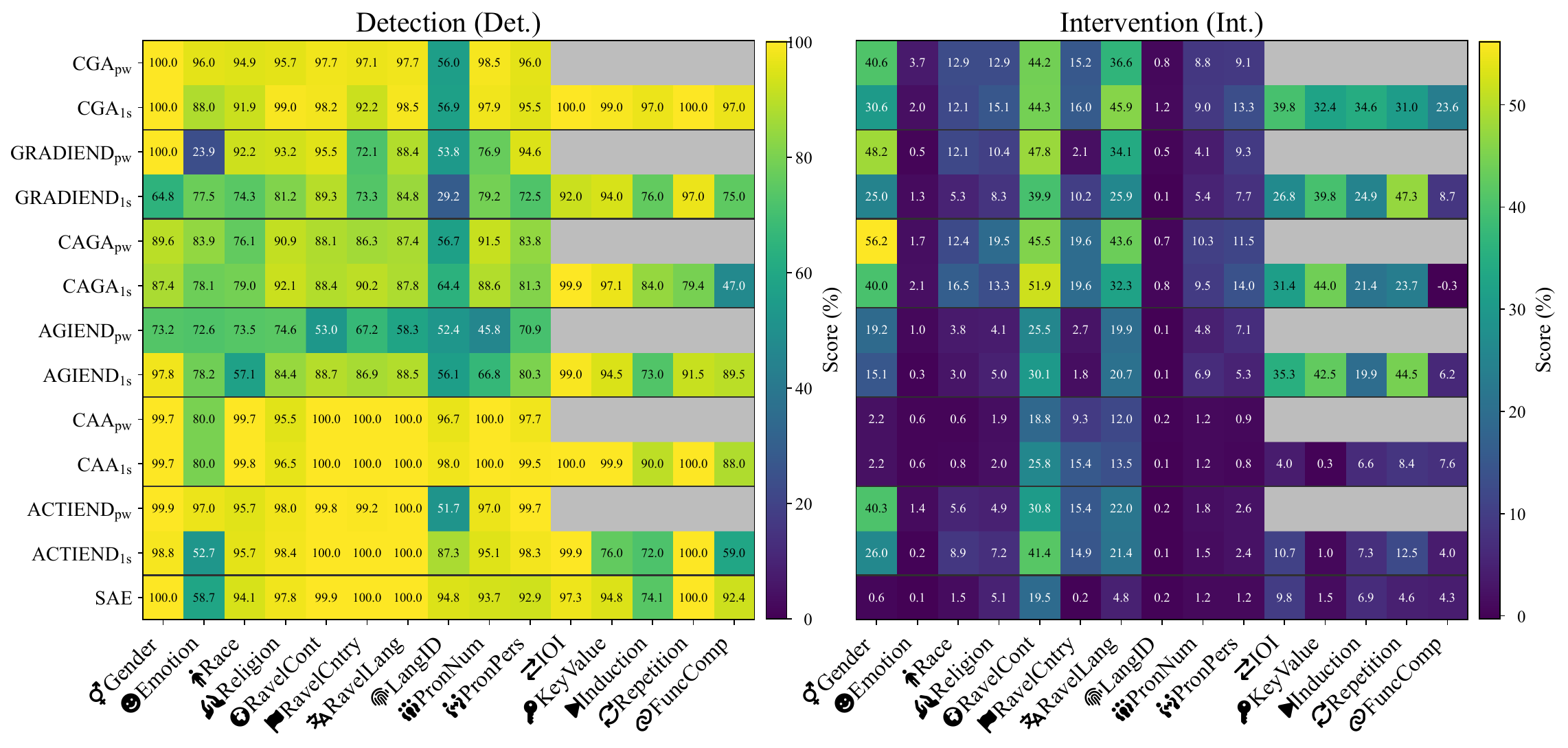}
    \caption{\rev{\deepseek\ detection and intervention scores per task.}}
    \label{fig:task-heatmap-deepseek}
\end{figure}

\begin{figure}[!tp]
    \centering
    \includegraphics[width=\linewidth]{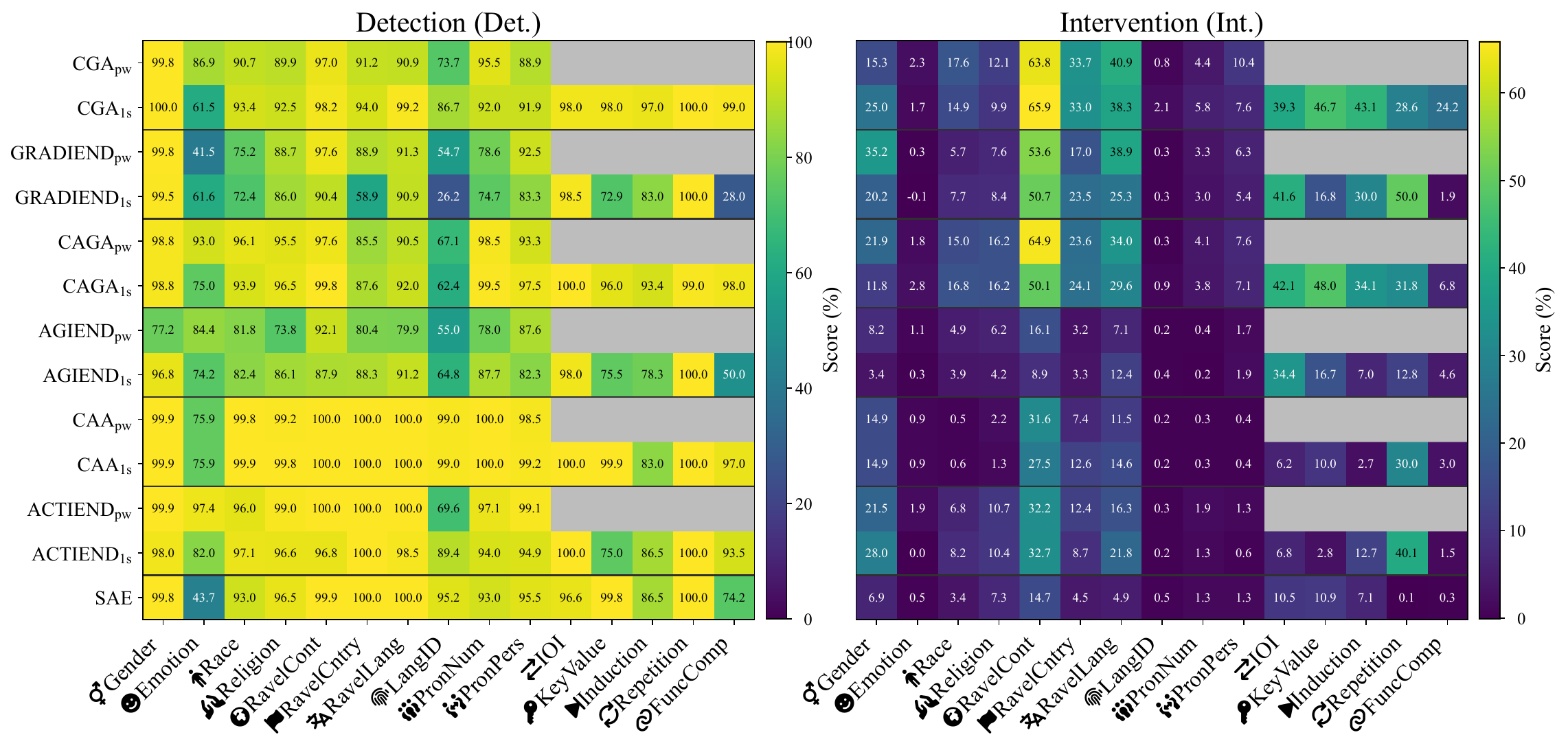}
    \caption{\llama\ detection and intervention scores per task.}
    \label{fig:task-heatmap-llama}
\end{figure}

For completeness, we report the task-level quantities underlying all aggregate
results in Tables~\ref{tab:summary-concatenated-detection-pythia70mdeduped}--\ref{tab:summary-concatenated-intervention-llama3.18b}. %, and further provide a task-averaged view on this in Table~\ref{tab:summary-methods}.
Further, we provide model-based views in Figures~\ref{fig:task-heatmap-pythia}-\ref{fig:task-heatmap-llama} of the averaged version in Figure~\ref{fig:task-heatmap}.

\begin{figure}[!tp]
    \centering
    \includegraphics[width=\linewidth]{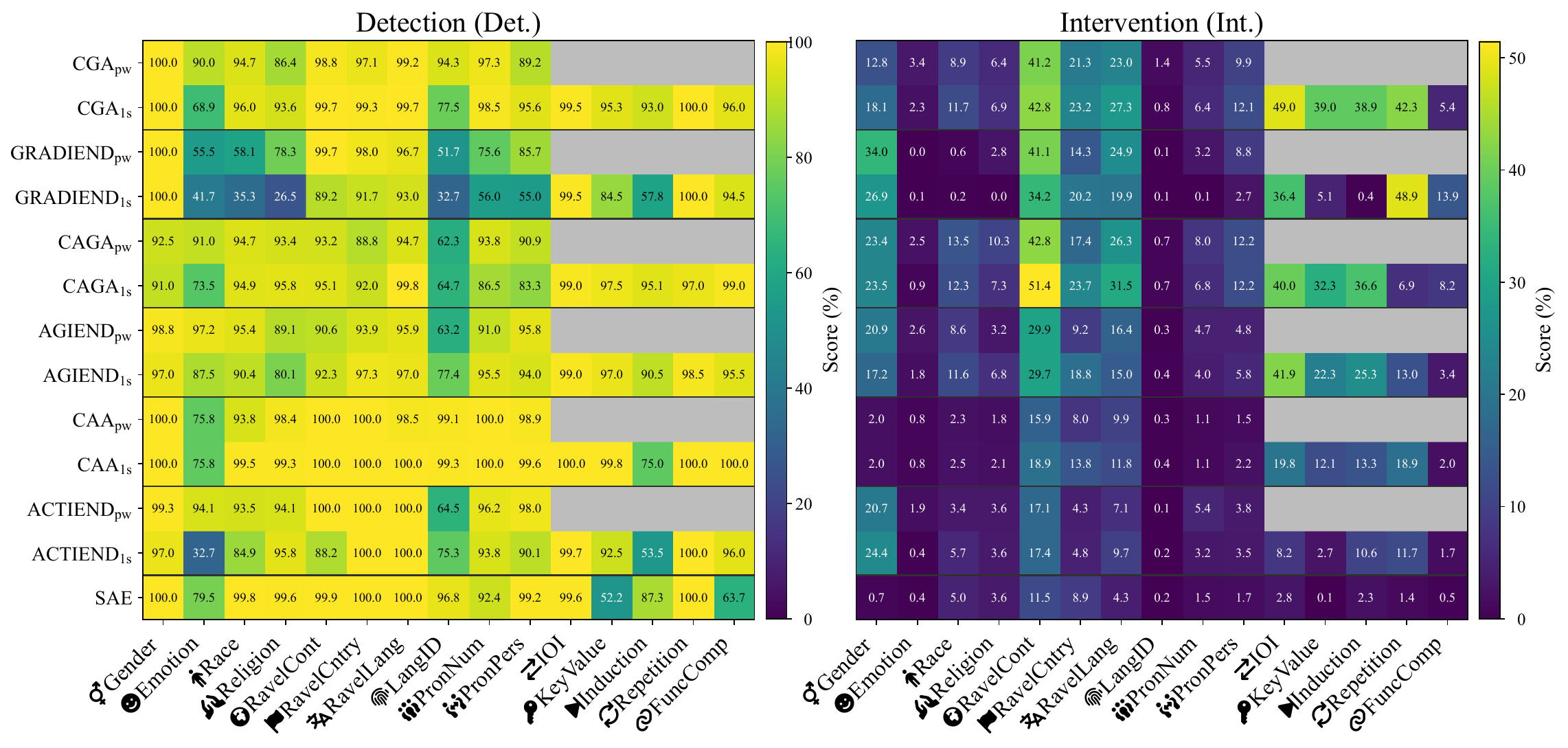}
    \caption{\rev{\qwenS\ detection and intervention scores per task.}}
    \label{fig:task-heatmap-qwen}
\end{figure}
\begin{figure}[!tp]
    \centering
    \includegraphics[width=\linewidth]{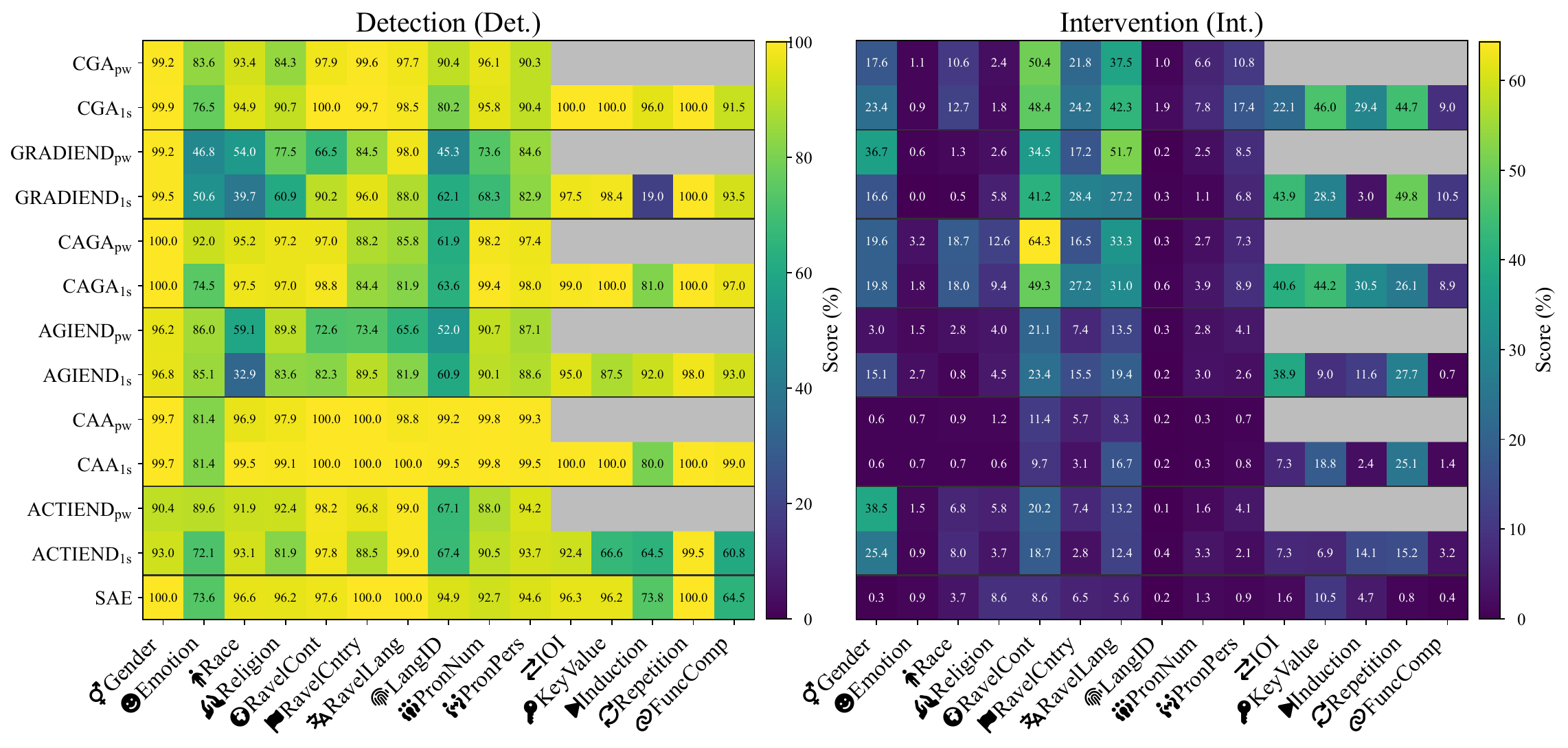}
    \caption{\rev{\qwenL\ detection and intervention scores per task.}}
    \label{fig:task-heatmap-qwen}
\end{figure}

\begin{table*}[p]
  \centering
  \tiny
  \setlength{\tabcolsep}{1pt}
  \renewcommand{\arraystretch}{0.8}
  \caption{Detection raw results for \pythia.}
  \label{tab:summary-concatenated-detection-pythia70mdeduped}
\begin{tabular}{l*{15}{c}}
    \toprule
    Method & \rotatebox{60}{\taskGender} & \rotatebox{60}{\taskEmotion} & \rotatebox{60}{\taskRace} & \rotatebox{60}{\taskReligion} & \rotatebox{60}{\taskRavelCont} & \rotatebox{60}{\taskRavelCountry} & \rotatebox{60}{\taskRavelLang} & \rotatebox{60}{\taskLangID} & \rotatebox{60}{\taskPronNum} & \rotatebox{60}{\taskPronPers} & \rotatebox{60}{\taskMIBIOI} & \rotatebox{60}{\taskKeyValue} & \rotatebox{60}{\taskInduction} & \rotatebox{60}{\taskRepetition} & \rotatebox{60}{\taskFuncComp} \\
    \midrule
    \multicolumn{16}{c}{AUC$_n$} \\
    \midrule
    CGA$_{\mathrm{pw}}$ & 0.966 & 0.783 & 0.866 & 0.919 & 0.945 & 0.945 & 0.836 & 0.593 & 0.981 & 0.902 & - & - & - & - & - \\
    CGA$_{\mathrm{1s}}$ & 0.951 & 0.687 & 0.912 & 0.916 & 0.890 & 0.889 & 0.789 & 0.604 & 0.977 & 0.901 & 0.882 & 0.737 & 0.729 & 0.990 & 0.990 \\
    \SummaryMethodRule{16}
    GRADIEND$_{\mathrm{pw}}$ & 0.995 & 0.902 & 0.919 & 0.928 & 0.930 & 0.919 & 0.932 & 0.499 & 0.988 & 0.993 & - & - & - & - & - \\
    GRADIEND$_{\mathrm{1s}}$ & 0.997 & 0.545 & 0.919 & 0.910 & 0.925 & 0.968 & 0.868 & 0.541 & 0.995 & 0.989 & 0.827 & 0.543 & 0.649 & 0.999 & 0.474 \\
    \SummarySignalRule{16}
    CAGA$_{\mathrm{pw}}$ & \textbf{1.000} & 0.851 & 0.967 & 0.966 & 0.966 & 0.859 & \textbf{1.000} & 0.774 & \textbf{1.000} & 0.977 & - & - & - & - & - \\
    CAGA$_{\mathrm{1s}}$ & \textbf{1.000} & 0.893 & \textbf{0.994} & 0.974 & 0.957 & 0.927 & \textbf{1.000} & 0.739 & \textbf{1.000} & \textbf{1.000} & 0.963 & 0.636 & 0.656 & \textbf{1.000} & 0.892 \\
\bottomrule
  \end{tabular}
\end{table*}

\begin{table*}[p]
  \centering
  \tiny
  \setlength{\tabcolsep}{1pt}
  \renewcommand{\arraystretch}{0.8}
  \caption{Intervention raw results for \pythia.}
  \label{tab:summary-concatenated-intervention-pythia70mdeduped}
  \begin{tabular}{l*{15}{c}}
    \toprule
    Method & \rotatebox{60}{\taskGender} & \rotatebox{60}{\taskEmotion} & \rotatebox{60}{\taskRace} & \rotatebox{60}{\taskReligion} & \rotatebox{60}{\taskRavelCont} & \rotatebox{60}{\taskRavelCountry} & \rotatebox{60}{\taskRavelLang} & \rotatebox{60}{\taskLangID} & \rotatebox{60}{\taskPronNum} & \rotatebox{60}{\taskPronPers} & \rotatebox{60}{\taskMIBIOI} & \rotatebox{60}{\taskKeyValue} & \rotatebox{60}{\taskInduction} & \rotatebox{60}{\taskRepetition} & \rotatebox{60}{\taskFuncComp} \\
    \midrule
 \multicolumn{16}{c}{$\Delta P^{+}$} \\
    \midrule
    CGA$_{\mathrm{pw}}$ & 0.197 & 0.010 & 0.018 & 0.012 & 0.028 & 0.025 & 0.050 & 0.000 & 0.073 & 0.045 & - & - & - & - & - \\
    CGA$_{\mathrm{1s}}$ & 0.165 & 0.001 & 0.025 & 0.009 & 0.035 & 0.015 & 0.050 & 0.000 & 0.076 & 0.147 & 0.012 & 0.083 & \textbf{0.019} & 0.106 & \textbf{0.787} \\
    \SummaryMethodRule{16}
    GRADIEND$_{\mathrm{pw}}$ & 0.093 & \textbf{0.171} & 0.042 & 0.032 & \textbf{0.056} & 0.039 & 0.072 & \textbf{0.002} & 0.064 & 0.074 & - & - & - & - & - \\
    GRADIEND$_{\mathrm{1s}}$ & 0.213 & 0.001 & 0.072 & 0.031 & 0.042 & \textbf{0.069} & 0.063 & 0.000 & \textbf{0.160} & \textbf{0.148} & \textbf{0.014} & -0.000 & -0.000 & \textbf{0.145} & 0.000 \\
    \SummarySignalRule{16}
    CAGA$_{\mathrm{pw}}$ & 0.052 & 0.001 & 0.030 & 0.013 & 0.027 & 0.008 & \textbf{0.093} & -0.000 & 0.033 & 0.076 & - & - & - & - & - \\
    CAGA$_{\mathrm{1s}}$ & 0.042 & 0.002 & 0.034 & 0.031 & 0.012 & 0.003 & 0.044 & 0.000 & 0.079 & 0.142 & 0.014 & \textbf{0.135} & 0.005 & 0.092 & 0.015 \\
\bottomrule
  \end{tabular}
\end{table*}

\begin{table*}[p]
  \centering
  \tiny
  \setlength{\tabcolsep}{1pt}
  \renewcommand{\arraystretch}{0.8}
  \caption{Detection raw results for \gpttwo.}
  \label{tab:summary-concatenated-detection-gpt2small}
  \begin{tabular}{l*{15}{c}}
    \toprule
    Method & \rotatebox{60}{\taskGender} & \rotatebox{60}{\taskEmotion} & \rotatebox{60}{\taskRace} & \rotatebox{60}{\taskReligion} & \rotatebox{60}{\taskRavelCont} & \rotatebox{60}{\taskRavelCountry} & \rotatebox{60}{\taskRavelLang} & \rotatebox{60}{\taskLangID} & \rotatebox{60}{\taskPronNum} & \rotatebox{60}{\taskPronPers} & \rotatebox{60}{\taskMIBIOI} & \rotatebox{60}{\taskKeyValue} & \rotatebox{60}{\taskInduction} & \rotatebox{60}{\taskRepetition} & \rotatebox{60}{\taskFuncComp} \\
    \midrule

        \multicolumn{16}{c}{AUC$_n$} \\
    \midrule
    CGA$_{\mathrm{pw}}$ & 0.994 & 0.823 & 0.902 & 0.952 & 0.998 & 0.992 & 0.992 & 0.654 & 0.862 & 0.941 & - & - & - & - & - \\
    CGA$_{\mathrm{1s}}$ & 0.987 & 0.769 & 0.915 & 0.921 & 1.000 & 0.988 & 0.997 & 0.723 & 0.845 & 0.781 & 0.997 & 0.644 & 0.935 & \textbf{1.000} & 0.921 \\
    \SummaryMethodRule{16}
    GRADIEND$_{\mathrm{pw}}$ & 1.000 & 0.908 & 0.926 & 0.937 & 0.993 & 0.953 & 0.991 & 0.681 & 0.955 & 0.988 & - & - & - & - & - \\
    GRADIEND$_{\mathrm{1s}}$ & 1.000 & 0.744 & 0.923 & 0.907 & 0.900 & 0.997 & 0.992 & 0.616 & 0.973 & 0.961 & 0.997 & 0.746 & 0.998 & 0.997 & 0.705 \\
    \SummarySignalRule{16}
    CAGA$_{\mathrm{pw}}$ & \textbf{1.000} & \textbf{0.990} & 0.982 & 0.983 & 0.845 & 0.835 & 0.842 & 0.692 & 0.995 & 0.933 & - & - & - & - & - \\
    CAGA$_{\mathrm{1s}}$ & \textbf{1.000} & 0.922 & 0.995 & 0.974 & 0.845 & 0.835 & 0.828 & 0.706 & \textbf{1.000} & 0.993 & 1.000 & 0.965 & 0.953 & \textbf{1.000} & \textbf{1.000} \\
\bottomrule
  \end{tabular}
\end{table*}

\begin{table*}[p]
  \centering
  \tiny
  \setlength{\tabcolsep}{1pt}
  \renewcommand{\arraystretch}{0.8}
  \caption{Intervention raw results for \gpttwo.}
  \label{tab:summary-concatenated-intervention-gpt2small}
  \begin{tabular}{l*{15}{c}}
    \toprule
    Method & \rotatebox{60}{\taskGender} & \rotatebox{60}{\taskEmotion} & \rotatebox{60}{\taskRace} & \rotatebox{60}{\taskReligion} & \rotatebox{60}{\taskRavelCont} & \rotatebox{60}{\taskRavelCountry} & \rotatebox{60}{\taskRavelLang} & \rotatebox{60}{\taskLangID} & \rotatebox{60}{\taskPronNum} & \rotatebox{60}{\taskPronPers} & \rotatebox{60}{\taskMIBIOI} & \rotatebox{60}{\taskKeyValue} & \rotatebox{60}{\taskInduction} & \rotatebox{60}{\taskRepetition} & \rotatebox{60}{\taskFuncComp} \\
    \midrule

\multicolumn{16}{c}{$\Delta P^{+}$} \\
    \midrule
    CGA$_{\mathrm{pw}}$ & 0.148 & 0.008 & 0.074 & \textbf{0.214} & 0.084 & 0.051 & 0.101 & 0.000 & 0.066 & 0.060 & - & - & - & - & - \\
    CGA$_{\mathrm{1s}}$ & 0.093 & 0.002 & \textbf{0.104} & 0.049 & 0.117 & 0.031 & 0.103 & 0.000 & 0.024 & 0.031 & 0.179 & \textbf{0.197} & 0.131 & \textbf{0.314} & \textbf{0.464} \\
    \SummaryMethodRule{16}
    GRADIEND$_{\mathrm{pw}}$ & 0.055 & 0.009 & 0.019 & 0.017 & 0.106 & 0.030 & 0.094 & \textbf{0.000} & 0.030 & 0.022 & - & - & - & - & - \\
    GRADIEND$_{\mathrm{1s}}$ & \textbf{0.276} & 0.001 & 0.066 & 0.036 & 0.120 & 0.046 & \textbf{0.166} & 0.000 & 0.045 & 0.049 & \textbf{0.346} & 0.099 & \textbf{0.413} & 0.313 & 0.010 \\
    \SummarySignalRule{16}
    CAGA$_{\mathrm{pw}}$ & 0.121 & 0.003 & 0.052 & 0.055 & 0.089 & 0.051 & 0.112 & 0.000 & 0.047 & \textbf{0.066} & - & - & - & - & - \\
    CAGA$_{\mathrm{1s}}$ & 0.093 & 0.001 & 0.054 & 0.087 & 0.166 & 0.039 & 0.091 & -0.000 & 0.043 & 0.054 & 0.317 & 0.164 & 0.157 & 0.286 & 0.209 \\
\bottomrule
  \end{tabular}
\end{table*}

%%%%% Gemma-3-1B %%%%%%%%%%%

\begin{table*}[p]
  \centering
  \tiny
  \setlength{\tabcolsep}{1pt}
  \renewcommand{\arraystretch}{0.8}
  \caption{\rev{Detection raw results for \gemmaS.}}
  \label{tab:summary-concatenated-detection-gemma3}
  \begin{tabular}{l*{15}{c}}
    \toprule
    Method & \rotatebox{60}{\taskGender} & \rotatebox{60}{\taskEmotion} & \rotatebox{60}{\taskRace} & \rotatebox{60}{\taskReligion} & \rotatebox{60}{\taskRavelCont} & \rotatebox{60}{\taskRavelCountry} & \rotatebox{60}{\taskRavelLang} & \rotatebox{60}{\taskLangID} & \rotatebox{60}{\taskPronNum} & \rotatebox{60}{\taskPronPers} & \rotatebox{60}{\taskMIBIOI} & \rotatebox{60}{\taskKeyValue} & \rotatebox{60}{\taskInduction} & \rotatebox{60}{\taskRepetition} & \rotatebox{60}{\taskFuncComp} \\
    \midrule
   \multicolumn{16}{c}{AUC$_n$} \\
    \midrule
    CGA$_{\mathrm{pw}}$ & \textbf{1.000} & 0.982 & 0.981 & 0.984 & 0.999 & 0.999 & \textbf{1.000} & 0.655 & 0.991 & 0.986 & - & - & - & - & - \\
    CGA$_{\mathrm{1s}}$ & \textbf{1.000} & 0.880 & 0.984 & 0.994 & \textbf{1.000} & 1.000 & 1.000 & 0.772 & \textbf{1.000} & 0.997 & \textbf{1.000} & 0.990 & 1.000 & \textbf{1.000} & 1.000 \\
    \SummaryMethodRule{16}
    GRADIEND$_{\mathrm{pw}}$ & \textbf{1.000} & \textbf{1.000} & 0.936 & 0.944 & 0.831 & 0.742 & 0.871 & 0.574 & 0.994 & 0.999 & - & - & - & - & - \\
    GRADIEND$_{\mathrm{1s}}$ & 0.727 & 0.871 & 0.789 & 0.975 & 0.922 & 0.787 & 0.783 & 0.796 & 0.994 & 0.998 & 0.999 & 0.989 & \textbf{1.000} & 0.975 & 0.996 \\
    \SummarySignalRule{16}
    CAGA$_{\mathrm{pw}}$ & 0.881 & 0.900 & 0.899 & 0.904 & 0.973 & 0.968 & 0.954 & 0.717 & 0.932 & 0.925 & - & - & - & - & - \\
    CAGA$_{\mathrm{1s}}$ & 0.881 & 0.687 & 0.863 & 0.827 & 0.955 & 0.974 & 0.956 & 0.826 & 0.951 & 0.953 & 1.000 & 0.999 & 0.978 & 0.973 & 0.999 \\
\bottomrule
  \end{tabular}
\end{table*}

\begin{table*}[p]
  \centering
  \tiny
  \setlength{\tabcolsep}{1pt}
  \renewcommand{\arraystretch}{0.8}
  \caption{\rev{Intervention raw results for \gemmaS.}}
  \label{tab:summary-concatenated-intervention-gemma3S}
  \begin{tabular}{l*{15}{c}}
    \toprule
    Method & \rotatebox{60}{\taskGender} & \rotatebox{60}{\taskEmotion} & \rotatebox{60}{\taskRace} & \rotatebox{60}{\taskReligion} & \rotatebox{60}{\taskRavelCont} & \rotatebox{60}{\taskRavelCountry} & \rotatebox{60}{\taskRavelLang} & \rotatebox{60}{\taskLangID} & \rotatebox{60}{\taskPronNum} & \rotatebox{60}{\taskPronPers} & \rotatebox{60}{\taskMIBIOI} & \rotatebox{60}{\taskKeyValue} & \rotatebox{60}{\taskInduction} & \rotatebox{60}{\taskRepetition} & \rotatebox{60}{\taskFuncComp} \\
    \midrule
   \multicolumn{16}{c}{$\Delta P^{+}$} \\
    \midrule
    CGA$_{\mathrm{pw}}$ & 0.067 & \textbf{0.052} & 0.060 & 0.016 & 0.054 & 0.055 & 0.000 & 0.000 & 0.077 & 0.075 & - & - & - & - & - \\
    CGA$_{\mathrm{1s}}$ & 0.148 & 0.001 & 0.052 & 0.022 & 0.288 & 0.053 & 0.000 & 0.000 & 0.064 & 0.108 & 0.350 & 0.183 & 0.188 & \textbf{0.160} & 0.281 \\
    \SummaryMethodRule{16}
    GRADIEND$_{\mathrm{pw}}$ & \textbf{0.444} & 0.045 & \textbf{0.074} & \textbf{0.139} & 0.101 & 0.077 & 0.000 & 0.000 & \textbf{0.124} & \textbf{0.217} & - & - & - & - & - \\
    GRADIEND$_{\mathrm{1s}}$ & 0.417 & 0.000 & 0.028 & 0.101 & 0.095 & 0.025 & 0.000 & 0.000 & 0.070 & 0.147 & 0.211 & \textbf{0.737} & \textbf{0.648} & 0.059 & \textbf{0.439} \\
    \SummarySignalRule{16}
    CAGA$_{\mathrm{pw}}$ & 0.066 & 0.001 & 0.031 & 0.010 & 0.189 & 0.028 & 0.053 & 0.000 & 0.038 & 0.071 & - & - & - & - & - \\
    CAGA$_{\mathrm{1s}}$ & 0.158 & -0.000 & 0.015 & 0.015 & 0.152 & 0.025 & \textbf{0.077} & 0.000 & 0.048 & 0.083 & \textbf{0.382} & 0.247 & 0.225 & 0.092 & 0.085 \\
\bottomrule
  \end{tabular}
\end{table*}

%%%%% Qwen 3.5 2B %%%%%%%%%%%

\begin{table*}[p]
  \centering
  \tiny
  \setlength{\tabcolsep}{1pt}
  \renewcommand{\arraystretch}{0.8}
  \caption{\rev{Detection raw results for \qwenS.}}
  \label{tab:summary-concatenated-detection-qwen3-2B}
  \begin{tabular}{l*{15}{c}}
    \toprule
    Method & \rotatebox{60}{\taskGender} & \rotatebox{60}{\taskEmotion} & \rotatebox{60}{\taskRace} & \rotatebox{60}{\taskReligion} & \rotatebox{60}{\taskRavelCont} & \rotatebox{60}{\taskRavelCountry} & \rotatebox{60}{\taskRavelLang} & \rotatebox{60}{\taskLangID} & \rotatebox{60}{\taskPronNum} & \rotatebox{60}{\taskPronPers} & \rotatebox{60}{\taskMIBIOI} & \rotatebox{60}{\taskKeyValue} & \rotatebox{60}{\taskInduction} & \rotatebox{60}{\taskRepetition} & \rotatebox{60}{\taskFuncComp} \\
    \midrule
    \multicolumn{16}{c}{AUC$_n$} \\
    \midrule
   CGA$_{\mathrm{pw}}$ & \textbf{1.000} & 0.900 & 0.989 & 0.925 & 0.997 & 1.000 & \textbf{1.000} & 0.971 & 0.988 & 0.973 & - & - & - & - & - \\
    CGA$_{\mathrm{1s}}$ & \textbf{1.000} & 0.689 & 0.988 & 0.939 & 1.000 & 0.999 & 0.999 & 0.921 & 0.992 & 0.973 & 0.998 & 0.953 & 0.941 & \textbf{1.000} & 0.998 \\
    \SummaryMethodRule{16}
    GRADIEND$_{\mathrm{pw}}$ & \textbf{1.000} & 0.614 & 0.623 & 0.795 & 0.999 & 0.980 & 0.999 & 0.517 & 0.756 & 0.870 & - & - & - & - & - \\
    GRADIEND$_{\mathrm{1s}}$ & \textbf{1.000} & 0.541 & 0.598 & 0.583 & 0.974 & 0.966 & 0.991 & 0.565 & 0.660 & 0.759 & \textbf{1.000} & 0.952 & 0.637 & \textbf{1.000} & 0.973 \\
    \SummarySignalRule{16}
    CAGA$_{\mathrm{pw}}$ & 0.925 & 0.916 & 0.956 & 0.968 & 0.950 & 0.924 & 0.992 & 0.629 & 0.938 & 0.960 & - & - & - & - & - \\
    CAGA$_{\mathrm{1s}}$ & 0.975 & 0.778 & 0.970 & 0.965 & 0.951 & 0.920 & 0.999 & 0.693 & 0.930 & 0.946 & \textbf{1.000} & 0.996 & 0.951 & 0.999 & 1.000 \\
\bottomrule
  \end{tabular}
\end{table*}

\begin{table*}[p]
  \centering
  \tiny
  \setlength{\tabcolsep}{1pt}
  \renewcommand{\arraystretch}{0.8}
  \caption{\rev{Intervention raw results for \qwenS.}}
  \label{tab:summary-concatenated-intervention-qwen}
  \begin{tabular}{l*{15}{c}}
    \toprule
    Method & \rotatebox{60}{\taskGender} & \rotatebox{60}{\taskEmotion} & \rotatebox{60}{\taskRace} & \rotatebox{60}{\taskReligion} & \rotatebox{60}{\taskRavelCont} & \rotatebox{60}{\taskRavelCountry} & \rotatebox{60}{\taskRavelLang} & \rotatebox{60}{\taskLangID} & \rotatebox{60}{\taskPronNum} & \rotatebox{60}{\taskPronPers} & \rotatebox{60}{\taskMIBIOI} & \rotatebox{60}{\taskKeyValue} & \rotatebox{60}{\taskInduction} & \rotatebox{60}{\taskRepetition} & \rotatebox{60}{\taskFuncComp} \\
    \midrule
   \multicolumn{16}{c}{$\Delta P^{+}$} \\
    \midrule
    CGA$_{\mathrm{pw}}$ & 0.022 & 0.011 & 0.044 & 0.033 & 0.338 & 0.095 & 0.149 & 0.000 & 0.025 & 0.049 & - & - & - & - & - \\
    CGA$_{\mathrm{1s}}$ & 0.125 & 0.001 & \textbf{0.098} & 0.042 & 0.392 & 0.045 & 0.199 & 0.000 & 0.051 & 0.089 & \textbf{0.528} & \textbf{0.682} & \textbf{0.594} & \textbf{0.171} & 0.064 \\
    \SummaryMethodRule{16}
    GRADIEND$_{\mathrm{pw}}$ & \textbf{0.433} & 0.000 & 0.000 & 0.010 & 0.397 & 0.046 & 0.221 & -0.000 & 0.018 & 0.080 & - & - & - & - & - \\
    GRADIEND$_{\mathrm{1s}}$ & 0.269 & 0.000 & 0.000 & 0.000 & 0.318 & 0.047 & 0.134 & -0.000 & 0.001 & 0.004 & 0.318 & 0.065 & -0.000 & 0.169 & \textbf{0.223} \\
    \SummarySignalRule{16}
    CAGA$_{\mathrm{pw}}$ & 0.155 & 0.005 & 0.004 & \textbf{0.092} & 0.400 & \textbf{0.128} & 0.236 & 0.000 & 0.068 & \textbf{0.109} & - & - & - & - & - \\
    CAGA$_{\mathrm{1s}}$ & 0.292 & 0.000 & 0.012 & 0.061 & \textbf{0.540} & 0.009 & \textbf{0.332} & 0.000 & \textbf{0.086} & 0.100 & 0.386 & 0.574 & 0.545 & 0.033 & 0.110 \\
\bottomrule
  \end{tabular}
\end{table*}

%%%%% Gemma-2-2B %%%%%%%%%%%

\begin{table*}[p]
  \centering
  \tiny
  \setlength{\tabcolsep}{1pt}
  \renewcommand{\arraystretch}{0.8}
  \caption{\rev{Detection raw results for \gemma.}}
  \label{tab:summary-concatenated-detection-gemma2}
  \begin{tabular}{l*{15}{c}}
    \toprule
    Method & \rotatebox{60}{\taskGender} & \rotatebox{60}{\taskEmotion} & \rotatebox{60}{\taskRace} & \rotatebox{60}{\taskReligion} & \rotatebox{60}{\taskRavelCont} & \rotatebox{60}{\taskRavelCountry} & \rotatebox{60}{\taskRavelLang} & \rotatebox{60}{\taskLangID} & \rotatebox{60}{\taskPronNum} & \rotatebox{60}{\taskPronPers} & \rotatebox{60}{\taskMIBIOI} & \rotatebox{60}{\taskKeyValue} & \rotatebox{60}{\taskInduction} & \rotatebox{60}{\taskRepetition} & \rotatebox{60}{\taskFuncComp} \\
    \midrule
   \multicolumn{16}{c}{AUC$_n$} \\
    \midrule
    CGA$_{\mathrm{pw}}$ & 1.000 & 0.994 & 0.973 & 0.964 & 0.999 & 0.997 & 1.000 & 0.628 & 0.978 & 0.954 & - & - & - & - & - \\
    CGA$_{\mathrm{1s}}$ & 0.893 & 0.788 & 0.982 & 0.985 & 0.999 & 0.998 & \textbf{1.000} & 0.720 & 0.990 & 0.976 & \textbf{1.000} & 0.999 & \textbf{1.000} & \textbf{1.000} & \textbf{1.000} \\
    \SummaryMethodRule{16}
    GRADIEND$_{\mathrm{pw}}$ & 0.856 & 0.987 & 0.934 & 0.950 & 0.986 & 0.882 & 0.992 & 0.694 & 0.972 & 0.997 & - & - & - & - & - \\
    GRADIEND$_{\mathrm{1s}}$ & 0.997 & 0.902 & 0.955 & 0.916 & 0.984 & 0.915 & 0.986 & 0.678 & 0.987 & 0.982 & \textbf{1.000} & 0.694 & 0.711 & 0.980 & 1.000 \\
    \SummarySignalRule{16}
    CAGA$_{\mathrm{pw}}$ & \textbf{1.000} & 0.983 & 0.964 & 0.958 & 0.999 & 1.000 & 1.000 & 0.813 & 0.924 & 0.954 & - & - & - & - & - \\
    CAGA$_{\mathrm{1s}}$ & 0.722 & 0.811 & 0.916 & 0.932 & 0.999 & 0.995 & \textbf{1.000} & 0.887 & 0.904 & 0.906 & 0.981 & \textbf{1.000} & 0.918 & 0.925 & \textbf{1.000} \\
\bottomrule
  \end{tabular}
\end{table*}

\begin{table*}[p]
  \centering
  \tiny
  \setlength{\tabcolsep}{1pt}
  \renewcommand{\arraystretch}{0.8}
  \caption{\rev{Intervention raw results for \gemma.}}
  \label{tab:summary-concatenated-intervention-gemma2}
  \begin{tabular}{l*{15}{c}}
    \toprule
    Method & \rotatebox{60}{\taskGender} & \rotatebox{60}{\taskEmotion} & \rotatebox{60}{\taskRace} & \rotatebox{60}{\taskReligion} & \rotatebox{60}{\taskRavelCont} & \rotatebox{60}{\taskRavelCountry} & \rotatebox{60}{\taskRavelLang} & \rotatebox{60}{\taskLangID} & \rotatebox{60}{\taskPronNum} & \rotatebox{60}{\taskPronPers} & \rotatebox{60}{\taskMIBIOI} & \rotatebox{60}{\taskKeyValue} & \rotatebox{60}{\taskInduction} & \rotatebox{60}{\taskRepetition} & \rotatebox{60}{\taskFuncComp} \\
    \midrule
   \multicolumn{16}{c}{$\Delta P^{+}$} \\
    \midrule
    CGA$_{\mathrm{pw}}$ & \textbf{0.491} & 0.005 & 0.087 & 0.019 & 0.357 & 0.133 & 0.111 & 0.000 & 0.002 & 0.072 & - & - & - & - & - \\
    CGA$_{\mathrm{1s}}$ & 0.028 & 0.000 & 0.048 & 0.031 & 0.536 & 0.040 & 0.239 & -0.000 & 0.038 & 0.110 & 0.227 & \textbf{0.757} & 0.229 & \textbf{0.182} & 0.049 \\
    \SummaryMethodRule{16}
    GRADIEND$_{\mathrm{pw}}$ & -0.000 & \textbf{0.045} & 0.138 & \textbf{0.221} & 0.515 & 0.200 & 0.232 & -0.000 & \textbf{0.120} & 0.177 & - & - & - & - & - \\
    GRADIEND$_{\mathrm{1s}}$ & 0.317 & 0.000 & 0.028 & 0.009 & \textbf{0.556} & 0.003 & 0.183 & -0.000 & 0.057 & 0.099 & 0.242 & 0.023 & 0.000 & 0.181 & 0.063 \\
    \SummarySignalRule{16}
    CAGA$_{\mathrm{pw}}$ & 0.083 & 0.002 & \textbf{0.138} & 0.149 & 0.524 & 0.084 & 0.060 & 0.000 & 0.064 & 0.098 & - & - & - & - & - \\
    CAGA$_{\mathrm{1s}}$ & 0.070 & -0.000 & 0.100 & 0.017 & 0.323 & 0.017 & 0.033 & 0.000 & 0.053 & 0.077 & 0.002 & 0.582 & \textbf{0.330} & 0.098 & \textbf{0.068} \\
\bottomrule
  \end{tabular}
\end{table*}

%%%%% Gemma-3-4B %%%%%%%%%%%

\begin{table*}[p]
  \centering
  \tiny
  \setlength{\tabcolsep}{1pt}
  \renewcommand{\arraystretch}{0.8}
  \caption{\rev{Detection raw results for \gemmaM.}}
  \label{tab:summary-concatenated-detection-gemma3-4b}
  \begin{tabular}{l*{15}{c}}
    \toprule
    Method & \rotatebox{60}{\taskGender} & \rotatebox{60}{\taskEmotion} & \rotatebox{60}{\taskRace} & \rotatebox{60}{\taskReligion} & \rotatebox{60}{\taskRavelCont} & \rotatebox{60}{\taskRavelCountry} & \rotatebox{60}{\taskRavelLang} & \rotatebox{60}{\taskLangID} & \rotatebox{60}{\taskPronNum} & \rotatebox{60}{\taskPronPers} & \rotatebox{60}{\taskMIBIOI} & \rotatebox{60}{\taskKeyValue} & \rotatebox{60}{\taskInduction} & \rotatebox{60}{\taskRepetition} & \rotatebox{60}{\taskFuncComp} \\
    \midrule
    \multicolumn{16}{c}{AUC$_n$} \\
    \midrule
    CGA$_{\mathrm{pw}}$ & \textbf{1.000} & 0.973 & 0.985 & 0.988 & 0.977 & 0.998 & 1.000 & 0.639 & 0.996 & 0.994 & - & - & - & - & - \\
    CGA$_{\mathrm{1s}}$ & \textbf{1.000} & 0.901 & 0.978 & \textbf{0.999} & 0.987 & 0.999 & \textbf{1.000} & 0.797 & 0.999 & 0.999 & \textbf{1.000} & \textbf{1.000} & \textbf{1.000} & \textbf{1.000} & \textbf{1.000} \\
    \SummaryMethodRule{16}
    GRADIEND$_{\mathrm{pw}}$ & \textbf{1.000} & 0.641 & 0.941 & 0.943 & 0.965 & 0.964 & 0.978 & 0.548 & 0.999 & 0.999 & - & - & - & - & - \\
    GRADIEND$_{\mathrm{1s}}$ & 0.992 & 0.518 & 0.909 & 0.934 & 0.949 & 0.937 & 0.858 & 0.605 & 0.978 & 0.790 & \textbf{1.000} & 0.529 & 0.501 & 1.000 & 0.871 \\
    \SummarySignalRule{16}
    CAGA$_{\mathrm{pw}}$ & 0.937 & 0.713 & 0.923 & 0.843 & 0.944 & 0.953 & 0.969 & 0.780 & 0.949 & 0.894 & - & - & - & - & - \\
    CAGA$_{\mathrm{1s}}$ & 0.892 & 0.747 & 0.868 & 0.739 & 0.957 & 0.982 & 0.973 & 0.752 & 0.929 & 0.892 & \textbf{1.000} & \textbf{1.000} & 0.981 & 0.999 & 1.000 \\
\bottomrule
  \end{tabular}
\end{table*}

\begin{table*}[p]
  \centering
  \tiny
  \setlength{\tabcolsep}{1pt}
  \renewcommand{\arraystretch}{0.8}
  \caption{Intervention raw results for \gemmaM.}
  \label{tab:summary-concatenated-intervention-gemma3L}
  \begin{tabular}{l*{15}{c}}
    \toprule
    Method & \rotatebox{60}{\taskGender} & \rotatebox{60}{\taskEmotion} & \rotatebox{60}{\taskRace} & \rotatebox{60}{\taskReligion} & \rotatebox{60}{\taskRavelCont} & \rotatebox{60}{\taskRavelCountry} & \rotatebox{60}{\taskRavelLang} & \rotatebox{60}{\taskLangID} & \rotatebox{60}{\taskPronNum} & \rotatebox{60}{\taskPronPers} & \rotatebox{60}{\taskMIBIOI} & \rotatebox{60}{\taskKeyValue} & \rotatebox{60}{\taskInduction} & \rotatebox{60}{\taskRepetition} & \rotatebox{60}{\taskFuncComp} \\
    \midrule
  \multicolumn{16}{c}{$\Delta P^{+}$} \\
    \midrule
    CGA$_{\mathrm{pw}}$ & 0.374 & \textbf{0.030} & 0.051 & 0.042 & 0.230 & 0.140 & 0.217 & 0.000 & 0.018 & 0.087 & - & - & - & - & - \\
    CGA$_{\mathrm{1s}}$ & 0.338 & 0.000 & 0.063 & 0.115 & 0.247 & \textbf{0.156} & 0.269 & 0.000 & 0.093 & 0.148 & 0.285 & \textbf{0.680} & \textbf{0.232} & 0.202 & \textbf{0.094} \\
    \SummaryMethodRule{16}
    GRADIEND$_{\mathrm{pw}}$ & 0.439 & 0.003 & \textbf{0.106} & \textbf{0.164} & 0.331 & 0.144 & \textbf{0.285} & -0.000 & \textbf{0.138} & \textbf{0.500} & - & - & - & - & - \\
    GRADIEND$_{\mathrm{1s}}$ & \textbf{0.566} & 0.000 & 0.031 & 0.043 & \textbf{0.333} & 0.097 & 0.267 & 0.000 & 0.050 & 0.035 & 0.084 & 0.005 & 0.016 & \textbf{0.202} & 0.013 \\
    \SummarySignalRule{16}
    CAGA$_{\mathrm{pw}}$ & 0.003 & 0.003 & 0.010 & 0.002 & 0.192 & 0.054 & 0.103 & 0.000 & 0.014 & 0.051 & - & - & - & - & - \\
    CAGA$_{\mathrm{1s}}$ & 0.002 & 0.000 & 0.012 & 0.002 & 0.078 & 0.010 & 0.181 & 0.000 & 0.020 & 0.107 & \textbf{0.380} & 0.387 & 0.201 & 0.028 & 0.025 \\
\bottomrule
  \end{tabular}
\end{table*}

%%%%% Mistral 7B %%%%%%%%%%%

\begin{table*}[p]
  \centering
  \tiny
  \setlength{\tabcolsep}{1pt}
  \renewcommand{\arraystretch}{0.8}
  \caption{\rev{Detection raw results for \mistral.}}
  \label{tab:summary-concatenated-detection-mistral}
  \begin{tabular}{l*{15}{c}}
    \toprule
    Method & \rotatebox{60}{\taskGender} & \rotatebox{60}{\taskEmotion} & \rotatebox{60}{\taskRace} & \rotatebox{60}{\taskReligion} & \rotatebox{60}{\taskRavelCont} & \rotatebox{60}{\taskRavelCountry} & \rotatebox{60}{\taskRavelLang} & \rotatebox{60}{\taskLangID} & \rotatebox{60}{\taskPronNum} & \rotatebox{60}{\taskPronPers} & \rotatebox{60}{\taskMIBIOI} & \rotatebox{60}{\taskKeyValue} & \rotatebox{60}{\taskInduction} & \rotatebox{60}{\taskRepetition} & \rotatebox{60}{\taskFuncComp} \\
    \midrule
    \multicolumn{16}{c}{AUC$_n$} \\
    \midrule
    CGA$_{\mathrm{pw}}$ & \textbf{1.000} & 0.975 & 0.972 & 0.958 & 1.000 & 0.999 & \textbf{1.000} & 0.664 & \textbf{0.996} & 0.972 & - & - & - & - & - \\
    CGA$_{\mathrm{1s}}$ & \textbf{1.000} & 0.944 & 0.995 & 0.985 & 1.000 & \textbf{1.000} & \textbf{1.000} & 0.654 & 0.978 & 0.980 & \textbf{1.000} & \textbf{1.000} & \textbf{1.000} & \textbf{1.000} & \textbf{1.000} \\
    \SummaryMethodRule{16}
    GRADIEND$_{\mathrm{pw}}$ & \textbf{1.000} & 0.936 & 0.825 & 0.923 & 0.943 & 0.869 & 0.939 & 0.603 & 0.625 & 0.971 & - & - & - & - & - \\
    GRADIEND$_{\mathrm{1s}}$ & 0.981 & 0.766 & 0.929 & 0.939 & 0.928 & 0.865 & 0.928 & 0.752 & 0.941 & 0.968 & 0.999 & 0.989 & 0.974 & 0.997 & 0.989 \\
    \SummarySignalRule{16}
    CAGA$_{\mathrm{pw}}$ & \textbf{1.000} & \textbf{0.997} & 0.981 & 0.990 & 0.954 & 1.000 & 0.992 & 0.564 & 0.983 & 0.974 & - & - & - & - & - \\
    CAGA$_{\mathrm{1s}}$ & \textbf{1.000} & 0.843 & 0.982 & \textbf{0.992} & 0.981 & \textbf{1.000} & 0.999 & 0.541 & 0.989 & 0.989 & \textbf{1.000} & \textbf{1.000} & 1.000 & \textbf{1.000} & 0.995 \\
\bottomrule
  \end{tabular}
\end{table*}

\begin{table*}[p]
  \centering
  \tiny
  \setlength{\tabcolsep}{1pt}
  \renewcommand{\arraystretch}{0.8}
  \caption{Intervention raw results for \mistral.}
  \label{tab:summary-concatenated-intervention-mistral}
  \begin{tabular}{l*{15}{c}}
    \toprule
    Method & \rotatebox{60}{\taskGender} & \rotatebox{60}{\taskEmotion} & \rotatebox{60}{\taskRace} & \rotatebox{60}{\taskReligion} & \rotatebox{60}{\taskRavelCont} & \rotatebox{60}{\taskRavelCountry} & \rotatebox{60}{\taskRavelLang} & \rotatebox{60}{\taskLangID} & \rotatebox{60}{\taskPronNum} & \rotatebox{60}{\taskPronPers} & \rotatebox{60}{\taskMIBIOI} & \rotatebox{60}{\taskKeyValue} & \rotatebox{60}{\taskInduction} & \rotatebox{60}{\taskRepetition} & \rotatebox{60}{\taskFuncComp} \\
    \midrule
   \multicolumn{16}{c}{$\Delta P^{+}$} \\
    \midrule
     CGA$_{\mathrm{pw}}$ & 0.174 & \textbf{0.006} & 0.020 & 0.029 & 0.272 & 0.008 & 0.296 & 0.000 & 0.017 & 0.039 & - & - & - & - & - \\
    CGA$_{\mathrm{1s}}$ & 0.010 & 0.001 & 0.104 & 0.046 & \textbf{0.644} & 0.002 & 0.168 & 0.000 & 0.061 & 0.076 & \textbf{0.603} & 0.778 & 0.554 & 0.216 & 0.243 \\
    \SummaryMethodRule{16}
    GRADIEND$_{\mathrm{pw}}$ & \textbf{0.487} & 0.006 & 0.046 & 0.060 & 0.446 & 0.102 & \textbf{0.360} & 0.000 & 0.005 & 0.043 & - & - & - & - & - \\
    GRADIEND$_{\mathrm{1s}}$ & 0.034 & 0.001 & 0.009 & 0.050 & 0.327 & 0.001 & 0.058 & 0.000 & 0.035 & 0.042 & 0.538 & \textbf{0.786} & 0.545 & \textbf{0.246} & 0.007 \\
    \SummarySignalRule{16}
    CAGA$_{\mathrm{pw}}$ & 0.205 & 0.001 & \textbf{0.124} & \textbf{0.101} & 0.491 & \textbf{0.250} & 0.324 & 0.000 & 0.045 & 0.076 & - & - & - & - & - \\
    CAGA$_{\mathrm{1s}}$ & 0.265 & 0.001 & 0.099 & 0.063 & 0.388 & 0.003 & 0.092 & -0.000 & \textbf{0.063} & \textbf{0.125} & 0.552 & 0.768 & \textbf{0.571} & 0.126 & \textbf{0.248} \\
\bottomrule
  \end{tabular}
\end{table*}

%%%%% DeepSeek %%%%%%%%%%%

\begin{table*}[p]
  \centering
  \tiny
  \setlength{\tabcolsep}{1pt}
  \renewcommand{\arraystretch}{0.8}
  \caption{\rev{Detection raw results for \deepseek.}}
  \label{tab:summary-concatenated-detection-deepseek}
  \begin{tabular}{l*{15}{c}}
    \toprule
    Method & \rotatebox{60}{\taskGender} & \rotatebox{60}{\taskEmotion} & \rotatebox{60}{\taskRace} & \rotatebox{60}{\taskReligion} & \rotatebox{60}{\taskRavelCont} & \rotatebox{60}{\taskRavelCountry} & \rotatebox{60}{\taskRavelLang} & \rotatebox{60}{\taskLangID} & \rotatebox{60}{\taskPronNum} & \rotatebox{60}{\taskPronPers} & \rotatebox{60}{\taskMIBIOI} & \rotatebox{60}{\taskKeyValue} & \rotatebox{60}{\taskInduction} & \rotatebox{60}{\taskRepetition} & \rotatebox{60}{\taskFuncComp} \\
    \midrule
    \multicolumn{16}{c}{AUC$_n$} \\
    \midrule
    CGA$_{\mathrm{pw}}$ & \textbf{1.000} & 0.988 & 0.976 & 0.978 & 0.996 & 1.000 & 0.999 & 0.629 & 0.994 & 0.984 & - & - & - & - & - \\
    CGA$_{\mathrm{1s}}$ & \textbf{1.000} & 0.924 & 0.967 & \textbf{1.000} & 0.999 & 1.000 & \textbf{1.000} & 0.787 & 0.996 & 0.994 & \textbf{1.000} & \textbf{1.000} & \textbf{1.000} & \textbf{1.000} & \textbf{1.000} \\
    \SummaryMethodRule{16}
    GRADIEND$_{\mathrm{pw}}$ & 1.000 & 0.957 & 0.922 & 0.936 & 0.955 & 0.721 & 0.923 & 0.538 & 0.769 & 0.980 & - & - & - & - & - \\
    GRADIEND$_{\mathrm{1s}}$ & 0.881 & 0.775 & 0.904 & 0.936 & 0.942 & 0.960 & 0.947 & 0.565 & 0.901 & 0.916 & 0.996 & 0.994 & 0.987 & 0.990 & 0.904 \\
    \SummarySignalRule{16}
    CAGA$_{\mathrm{pw}}$ & 0.896 & 0.839 & 0.761 & 0.913 & 0.909 & 0.924 & 0.874 & 0.674 & 0.915 & 0.840 & - & - & - & - & - \\
    CAGA$_{\mathrm{1s}}$ & 0.874 & 0.888 & 0.790 & 0.921 & 0.884 & 0.919 & 0.878 & 0.696 & 0.886 & 0.867 & 0.999 & 0.971 & 0.927 & 0.794 & 0.724 \\
\bottomrule
  \end{tabular}
\end{table*}

\begin{table*}[p]
  \centering
  \tiny
  \setlength{\tabcolsep}{1pt}
  \renewcommand{\arraystretch}{0.8}
  \caption{Intervention raw results for \deepseek.}
  \label{tab:summary-concatenated-intervention-deepseek}
  \begin{tabular}{l*{15}{c}}
    \toprule
    Method & \rotatebox{60}{\taskGender} & \rotatebox{60}{\taskEmotion} & \rotatebox{60}{\taskRace} & \rotatebox{60}{\taskReligion} & \rotatebox{60}{\taskRavelCont} & \rotatebox{60}{\taskRavelCountry} & \rotatebox{60}{\taskRavelLang} & \rotatebox{60}{\taskLangID} & \rotatebox{60}{\taskPronNum} & \rotatebox{60}{\taskPronPers} & \rotatebox{60}{\taskMIBIOI} & \rotatebox{60}{\taskKeyValue} & \rotatebox{60}{\taskInduction} & \rotatebox{60}{\taskRepetition} & \rotatebox{60}{\taskFuncComp} \\
    \midrule
   \multicolumn{16}{c}{$\Delta P^{+}$} \\
    \midrule
    CGA$_{\mathrm{pw}}$ & 0.317 & \textbf{0.010} & 0.027 & 0.062 & 0.296 & 0.084 & 0.203 & 0.000 & \textbf{0.088} & 0.039 & - & - & - & - & - \\
    CGA$_{\mathrm{1s}}$ & 0.099 & -0.001 & 0.020 & 0.090 & 0.315 & 0.043 & 0.349 & -0.000 & 0.030 & 0.047 & \textbf{0.483} & 0.337 & \textbf{0.447} & 0.117 & \textbf{0.330} \\
    \SummaryMethodRule{16}
    GRADIEND$_{\mathrm{pw}}$ & 0.484 & 0.000 & 0.075 & 0.057 & \textbf{0.482} & -0.000 & 0.286 & -0.000 & 0.019 & 0.048 & - & - & - & - & - \\
    GRADIEND$_{\mathrm{1s}}$ & 0.043 & 0.000 & 0.013 & 0.009 & 0.281 & 0.003 & 0.102 & -0.000 & 0.022 & 0.022 & 0.273 & 0.387 & 0.250 & 0.174 & 0.060 \\
    \SummarySignalRule{16}
    CAGA$_{\mathrm{pw}}$ & \textbf{0.574} & 0.001 & \textbf{0.097} & \textbf{0.216} & 0.406 & 0.146 & \textbf{0.438} & 0.000 & 0.074 & \textbf{0.109} & - & - & - & - & - \\
    CAGA$_{\mathrm{1s}}$ & 0.293 & -0.000 & 0.041 & 0.061 & 0.367 & 0.066 & 0.059 & 0.000 & 0.069 & 0.070 & 0.320 & \textbf{0.494} & 0.184 & \textbf{0.191} & -0.002 \\
\bottomrule
  \end{tabular}
\end{table*}

\begin{table*}[p]
  \centering
  \tiny
  \setlength{\tabcolsep}{1pt}
  \renewcommand{\arraystretch}{0.8}
  \caption{Detection raw results for \llama.}
  \label{tab:summary-concatenated-detection-llama3.18b}
  \begin{tabular}{l*{15}{c}}
    \toprule
    Method & \rotatebox{60}{\taskGender} & \rotatebox{60}{\taskEmotion} & \rotatebox{60}{\taskRace} & \rotatebox{60}{\taskReligion} & \rotatebox{60}{\taskRavelCont} & \rotatebox{60}{\taskRavelCountry} & \rotatebox{60}{\taskRavelLang} & \rotatebox{60}{\taskLangID} & \rotatebox{60}{\taskPronNum} & \rotatebox{60}{\taskPronPers} & \rotatebox{60}{\taskMIBIOI} & \rotatebox{60}{\taskKeyValue} & \rotatebox{60}{\taskInduction} & \rotatebox{60}{\taskRepetition} & \rotatebox{60}{\taskFuncComp} \\
    \midrule
    \multicolumn{16}{c}{AUC$_n$} \\
    \midrule
    CGA$_{\mathrm{pw}}$ & \textbf{1.000} & 0.869 & 0.969 & 0.947 & 1.000 & 0.992 & \textbf{1.000} & 0.801 & 0.980 & 0.911 & - & - & - & - & - \\
    CGA$_{\mathrm{1s}}$ & \textbf{1.000} & 0.813 & 0.945 & 0.925 & 0.998 & 0.940 & 1.000 & 0.952 & 0.996 & 0.967 & \textbf{1.000} & 1.000 & \textbf{1.000} & \textbf{1.000} & 0.999 \\
    \SummaryMethodRule{16}
    GRADIEND$_{\mathrm{pw}}$ & 1.000 & 0.590 & 0.752 & 0.893 & 0.994 & 0.891 & 0.985 & 0.591 & 0.847 & 0.968 & - & - & - & - & - \\
    GRADIEND$_{\mathrm{1s}}$ & 0.999 & 0.652 & 0.903 & 0.872 & 0.991 & 0.878 & 0.941 & 0.599 & 0.846 & 0.910 & 1.000 & 0.729 & 0.954 & \textbf{1.000} & 0.617 \\
    \SummarySignalRule{16}
    CAGA$_{\mathrm{pw}}$ & 1.000 & 0.930 & 0.965 & 0.990 & 1.000 & 0.855 & 0.905 & 0.680 & 0.990 & 0.933 & - & - & - & - & - \\
    CAGA$_{\mathrm{1s}}$ & \textbf{1.000} & 0.813 & 0.985 & 0.995 & 1.000 & 0.876 & 0.920 & 0.675 & 1.000 & 0.981 & \textbf{1.000} & \textbf{1.000} & 0.944 & 0.998 & 0.989 \\
\bottomrule
  \end{tabular}
\end{table*}

\begin{table*}[p]
  \centering
  \tiny
  \setlength{\tabcolsep}{1pt}
  \renewcommand{\arraystretch}{0.8}
  \caption{Intervention raw results for \llama.}
  \label{tab:summary-concatenated-intervention-llama3.18b}
  \begin{tabular}{l*{15}{c}}
    \toprule
    Method & \rotatebox{60}{\taskGender} & \rotatebox{60}{\taskEmotion} & \rotatebox{60}{\taskRace} & \rotatebox{60}{\taskReligion} & \rotatebox{60}{\taskRavelCont} & \rotatebox{60}{\taskRavelCountry} & \rotatebox{60}{\taskRavelLang} & \rotatebox{60}{\taskLangID} & \rotatebox{60}{\taskPronNum} & \rotatebox{60}{\taskPronPers} & \rotatebox{60}{\taskMIBIOI} & \rotatebox{60}{\taskKeyValue} & \rotatebox{60}{\taskInduction} & \rotatebox{60}{\taskRepetition} & \rotatebox{60}{\taskFuncComp} \\
    \midrule
   \multicolumn{16}{c}{$\Delta P^{+}$} \\
    \midrule
    CGA$_{\mathrm{pw}}$ & 0.035 & 0.005 & 0.089 & 0.084 & 0.493 & 0.153 & 0.327 & 0.000 & 0.017 & \textbf{0.082} & - & - & - & - & - \\
    CGA$_{\mathrm{1s}}$ & 0.215 & 0.000 & 0.049 & 0.054 & 0.513 & 0.045 & 0.205 & 0.000 & \textbf{0.050} & 0.047 & 0.524 & 0.811 & \textbf{0.539} & 0.166 & \textbf{0.435} \\
    \SummaryMethodRule{16}
    GRADIEND$_{\mathrm{pw}}$ & \textbf{0.398} & 0.000 & 0.034 & 0.037 & 0.501 & 0.064 & \textbf{0.416} & -0.000 & 0.025 & 0.034 & - & - & - & - & - \\
    GRADIEND$_{\mathrm{1s}}$ & 0.081 & 0.000 & 0.014 & 0.006 & 0.417 & 0.024 & 0.228 & 0.000 & 0.024 & 0.046 & \textbf{0.560} & 0.252 & 0.298 & \textbf{0.170} & 0.034 \\
    \SummarySignalRule{16}
    CAGA$_{\mathrm{pw}}$ & 0.236 & 0.001 & 0.074 & \textbf{0.153} & \textbf{0.727} & \textbf{0.211} & 0.275 & 0.000 & 0.031 & 0.059 & - & - & - & - & - \\
    CAGA$_{\mathrm{1s}}$ & 0.071 & 0.000 & \textbf{0.197} & 0.124 & 0.220 & 0.027 & 0.089 & 0.000 & 0.033 & 0.050 & 0.554 & \textbf{0.837} & 0.399 & 0.114 & 0.090 \\
\bottomrule
  \end{tabular}
\end{table*}

%%%%% Gemma-3-1B %%%%%%%%%%%

\begin{table*}[p]
  \centering
  \tiny
  \setlength{\tabcolsep}{1pt}
  \renewcommand{\arraystretch}{0.8}
  \caption{\rev{Detection raw results for \qwenL.}}
  \label{tab:summary-concatenated-detection-qwenL}
  \begin{tabular}{l*{15}{c}}
    \toprule
    Method & \rotatebox{60}{\taskGender} & \rotatebox{60}{\taskEmotion} & \rotatebox{60}{\taskRace} & \rotatebox{60}{\taskReligion} & \rotatebox{60}{\taskRavelCont} & \rotatebox{60}{\taskRavelCountry} & \rotatebox{60}{\taskRavelLang} & \rotatebox{60}{\taskLangID} & \rotatebox{60}{\taskPronNum} & \rotatebox{60}{\taskPronPers} & \rotatebox{60}{\taskMIBIOI} & \rotatebox{60}{\taskKeyValue} & \rotatebox{60}{\taskInduction} & \rotatebox{60}{\taskRepetition} & \rotatebox{60}{\taskFuncComp} \\
    \midrule
    \multicolumn{16}{c}{AUC$_n$} \\
    \midrule
    CGA$_{\mathrm{pw}}$ & \textbf{1.000} & 0.836 & 0.982 & 0.908 & 1.000 & 0.999 & 1.000 & 0.933 & 0.973 & 0.938 & - & - & - & - & - \\
    CGA$_{\mathrm{1s}}$ & 0.999 & 0.866 & 0.985 & 0.922 & \textbf{1.000} & 0.999 & 0.997 & 0.917 & 0.982 & 0.955 & \textbf{1.000} & \textbf{1.000} & \textbf{1.000} & \textbf{1.000} & 1.000 \\
    \SummaryMethodRule{16}
    GRADIEND$_{\mathrm{pw}}$ & \textbf{1.000} & 0.600 & 0.550 & 0.775 & 0.721 & 0.890 & 0.999 & 0.576 & 0.736 & 0.846 & - & - & - & - & - \\
    GRADIEND$_{\mathrm{1s}}$ & 0.999 & 0.598 & 0.760 & 0.798 & 0.993 & 0.976 & 0.969 & 0.699 & 0.806 & 0.844 & \textbf{1.000} & 0.984 & 0.621 & \textbf{1.000} & 0.965 \\
    \SummarySignalRule{16}
    CAGA$_{\mathrm{pw}}$ & 1.000 & \textbf{0.998} & 0.982 & 0.980 & 0.999 & 0.921 & 0.858 & 0.619 & 0.997 & 0.974 & - & - & - & - & - \\
    CAGA$_{\mathrm{1s}}$ & \textbf{1.000} & 0.907 & 0.983 & 0.979 & 0.999 & 0.844 & 0.819 & 0.774 & 0.999 & 0.989 & \textbf{1.000} & \textbf{1.000} & 0.986 & \textbf{1.000} & 1.000 \\
\bottomrule
  \end{tabular}
\end{table*}

\begin{table*}[p]
  \centering
  \tiny
  \setlength{\tabcolsep}{1pt}
  \renewcommand{\arraystretch}{0.8}
  \caption{Intervention raw results for \qwenL.}
  \label{tab:summary-concatenated-intervention-qwenL}
  \begin{tabular}{l*{15}{c}}
    \toprule
    Method & \rotatebox{60}{\taskGender} & \rotatebox{60}{\taskEmotion} & \rotatebox{60}{\taskRace} & \rotatebox{60}{\taskReligion} & \rotatebox{60}{\taskRavelCont} & \rotatebox{60}{\taskRavelCountry} & \rotatebox{60}{\taskRavelLang} & \rotatebox{60}{\taskLangID} & \rotatebox{60}{\taskPronNum} & \rotatebox{60}{\taskPronPers} & \rotatebox{60}{\taskMIBIOI} & \rotatebox{60}{\taskKeyValue} & \rotatebox{60}{\taskInduction} & \rotatebox{60}{\taskRepetition} & \rotatebox{60}{\taskFuncComp} \\
    \midrule
        \multicolumn{16}{c}{$\Delta P^{+}$} \\
    \midrule
    CGA$_{\mathrm{pw}}$ & 0.023 & 0.002 & 0.037 & 0.011 & 0.388 & 0.045 & 0.275 & \textbf{0.001} & 0.047 & 0.086 & - & - & - & - & - \\
    CGA$_{\mathrm{1s}}$ & 0.117 & 0.001 & 0.017 & 0.007 & 0.290 & 0.005 & 0.351 & -0.000 & \textbf{0.054} & \textbf{0.151} & 0.099 & \textbf{0.774} & 0.404 & 0.178 & 0.061 \\
    \SummaryMethodRule{16}
    GRADIEND$_{\mathrm{pw}}$ & \textbf{0.456} & 0.000 & 0.001 & 0.011 & 0.304 & 0.059 & \textbf{0.537} & 0.000 & 0.002 & 0.062 & - & - & - & - & - \\
    GRADIEND$_{\mathrm{1s}}$ & 0.055 & 0.000 & 0.000 & 0.002 & 0.148 & -0.000 & 0.041 & 0.000 & 0.008 & 0.029 & \textbf{0.495} & 0.428 & 0.000 & \textbf{0.204} & 0.088 \\
    \SummarySignalRule{16}
    CAGA$_{\mathrm{pw}}$ & 0.130 & 0.003 & \textbf{0.159} & \textbf{0.121} & \textbf{0.652} & \textbf{0.107} & 0.242 & 0.000 & 0.024 & 0.043 & - & - & - & - & - \\
    CAGA$_{\mathrm{1s}}$ & 0.188 & 0.001 & 0.123 & 0.032 & 0.310 & 0.013 & 0.135 & 0.000 & 0.031 & 0.052 & 0.460 & 0.742 & \textbf{0.419} & 0.133 & \textbf{0.108} \\
\bottomrule
  \end{tabular}
\end{table*}

\end{document}